\documentclass[twoside,11pt]{article}
\usepackage{amsthm}
\usepackage[nohyperref]{jmlr2e}
\input{"./preamble/packages"}

\makeatletter
\newcommand*{\addFileDependency}[1]{
  \typeout{(#1)}
  \@addtofilelist{#1}
  \IfFileExists{#1}{}{\typeout{No file #1.}}
}
\makeatother

\newacronym{ID}{ID}{Integrated Dirichlet}
\newacronym{NGP}{NGP}{Normalized Gaussian Process}
\newacronym{CE}{CE}{Cross Entropy}

\newacronym{NLU}{NLU}{natural language understanding}
\newacronym{MCMC}{MCMC}{Markov chain Monte Carlo}
\newacronym{DST}{DST}{Dempster–Shafer theory}
\newacronym{bba}{bba}{basic belief assignment}
\newacronym{BMA}{BMA}{Bayesian model averaging}

\newacronym{DNN}{DNN}{deep neural network}

\newacronym{BNN}{BNNs}{Bayesian neural networks}
\newacronym{MCD}{MC Dropout}{Monte Carlo Dropout}
\newacronym{MCBN}{MCBN}{Monte Carlo Batch Normalization}
\newacronym{SWAG}{SWAG}{Stochastic Weight Averaging-Gaussian}

\newacronym{UQ}{UQ}{uncertainty quantification}

\newacronym{MAP}{MAP}{maximum a posterior}

\newacronym{CDF}{CDF}{cumulative distribution function}
\newacronym{PDF}{PDF}{probability density function}
\newacronym{PMF}{PMF}{probability mass function}
\newacronym{MGF}{MGF}{moment generating function}

\newacronym{GLM}{GLM}{generalized linear model}
\newacronym{GAM}{GAM}{generalized additive model}
\newacronym{MLE}{MLE}{maximum-likelihood}

\newacronym{HMC}{HMC}{Hamiltonian Monte Carlo}
\newacronym{VI}{VI}{Variational Inference}

\newacronym{CRPS}{CRPS}{\textit{continous ranked probability score}}
\newacronym{RMSE}{RMSE}{root mean squared error}
\newacronym{MSE}{MSE}{mean squared error}

\newacronym{KL}{KL}{Kullback-Leibler}
\newacronym{CvM}{CvM}{Cram\'{e}r von Mises}
\newacronym{ELBO}{ELBO}{\emph{evidence lower bound}}

\newacronym{BNE}{BNE}{\textit{Bayesian Nonparametric Ensemble}}
\newacronym{BAE}{BAE}{\textbf{Bayesian Additive Ensemble}}
\newacronym{stack}{stack}{\textbf{Bayesian Stacking}}
\newacronym{BME}{BME}{\textbf{Bayesian Mixture of Experts}}

\newacronym{CVI}{Calibrated VI}{\textbf{Calibrated Variational Inference}}
\newacronym{EnKF}{EnKF}{Ensemble Kalman Filters}

\newacronym{RKHS}{RKHS}{reproducing kernel Hilbert space}
\newacronym{RBF}{RBF}{radial basis function}
\newacronym{CI}{CI}{Coverage Index}

\newacronym[plural=Gaussian processes,firstplural=Gaussian processes (GPs)]{GP}{GP}{Gaussian process}
\newacronym{CGP}{CGP}{\textit{constrained Gaussian process}}

\newacronym{SGD}{SGD}{stochastic gradient descent}

\newacronym{ECE}{ECE}{Expected Calibration Error}
\newacronym{EB}{EB}{empirical Bayes}

\newacronym{DDGP}{DDGP}{deep Dirichlet-Gaussian process}
\newacronym{ARD}{ARD}{automatic relevance determination}

\newacronym{IND}{IND}{in-domain}
\newacronym{OOD}{OOD}{out-of-domain}

\newacronym{SNGP}{SNGP}{\textit{Spectral-normalized Neural Gaussian Process}}
\newacronym{SVDKL}{SVDKL}{Stochastic Variational Deep Kernel Learning}

\newacronym{BERT}{BERT}{Bidirectional Encoder Representations from Transformers}

\newacronym{RFF}{RFF}{random Fourier feature}

\newacronym{SN}{SN}{spectral normalization}
\newacronym{DUQ}{DUQ}{Deterministic Uncertainty Quantification}
\newacronym{IsoMax+}{IsoMax+}{Enhanced Isotropic Maximization}
\newacronym{DUQ-GP}{DUQ-GP}{Deterministic Uncertainty Quantification with GP Last Layer}
\newacronym{MCDGP}{MCD-GP}{Calibrated Deep Gaussian Process}

\newacronym{AUROC}{AUROC}{Area Under Receiver Operating Characteristic}
\newacronym{AUPR}{AUPR}{Area Under Precision-Recall}

\newacronym{BvM}{BvM}{Bernstein-von Mises}

\newacronym{SVHN}{SVHN}{Street View House Numbers}

\newacronym{ML}{ML}{machine-learning}

\newacronym{EP}{EP}{expectation propagation}

\newacronym{OOS}{OOS}{out-of-scope}

\newcommand{\softmax}{\mathrm{softmax}}
\newcommand{\sigmoid}{\mathrm{sigmoid}}
\renewcommand{\logit}{\mathrm{logit}}

\usepackage{lastpage}
\jmlrheading{23}{2022}{1-\pageref{LastPage}}{5/22; Revised
12/22}{12/22}{22-0479}{
Jeremiah Liu, Shreyas Padhy, Jie Ren, Zi Lin, Yeming Wen, Ghassen Jerfel, Zack Nado, Jasper Snoek, Dustin Tran and Balaji Lakshminarayanan
}

\ShortHeadings{
A Simple Approach to Improve  Single-Model Deep Uncertainty via Distance-Awareness
}
 {Liu, Padhy, Ren, Lin, Wen, Jerfel, Nado, Snoek, Tran and Lakshminarayanan}
\firstpageno{1}

\makeatletter
\newcommand{\printfnsymbol}[1]{%
  \textsuperscript{\@fnsymbol{#1}}%
}
\makeatother

\begin{document}

\title{
A Simple Approach to Improve  Single-Model Deep Uncertainty via Distance-Awareness
}

\author{
\name \hspace{-0.5em} 
Jeremiah Zhe Liu\thanks{\ \  Equal contribution.} \email jereliu@google.com \\
\name Shreyas Padhy\footnotemark[1] \ \thanks{\ \ Work done at Google.} \email sp2058@eng.cam.ac.uk \\
\name Jie Ren\footnotemark[1] \email jjren@google.com \\
\name Zi Lin\footnotemark[2]  \email zil061@ucsd.edu \\
\name Yeming Wen\footnotemark[2]  \email ywen@utexas.edu \\
\name Ghassen Jerfel\footnotemark[2]  \email ghassen@waymo.com \\
\name Zachary Nado \email znado@google.com \\
\name Jasper Snoek \email jsnoek@google.com \\
\name Dustin Tran \email trandustin@google.com \\
\name Balaji Lakshminarayanan \email balajiln@google.com \\
\addr 
Google Research, Mountain View, CA 94043, USA \\
}

\editor{Philipp Hennig}

\maketitle
\begin{abstract}
Accurate uncertainty quantification is a major challenge in deep learning, as neural networks can make overconfident errors and assign high confidence predictions to 
out-of-distribution (OOD) inputs. The most popular approaches to estimate predictive uncertainty in deep learning are methods that combine predictions from multiple neural networks, such as Bayesian neural networks (BNNs) and deep ensembles. However their practicality in real-time, industrial-scale applications are limited due to the high memory and computational cost. Furthermore, ensembles and BNNs do not necessarily fix all the issues with the underlying member networks. 
In this work, we study principled approaches to improve the  uncertainty property of a single network, based on a single, deterministic representation. By formalizing the uncertainty quantification as a minimax learning problem, we first identify \emph{distance awareness}, i.e., the model's ability to quantify the distance of a testing example from the training data, 
as a necessary condition for a DNN to achieve high-quality (i.e., minimax optimal) uncertainty estimation. We then propose \textit{Spectral-normalized Neural Gaussian Process} (SNGP), a simple method that improves the distance-awareness ability of modern DNNs with two simple changes: (1) applying spectral normalization to hidden weights to enforce bi-Lipschitz smoothness in representations and (2) replacing the last output layer with a Gaussian process layer. On a suite of vision and language understanding benchmarks and on modern architectures (Wide-ResNet and BERT), SNGP consistently outperforms other single-model approaches in prediction, calibration and out-of-domain detection.
Furthermore, SNGP provides complementary benefits to popular techniques such as deep ensembles and data augmentation, making it a simple and scalable building block for probabilistic deep learning. 
Code is open-sourced at \url{https://github.com/google/uncertainty-baselines}. 
\end{abstract}

\section{Introduction}
In recent years, \gls{DNN} models have become ubiquitous in large-scale, real-world applications. The examples range from self-driving, image recognition, natural language understanding, to scientific applications including genomic sequence identification, genetic variant calling, drug discovery and protein design \citep{larson_evaluation_2019, GUPTA2021100057, jumper2021highly, poplin2018universal, ren2019likelihood, han2021reliable, kivlichan-etal-2021-measuring, roy2022does}.
A key characteristic shared by these real-world tasks is their \textit{risk sensitivity}: a confidently wrong decision from the deep learning model can lead to ethical violations, misleading scientific conclusions, and even fatal accidents \citep{amodei_concrete_2016}.
Therefore, to ensure safe and responsible deployment of AI technologies to the real world, it is of utmost importance to develop efficient approaches that reliably improves a deep neural network's uncertainty quality without compromising its practical utility (i.e., in terms of accuracy and scalability). 

It is well-known that a naively trained modern deep network tends to perform poorly in uncertainty tasks. They can be poorly calibrated \citep{guo_calibration_2017} or assign high confidence predictions to \gls{OOD} inputs \citep{nguyen2015deep,hendrycks_baseline_2017, lakshminarayanan_simple_2017}.  
This has led to the development of probabilistic methods tailored for deep neural networks, with examples including \gls{BNN} \citep{blundell_bayesian_2012, osawa_practical_2019, wenzel_how_2020}, Monte Carlo (MC) Dropout \citep{gal_dropout_2016}, and Deep Ensembles \citep{lakshminarayanan_simple_2017}. 
A shared characteristic of these approaches is that they are \textit{ensemble-based} methods: they need to maintain distribution samples over millions of model parameters, or require multiple forward passes to produce a final prediction. Consequently, despite their success in academic studies, these approaches can be computationally prohibitive for real-world usage \citep{ovadia_can_2019, wilson2020bayesian}. Furthermore, ensembles and BNNs may not necessarily fix all the problems with the underlying neural network in the first place. 
For instance, if all the ensemble members consistently make same mistakes or produce high confidence predictions far away from the data, the ensemble would inherit this behavior and also produce high confidence predictions far away from the data. 

In this work, we seek to address this gap by exploring an alternative direction to improve the uncertainty quality of a \textit{single} \gls{DNN}, so that the model achieves high-quality uncertainty with only a \textit{single, deterministic representation}. That is, instead of improving uncertainty by ensembling over multiple representations, we focus on addressing the design choices in a single \gls{DNN} that hinders its uncertainty performance. Specifically, we study principled inductive biases (i.e., mathematical conditions) that a model's hidden representation and output layer should satisfy to obtain good uncertainty performance. Then, we propose a simple, general-purpose algorithm to implement such inductive bias into a deep neural network. We expect this direction to be fruitful: an effective single-model uncertainty approach  unlocks high-quality uncertainty estimation in resource-constrained, real-world settings where Monte Carlo sampling is not feasible (e.g., on-device, real-time prediction), and also can be combined with existing ensemble techniques to produce new probabilistic models that improves the state-of-the-art.

\begin{figure}[th!]
    \centering
    \includegraphics[width=0.99\columnwidth]{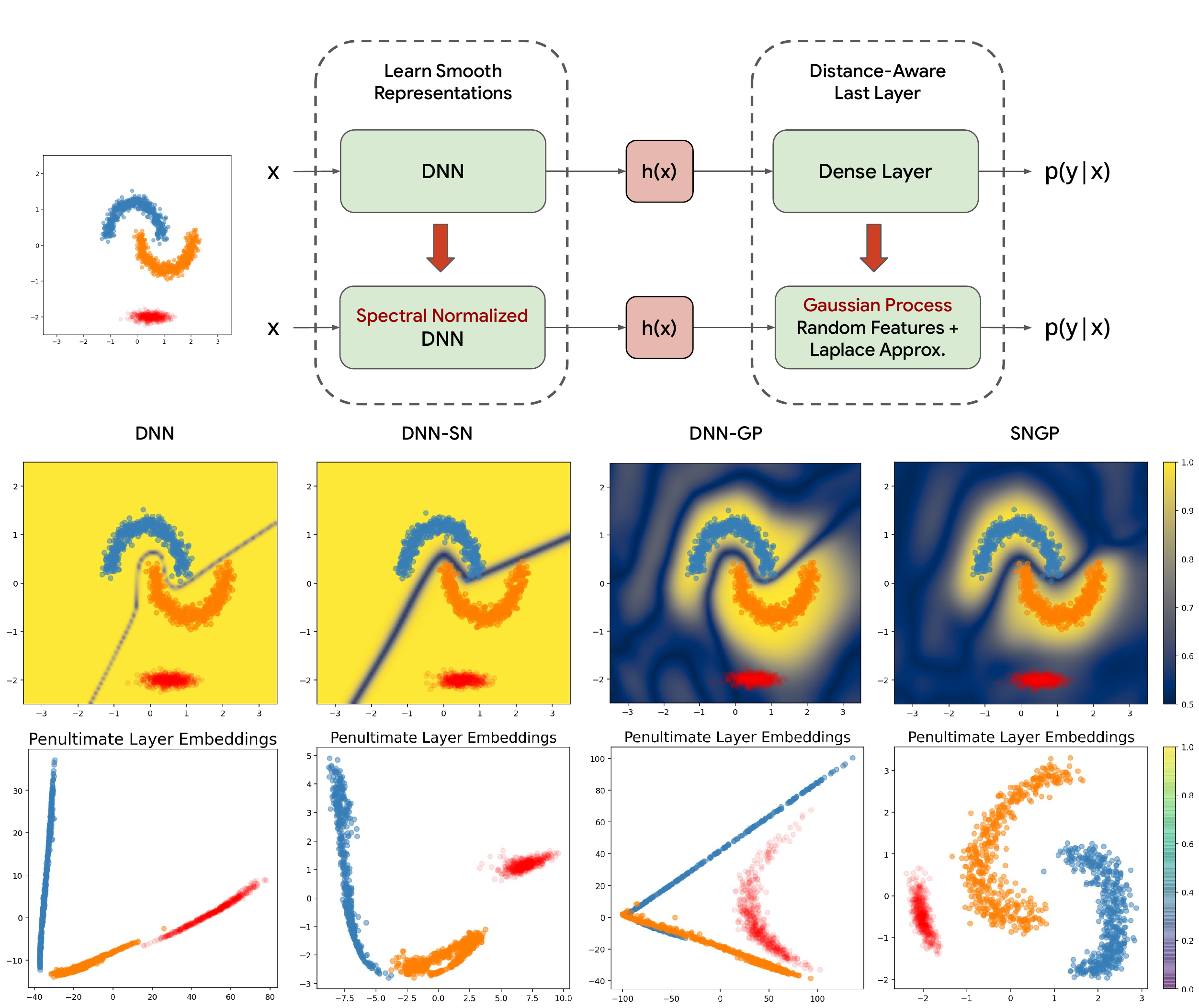}
    \caption{
    \small{
    SNGP improves the standard Deep Learning framework by (1) learning smooth representations from input data using Spectral normalization, and (2) enforcing distance-awareness in the 
    last layer of the neural network. In the \textit{two moons} example, a regular deep neural network \textbf{(DNN)} learns hidden embeddings that are highly degenerate and map both in-distribution and OOD points close together. In \textbf{DNN-SN}, the application of Spectral Norm improves the quality of hidden representations, however a dense last layer still results in high confidences away from the training manifold. In \textbf{DNN-GP}, a Gaussian Process last layer learns distance-aware confidences, however due to degenerate hidden representations, there is still high confidence away from the training manifold, especially near the OOD class. Finally, Spectral Norm is combined with a last-layer Gaussian Process in \textbf{SNGP}, where the model learns smooth representations of the training manifold, and shows low confidence away from the training distribution. In the top row, we plot the model confidence surface given by $\max{p(\bx)}$ for all methods, and in the bottom row we plot the 2-dimensional TSNE projection of the penultimate layer embeddings for a 6-layer ResNet model with 128 hidden units per layer.
    }}
    \label{fig:sngp_overview}
\end{figure}

Specifically, we first propose \textit{distance awareness}, i.e., a model's output prediction is aware of the distance between a new test example and the previously trained-upon examples, as an important property for the neural network to achieve high-quality uncertainty. Formally, this means that a predictive model $f(\bx)$ has the ability to output a scalar uncertainty estimate $u(\bx)$ that is monotonic with respect to an {\color{black} appropriate distance metric} between $\bx$ and the training examples (Definition \ref{def:da}). In the literature, several prior investigations have suggested the connection between a neural model's uncertainty quality and its ability to distinguish distances in the input space. For example, \cite{kristiadi_being_2020} shows that deep classifiers degrades in performance and becomes increasing overconfident when making predictions on examples that further from the support of the training set. On the other hand, \cite{van2021feature} and \cite{behrmann2021understanding} observe that deep models often collapse the representations of distinct examples onto the same location in the hidden space (i.e., ``feature collapse"), preventing the model from distinguishing between the familiar and unfamiliar examples and causing difficulty in detecting \gls{OOD} inputs. In this work, we formalize the intuition behind these empirical findings by providing a precise definition of the \textit{distance awareness}. We further provide rigorous arguments for the importance of the distance awareness property for a model's uncertainty performance, by casting uncertainty estimation as a minimax decision making problem and showing that distance awareness constitutes a necessary condition for obtaining the optimal solution (Section \ref{sec:theory}).

We then move to identify simple, general-purpose algorithm to implement the distance-awareness principle into a \gls{DNN} model. Figure~\ref{fig:sngp_overview} gives an high level overview.  We take inspiration from the classic probablistic learning literature, where a shallow \gls{GP}  model (when equipped with certain kernel) is known to perfectly achieve distance awareness \citep{rasmussen_gaussian_2006}.
However, the shallow \gls{GP} model tends to generalize poorly for high-dimensional data, since the common choices of kernel function lack the ability to adaptively capture the intrinsic structure underlying the data's high surface dimension, leading to the issue of curse of dimensionality \citep{bach_breaking_2017}. Later works mitigate this issue by equipping the \gls{GP} with a dimension-reduction feature extractor (e.g., implemented by a \gls{DNN})
\citep{salakhutdinov_bayesian_2008, damianou_deep_2013, wilson_deep_2015, wilson_stochastic_2016, calandra_manifold_2016, salimbeni_doubly_2017, bradshaw_adversarial_2017}. However, naively combining a \gls{DNN} with a \gls{GP} layer does not automatically guarantee distance awareness even with end-to-end training, as the hidden representation can still suffer the issue of feature collapse (as we will show in experiments, c.f.~Figures \ref{fig:sngp_overview} and \ref{fig:2d_exp}). 

To this end, we propose a simple and efficient algorithm for probabilistic deep learning which we term \gls{SNGP}. 
As shown in Figure \ref{fig:sngp_overview}, 
given an existing \gls{DNN} architecture, \gls{SNGP} makes two simple modifications to the neural network, namely 
\begin{enumerate}[leftmargin=2em]
    \item 
    adding \textit{spectral normalization} to the model's hidden layers,
and 
\item replacing the typical dense output layer with a distance-aware Gaussian Process.
\end{enumerate}

We show that spectral normalization improves the distance-preservation property of learned representations by bounding the hidden-space distance $||h(\bx) - h(\bx')||_H$ with respect to ${\color{black}{\color{black}d_X}(\bx, \bx')}$, where $\bx$ and $\bx'$ are two inputs to the feature extractor of the \glspl{DNN} $h(\bx)$, and $d_X$ is a {\color{black} suitable distance metric} defined for the data manifold (Proposition \ref{thm:resnet_lipschitz}). We then build a GP output layer on these representations $h(\bx)$. To ensure computational scalability, we use a Laplace approximation to the random feature expansion of the GP. This results in a model posterior that can be learned scalably and in closed form, and lets us efficiently compute the predictive uncertainty on individual inputs without having to resort to computationally expensive methods such as multiple forward passes (Section \ref{sec:method}).

We conduct a comprehensive study to investigate the behavior of \gls{SNGP} model across data modalities. As we will show, the SNGP approach  improves the calibration and \gls{OOD} detection performance of a deterministic \gls{DNN}, and  outperforms or is competitive with other single model uncertainty approaches. 
We further illustrate the method's scalability by adapting it to large-scale recognition tasks (i.e., ImageNet), and illustrate the method's generality by extending it to additional tasks in natural language understanding and genomics sequence identification (Section \ref{sec:experiments}). Finally, we show that SNGP can serve as a strong building block for  probabilistic deep learning approaches and provides complementary benefits to other state-of-the-art techniques such as ensembling (e.g., MC Dropout, Deep Ensemble) and data augmentation (e.g., AugMix), providing orthogonal improvements in uncertainty quantification (Section \ref{sec:building_block}).

\section{Theoretical Motivation for Distance Awareness}
\label{sec:theory}

\paragraph{Notation and Problem Setup} 
Let us consider a data-generating distribution {\color{black} $p^*(y, \bx) = p^*(y|\bx)p^*(\bx)$}, where $y \in \{1, \dots, K\}$ is the $K$-dimensional simplex of labels, and $\bx \in \Xsc \subset \real^d$ is the input data present on a manifold 
with a suitable metric $d_X: \Xsc \times \Xsc \rightarrow \real$. In particular, such a metric is 
tailored toward the geometry of $\Xsc$ such that{\color{black}, intuitively, the distance $d_X(\bx_1, \bx_2)$ between a pair of examples $\bx_1, \bx_2 \in \Xsc$ reflects a meaningful difference in the input space (e.g., for a pair of sentences $\bx_1, \bx_2$, $d_X(\bx_1, \bx_2)$ describes their semantic similarity rather than the token-level edit distance \citep{cer_semeval-2017_2017}).}
In a supervised learning setup, the goal is often to learn the conditional distribution $p^*(y|\bx)$, and the training data $\Dsc=\{y_i, \bx_i\}_{i=1}^N$ is often a subset of the full input space $\Xsc_{\texttt{IND}} \subset \Xsc$. Due to this fact, we can represent the conditional data-generating distribution $p^*(y | \bx)$ as a mixture of an \gls{IND} distribution $p_{\texttt{IND}}(y|\bx) = p^*(y|\bx, \bx \in \Xsc_{\texttt{IND}})$ and an \gls{OOD} distribution with non-overlapping support $p_{\texttt{OOD}}(y|\bx)=p^*(y|\bx, \bx \not\in \Xsc_{\texttt{IND}})$ \citep{meinke_towards_2020, scheirer_probability_2014}:
\begin{alignat}{4}
    p^*(y|\bx)
    &= p^*(y|\bx, \bx \in \Xsc_{\texttt{IND}}) \times p^*(\bx \in \Xsc_{\texttt{IND}}) 
    && + p^*(y|\bx, \bx \not\in \Xsc_{\texttt{IND}}) \times p^*(\bx \not\in \Xsc_{\texttt{IND}}).
    \label{eq:true_dist}
\end{alignat}
During the process of training, the model learns the in-domain predictive distribution $p^*(y|\bx, \bx \in \Xsc_{\texttt{IND}})$ from the training data $\Dsc$, but does not have knowledge about $p^*(y|\bx, \bx \not\in \Xsc_{\texttt{IND}})$.

In the example of an image classification model trained on MNIST, the out-of-domain space $\Xsc_{\texttt{OOD}} = \Xsc / \Xsc_{\texttt{IND}}$ is the space of all images that do not contain handwritten characters with the classes 0 to 9, which can include other datasets such as CIFAR-10 or ImageNet, with the potential for some overlap whenever numbers are present in such image examples.

In the example of a weather-service chatbot, the out-of-domain space $\Xsc_{\texttt{OOD}} = \Xsc / \Xsc_{\texttt{IND}}$ is the space of all natural utterances not related to weather queries, whose elements usually do not have a meaningful correspondence with the in-domain intent labels $y \in \{1, \dots, K\}$.
The out-of-domain distribution $p^*(y|\bx, \bx \not\in \Xsc_{\texttt{IND}})$ can in general be very different from the in-domain distribution $p^*(y|\bx, \bx \in \Xsc_{\texttt{IND}})$, and it is usually expected that the model will only generalize well within $\Xsc_{\texttt{IND}}$. However, during testing and deployment, the model is expected to construct a predictive distribution $p(y|\bx)$ for the entire input space, $\Xsc=\Xsc_{\texttt{IND}} \cup \Xsc_{\texttt{OOD}}$, since the gamut of input data that the model can be deployed on can come from anywhere, and not just a curated training distribution.




\vspace{-0.5em}
\subsection{Uncertainty Estimation as a Minimax Learning Problem} 
\label{sec:minimax}
In order to formulate uncertainty estimation as a learning problem under (\ref{eq:true_dist}), we need to define a loss function to measure a model $p(y|\bx)$'s quality of predictive uncertainty. One popular metric, the \gls{ECE}, is defined as $C(p, p^*) = E\big[ |E(y^*=\hat{y}|\hat{p}=p) - p | \big]$, and measures the difference in expectation between the model's predictive confidence (e.g., the maximum probability score) and its actual accuracy \citep{guo_calibration_2017,nixon2019measuring}. However, \gls{ECE} is not suitable as a loss function, since it does not have a unique minimum at the true solution $p = p^*$. Using the ECE directly can result in trivial counterexamples where a predictor ignores the input example and achieve perfect calibration by predicting randomly according to the marginal distribution of the labels \citep{gneiting_probabilistic_2007}.

To this end, a more theoretically sound
uncertainty metric needs to be obtained, which we accomplish by examining the rich literature of  \textit{strictly proper scoring rules} \citep{gneiting_strictly_2007} $s(.,p^*)$, which are loss functions that are uniquely minimized by the true distribution $p=p^*$.
These family of loss functions include a lot of the more commonly used examples such as log-loss and Brier score. Another nice property of proper scoring rules is that they're related to \gls{ECE}; both the log-loss and the Brier score are upper bounds of the calibration error, and this can be shown by the classic calibration-refinement decomposition   \citep{brocker2009reliability}. Therefore, it follows that if a proper scoring rule is minimized, it implies that the calibration error of the model is also being minimized.
We can now formalize the problem of uncertainty quantification as the problem of constructing an optimal predictive distribution $p(y|\bx)$ that minimizes the expected risk over all $\bx \in \Xsc$, i.e., an \textit{Uncertainty Risk Minimization} problem\footnote{It is interesting to note that, as a special case, (\ref{eq:brier_risk}) reduces back to the familiar \gls{MLE} objective when $s$ is the logarithm score. See \cite{gneiting_strictly_2007} for further detail. However, the analysis here goes beyond the scope of empirical risk parameter estimation, as it focuses on risk functional in terms of a infinite-dimensional parameter $p$ and concerns the out-of-domain situations where the training data is not available.}:
\begin{align}
    \inf_{p \in \Psc} S(p, p^*) = 
    \inf_{p \in \Psc} \underset{(\bx, y) \in \Xsc \times \Ysc}{E}
    \big[s \big( p(y|\bx), p^*(y|\bx) \big) \big].
    \label{eq:brier_risk}
\end{align}
Unfortunately, we cannot minimize (\ref{eq:brier_risk}) over the entire input space $\Xsc$, even with access to infinite amounts of in-domain data. This is due to the fact that during the training process, the data is collected only from $\Xsc_{\texttt{IND}}$, and the true \gls{OOD} distribution $p^*(y|\bx, \bx \not\in \Xsc_{\texttt{IND}})$ can never be learned by the model, and therefore generalization is not guaranteed since we do not make the assumption that $p^*(y|\bx, \bx \in \Xsc_{\texttt{IND}})$ and $p^*(y|\bx, \bx \not\in \Xsc_{\texttt{IND}})$ are similar. In practice, using a model trained only with in-domain data to make predictions on \gls{OOD} can lead to arbitrarily bad results, since nature can contain many \gls{OOD} distributions $p^*(y|\bx, \bx \not\in \Xsc_{\texttt{IND}})$ that are very dissimilar with the training data.
This is clearly undesirable for safety-critical applications. 

We therefore reformulate the problem using a more prudent strategy;
minimize instead the \textit{worst-case} risk with respect to all possible $p^* \in \Psc^*$, {\color{black} where $\Psc^*$ is the space of possible data-generating distributions whose in-domain component $p^*(y|\bx, \bx \in \Xsc_{\texttt{IND}})$ generates the observational data, while whose out-of-domain component $p^*(y|\bx, \bx \not\in \Xsc_{\texttt{IND}})$ is unconstrained and can be arbitrary.}
That is, we seek to construct a $p(y|\bx)$ to minimize the \textit{Minimax Uncertainty Risk}:
\begin{align}
\underset{p \in \Psc}{\operatorname{inf}}  \Big[ \sup_{p^* \in \Psc^*} \; S(p, p^*) \Big].
\label{eq:minimax_loss}
\end{align}
where 
$S(p, p^*)$ is the expected risk as defined in (\ref{eq:brier_risk}).
This reformulation can be viewed from a game-theoretic lens; the uncertainty estimation task is acting as a two-player game with the model and nature, where the goal of the model is to produce a minimax strategy $p$ that minimizes the risk $S(p, p^*)$ against all possible (even adversarial) moves $p^*$ of nature.
Under the task of classification using the Brier score as a proper scoring rule, the solution to the minimax problem (\ref{eq:minimax_loss}) adopts a simple and elegant form: 
\begin{align}
    p(y |\bx) 
    &= 
    p(y|\bx, \bx \in \Xsc_{\textup{\texttt{IND}}}) \times p^*(\bx \in \Xsc_{\textup{\texttt{IND}}}) 
    + 
    p_{\textup{\texttt{uniform}}}(y|\bx, \bx \not\in \Xsc_{\textup{\texttt{IND}}}) \times p^*(\bx \not\in \Xsc_{\textup{\texttt{IND}}}).
    \label{eq:minimax_solution}
\end{align}
(\ref{eq:minimax_solution}) has a very intuitive understanding; we should trust the model for an input point that lies in the training data domain, and otherwise make a maximum entropy (uniform) prediction.\footnote{Noted that this uniform strategy is conservative and is derived under the minimax assumption that (1) the OOD distribution is adversarial, and (2) all the classes in $y$ are semantically distinct. That is, in the OOD regions, $p^*(y|\bx)$ is always as far away from $p(y|\bx)$ as possible and puts its probability mass on an class that is the most different from the model prediction. For example, for a cancer prognosis model, the $p^*$ puts the probability on a different cancer type that requires a completely different treatment. 
In practice, there exist situations where the OOD class is semantically related to an in-domain class (e.g., pickup truck v.s. truck for the CIFAR100 v.s. CIFAR10 OOD detection problem) such that the data is not completely adversarial. In this case, a non-uniform distribution is still sensible in the practical context. However that falls outside the scope of the minimax analysis that we consider here.
}
For the practice of uncertainty estimation, (\ref{eq:minimax_solution}) is conceptually important in that it verifies that 
there exists a unique optimal solution to the uncertainty estimation problem (\ref{eq:minimax_loss}). Furthermore, this optimal solution can be constructed conveniently as a mixture of a discrete uniform distribution $p_{\texttt{uniform}}$ and the in-domain predictive distribution $p(y|\bx, \bx \in \Xsc_{\textup{\texttt{IND}}})$ that the model has already learned from data, \textit{assuming one can quantify $p^*(\bx \in \Xsc_{\textup{\texttt{IND}}})$ well}. 
In fact, the expression (\ref{eq:minimax_solution}) can be shown to be optimal for a broad family of scoring rules known as the Bregman scores, which includes the Brier score and the widely used \textit{log score} as the special cases. 

\paragraph{Proof Sketch:}
{\color{black} The proof for (\ref{eq:minimax_solution}) relies on the following key lemma (proved in Appendix \ref{sec:minimax_lemma_proof}):

\begin{lemma}[$p_{\textup{\texttt{uniform}}}$ is Optimal for Minimax Bregman Score in $\bx \not\in \Xsc_{IND}$] $ $\\
Consider the Bregman score in (\ref{eq:bregman_score}). At a location $\bx \not\in \Xsc_{IND}$ where the model has no information about $p^*$ other than $\sum_{k=1}^K p(y_k|\bx) = 1$, the solution to the minimax problem
$$\inf_{p\in \Psc} \sup_{p^* \in \Psc^*} s(p, p^*|\bx)$$
is the discrete uniform distribution, i.e., $p_{\textup{\texttt{uniform}}}(y_k|\bx)=\frac{1}{K} \;\;\; \forall k \in \{1, \dots, K\}$.
\label{thm:minimax_lemma_main}
\end{lemma}
Using Lemma \ref{thm:minimax_lemma_main}, (\ref{eq:minimax_solution}) can be proved easily by decomposing $p^*$ into its in-domain and out-of-domain components and apply standard minimax arguments. A proof sketch is included below: 

\noindent{\it Proof Sketch.\quad}
Denote $\Xsc_{\texttt{OOD}} = \Xsc / \Xsc_{\texttt{IND}}$.
Decompose the overall Bregman risk by domain:
\begin{align*}
    S(p, p^*) 
    &= 
    E_{x \in \Xsc}\big(s(p, p^*|\bx )\big) = 
    \int_{\Xsc} s\big(p, p^*|\bx \big)p^*(\bx) d\bx \\
    &= S_{\texttt{IND}}(p, p^*) * p^*(\bx \in \Xsc_{\texttt{IND}}) +
    S_{\texttt{OOD}}(p, p^*) * p^*(\bx \in \Xsc_{\texttt{OOD}}).
\end{align*}
where we have denoted $S_{\texttt{IND}}(p, p^*)=E_{\bx \in \Xsc_{\texttt{IND}}}\big(s(p, p^*|\bx )\big)$ and $S_{\texttt{OOD}}(p, p^*)=E_{\bx \in \Xsc_{\texttt{OOD}}}\big(s(p, p^*|\bx )\big)$.
Noting that $S_{\texttt{IND}}(p, p^*)$ and $S_{\texttt{OOD}}(p, p^*)$ have disjoint support, we can decompose the minimax risk as follows: 
\begin{align}
    \inf_{p}\sup_{p^*} S(p, p^*) 
    &= 
    \inf_{p}\sup_{p^*} \big[S_{\texttt{IND}}(p, p^*)\big] * 
    p^*(\bx \in \Xsc_{\texttt{IND}}) + 
    \inf_{p}\sup_{p^*} \big[S_{\texttt{OOD}}(p, p^*)\big] * 
    p^*(\bx \in \Xsc_{\texttt{OOD}}),
    \label{eq:minimax_decomposed_main}
\end{align}
Since the model's predictive distribution is learned from data, the in-domain minimax risk is fixed. Therefore, we only need to show $p_{\texttt{uniform}}$ is the optimal and unique solution to the out-of-domain minimax risk $\inf_{p}\sup_{p^*} \big[S_{\texttt{OOD}}(p, p^*)\big]$. To this end, notice that for a given $p$:
\begin{align}
    \sup_{p^*\in \Psc^*} \big[S_{\texttt{OOD}}(p, p^*)\big] 
    = 
    \int_{\Xsc_{\texttt{OOD}}} 
    \sup_{p^*}[s(p, p^*|\bx)] p(\bx|\bx\in \Xsc_{\texttt{OOD}}) d\bx,
\end{align}
due to the fact that we don't impose assumption on $p^*$ (therefore $p^*$ is free to attain the global supreme by maximizing $s(p, p^*|\bx)$ at every single location $\bx \in \Xsc_{\texttt{OOD}}$). Furthermore, there exists $p$ that minimize $\sup_{p^*} s(p, p^*|\bx)$ at every location of $\bx\in \Xsc_{\texttt{OOD}}$, then it minimizes the integral \citep{berger_statistical_1985}. By Lemma \ref{thm:minimax_lemma_main}, such $p$ exists and is unique, i.e.:
\begin{align*}
    p_{\textup{\texttt{uniform}}} = \underset{p\in \Psc}{\mathrm{arginf}}\sup_{p^*\in \Psc^*} S_{\texttt{OOD}}(p, p^*).
\end{align*}
In conclusion, we have shown that $p_{\textup{\texttt{uniform}}}$ is the unique solution to $\inf_{p}\sup_{p^*} S_{\texttt{OOD}}(p, p^*)$, and therefore that the unique solution to (\ref{eq:minimax_decomposed_main}) is 
(\ref{eq:minimax_solution}).
\hfill\BlackBox

A complete version of the proof is available in Appendix \ref{sec:minimax_app}.
}


\vspace{-0.5em}
\subsection{Distance Awareness as a Necessary Condition}
\label{sec:da}

Following from Equation (\ref{eq:minimax_solution}), it stands to reason that a key recipe for a deep learning model to be able to reliably estimate predictive uncertainty 
is its ability to quantify (explicitly or implicitly) the domain probability $p(\bx \in \Xsc_{\textup{\texttt{IND}}})$.
This requires that the model have a good notion of the distance (or dissimilarity) between a testing example $\bx$ and the training data $\Xsc_{\texttt{IND}}$ with respect to a \textit{suitable} metric $d_X$ for the data manifold (e.g.,  \textit{semantic textual similarity} \citep{cer_semeval-2017_2017} for language data). Definition \ref{def:da} makes this notion more precise:
\begin{definition}[Distance Awareness] Consider a predictive distribution $p(y|\bx)$ trained on a domain $\Xsc_{\textup{\texttt{IND}}} \subset \Xsc$, where $(\Xsc, d_X)$ is the input data manifold equipped with a suitable metric $d_X$. We say $p(y|\bx)$ is \underline{input distance aware} if there exists $u(\bx)$ a summary statistic of $p(y|\bx)$ that quantifies model uncertainty  (e.g., entropy, predictive variance, etc) and reflects the distance between $\bx$ and the training data with respect to $d_X$, i.e.,
$$u(\bx) = v\big(
d(\bx, \Xsc_{\textup{\texttt{IND}}})\big).$$
Here, $v$ is a monotonic function and $d(\bx, \Xsc_{\textup{\texttt{IND}}}) = E_{\bx' \sim \Xsc_{\textup{\texttt{IND}}}} d_X(\bx, \bx')$
is the distance between $\bx$ and the training data domain. 
\label{def:da}
\end{definition}
In the literature, there have been many models that attempt to enforce \textit{distance-awareness}, and here we consider a classic model that satisfies this property: a \glsfirst{GP} with a \gls{RBF} kernel.
For a classification task, 
the predicted class probability $r(g(\bx))$ is a nonlinear transformation of the \gls{GP} posterior $g \sim GP$, i.e., $r(.)=\mathrm{sigmoid}(.)$ for binary classification. The predictive uncertainty can then be expressed by the posterior variance $u(\bz_{test})=\var(g(\bz_{test})) = 1 - \bk^{\top}_{test}\bV\bk_{test}$ for $\bk_{test,i}=\exp(-\frac{1}{2}||\bz_{test} - \bz_i||^2)$ and $\bV_{N \times N}$  a fixed matrix determined by data. 
Then $u(\bx^*)$ increases monotonically toward 1 as $\bx^*$ moves further away from $\Xsc_{\texttt{IND}}$
(\cite{rasmussen_gaussian_2006}, Chapter 3.4).
It follows from the expression (\ref{eq:minimax_solution}) that the \textit{distance awareness} property is important for both calibration and \gls{OOD} detection. However, there is no guarantee of this property for a typical deep learning model \citep{hein_why_2019}. For example, consider a discriminative deep classifier with a dense output layer $\logit(\bx) = h(\bx)^\top \bbeta$, where the model confidence (i.e., maximum predictive probability) is characterized by the magnitude of the class logits, which is defined by the inner product distances between the hidden representation $h(\bx)$ and the decision boundaries $\{\bbeta_k\}_{k=1}^K$ (see, e.g., Figure \ref{fig:sngp_overview}). 
As a result of this formulation, the model computes confidence for a point $\bx^*$ based not on its distance from the training data $\Xsc_{\texttt{IND}}$, but based on its distance from the decision boundaries, i.e., the model uncertainty is not \textit{distance aware}. {\color{black} \Cref{sec:discussion} provides further discussion.} 

\textbf{Two Conditions for Distance Awareness in Deep Learning}. 
The output of deep learning classifiers can be written as $\logit(\bx) = g \circ h(\bx)$,  which is composed of a hidden mapping $h: \Xsc \rightarrow \Hsc$  that map the input $\bx$ into a hidden representation space $h(\bx) \in \Hsc$, and an output layer $g$ that maps $h(\bx)$ to the label space. As shown in the previous section, this formulation is not traditionally \textit{input distance aware}, but it can be made to be so by imposing two conditions:
\textbf{(1)} make the output layer $g$
\textit{distance aware}, so it outputs 
an uncertainty metric reflecting distance in the hidden space $||h(\bx) - h(\bx')||_H$ (in practice, this can be achieved by using a \gls{GP} with a shift-invariant kernel as the output layer),
and \textbf{(2)} make the hidden mapping \textit{distance preserving} (defined below), so that the distance in the hidden space $||h(\bx) - h(\bx')||_H$ has a meaningful correspondence to the distance $d_X(\bx, \bx')$ in the data manifold. 
From a mathematical point of view, this is equivalent to requiring $h$ to satisfy the  \textit{bi-Lipschitz} condition \citep{searcod_metric_2006}:
\begin{align}
    L_1 \times d_X(\bx_1, \bx_2) \leq ||h(\bx_1) - h(\bx_2)||_H \leq L_2 \times d_X(\bx_1, \bx_2),
    \label{eq:dp}
\end{align}
for positive and bounded constants $0 < L_1 <  L_2$. For a deep learning model, the bi-Lipschitz condition (\ref{eq:dp}) usually leads the model's hidden space to preserve a {\color{black} meaningful distance in the input data manifold $\Xsc$ that is, e.g., effective for determining if an observation is in-distribution.}
This is due to the fact that the upper Lipschitz bound $||h(\bx_1) - h(\bx_2)||_H \leq L_2 \times d_X(\bx_1, \bx_2)$ is an important condition for the adversarial robustness of a deep network, which prevents the hidden representations $h(\bx)$ from being overly sensitive to the meaningless perturbations in the pixel space (e.g., Gaussian noise) \citep{ruan_reachability_2018, weng_evaluating_2018, tsuzuku_lipschitz-margin_2018, jacobsen_excessive_2018, sokolic_robust_2017}. {\color{black} On the other hand, 
the lower Lipschitz bound $||h(\bx_1) - h(\bx_2)||_H \geq L_1 \times d_X(\bx_1, \bx_2)$ prevents the hidden representation from collapsing the representations of distinct examples together, which otherwise leads to undesired invariance to the meaningful differences between examples (i.e., the concept of \textit{feature collapse})}
\citep{jacobsen_exploiting_2019, van2020uncertainty
}. When we combine these two inequalities and the properties they enforce, the bi-Lipschitz condition essentially encourages $h$ to be an \textit{approximately} isometric mapping, 
thereby ensuring that the learned representation space $H$ has a robust and meaningful correspondence with 
the geometry of the input data manifold $\Xsc$. Heuristically as well, machine learning methods usually tend to attempt to learn an approximately isometric and geometry-preserving mapping \citep{hauser_principles_2017, perrault-joncas_metric_2017, rousseau_residual_2020}. For example, deep image classifiers usually strive to learn a mapping from the image manifold to a hidden representation space where the input data is easily separable using a set of linear decision boundaries. Similarly, sentence encoders aim to project sentences into a vector space where the cosine distance can be used to measure the \textit{semantic textual similarity} between natural language sentences \citep{cer2018universal}. It has also been shown that preserving such \textit{approximate} isometry in a neural network is possible even after significant dimensionality reduction \citep{blum_random_2006}.


\section{Our Proposed Method: Spectral-normalized Neural Gaussian Process (SNGP)}
\label{sec:method}

In this section we formalize the definition of the \textit{\glsfirst{SNGP}} algorithm for modern residual-based \gls{DNN} (e.g., ResNet, Transformer). The \gls{SNGP} algorithm provides a simple approach to encode distance awareness into the output layer, and distance preservation into the hidden layers (i.e., the two properties introduced in Section \ref{sec:da}). We summarize the method in Algorithms \ref{alg:training}-\ref{alg:prediction}.


\vspace{-0.5em}
\subsection{Distance-aware Output Layer via Laplace-approximated Neural Gaussian Process}
\label{sec:gp}

\gls{SNGP} achieves distance-awareness in the output layer $g: \Hsc \rightarrow \Ysc$ by replacing a dense output layer with an approximate \glsfirst{GP} conditioned on the learned hidden representations of the \gls{DNN}, 
where the posterior variance of a test input $\bx^*$ is proportional to its $L_2$ distance from the training datapoints in the hidden space\footnote{In this work, we focus on the RBF kernel for its simplicity, however it is easy to extend our framework to other kernels (e.g., Mat\'{e}rn, MLP, or arccos kernel, etc) by modifying the activation function and the distribution of the random-feature mapping in Equation (\ref{eq:rff_gp}) \citep{choromanski_geometry_2018, liu2020random}.}. 

Given a training dataset  $\Dsc=\{y_i, \bx_i\}_{i=1}^N$ with $N$ data points, we denote the hidden representation of the \gls{DNN} defined for each training point as $h_i=h(\bx_i)$, and denote the Gaussian Process conditioned on the hidden representation as $\bg_{N \times 1} = [g(h_1), \dots, g(h_N)]^\top$. Then, the prior distribution of a \gls{GP} model equipped with an \gls{RBF} kernel is a multivariate normal:
\begin{align}
    \bg_{N \times 1}  \sim N(\bzero_{N \times 1}, \sigma^2 * \bK_{N \times N}), 
    \quad \mbox{where} \  
    \bK_{i, j} = \exp(-||h_i - h_j||_2^2/2),
    \label{eq:gp_prior}
\end{align}
where $\sigma^2$ is the kernel amplitude \citep{rasmussen_gaussian_2006}.  The model's posterior distribution is then calculated using Bayes Rule:
\begin{align}
    p(g|\Dsc) &\propto p(\Dsc|g) \times p(g),
    \label{eq:gp_posterior}
\end{align}
where  $p(g)$ is the GP prior as given by (\ref{eq:gp_prior}). $p(\Dsc|g)$ is the data likelihood for the task, e.g., for a regression task, it can be the exponentiated squared loss $p(\Dsc|g)= \exp(\sum_{i=1}^N(y_i - g_i)^2)$. For a binary classification task, it can be the exponentiated sigmoid cross-entropy loss $p(\Dsc|g)=\exp \big(\sum_{i=1}^N I(y_{i}=1) \, \log p_{i} + I(y_{i}=0) \, \log (1-p_{i}) \big)$ where $p_i = \mathrm{sigmoid}(g_i)$.
 
However, in the context of large-scale neural modeling, performing exact \gls{GP} inference (\ref{eq:gp_prior})-(\ref{eq:gp_posterior}) is difficult for two reasons: 
(i) inference with the exact prior (\ref{eq:gp_prior}) is \textit{computationally intractable}, due to the need of inverting a $N \times N$ kernel matrix $\bK$ which has a cubic time complexity $\mathcal{O}(N^3)$.
(ii) Inference with a non-Gaussian likelihood in (\ref{eq:gp_posterior}) is \textit{analytically intractable}, due to the difficulty of deriving the analytical integration over a non-conjugate likelihood. The existing literature commonly handles (i)-(ii) by deploying sophisticated inference algorithms such as structured \gls{VI} or \gls{MCMC} (see Section \ref{sec:related_app} for a full review). However, this invariably imposes nontrivial engineering and computational complexity to an existing deep learning system, rendering itself infeasible in many practical scenarios where the system scalability and maintainability is of high importance.

In this work, we propose to tackle the above challenge by performing Laplace-approximation inference to the \glspl{RFF} expansion of the \gls{GP} model \citep{rasmussen_gaussian_2006}, by (i) converting the \gls{GP} model  (\ref{eq:gp_prior}) into a featurized Bayesian linear model via the random-feature expansion \citep{rahimi_random_2008}. Then (ii) approximate the intractable posterior (\ref{eq:gp_posterior}) using Laplace approximation \citep{gelman_bayesian_2013}. Both are well-established methods in the statistical machine learning literature with rigorous theoretical guarantees \citep{rahimi2008uniform, degroot2012probability}. We re-iterate that the goal here is 
\emph{not} to develop another algorithm to approximate the exact \gls{GP} posterior, but to identify a simple, practical baseline approach to implement the distance-awareness property (Definition \ref{def:da}) into a neural model with minimal modification to the training pipeline for a deterministic \gls{DNN}. 
To this end, as we will show, the proposed approach leads to a closed-form posterior that can be trained end-to-end using the same pipeline for a deterministic \gls{DNN} and with comparable time complexity.
Empirically, the trained model maintains the generalization performance of a deterministic \gls{DNN} while illustrating improved uncertainty performance in calibration and in \gls{OOD} detection.


\paragraph{Random-feature Expansion of \gls{GP} Model.} First, we approximate the prior defined in (\ref{eq:gp_prior}) by defining a low-rank approximation of the kernel matrix $\bK=\bPhi\bPhi^\top$ by using random features \citep{rahimi_random_2008}, which gives us a \textit{random-feature Gaussian process}:
\begin{align}
    \bg_{N \times 1} \sim MVN(\bzero_{N \times 1}, \bPhi\bPhi^\top_{N \times N}), 
   \quad \mbox{where} \quad 
    \bPhi_{i, D_L \times 1} = 
    \sqrt{2 \sigma^2 /D_L} * 
    \cos(-\bW_L h_i + \bb_L).
    \label{eq:rff_gp}
\end{align}
Here, $h_i=h(\bx_i)$ represents the hidden representations of the penultimate layer with a dimensionality of $D_{L-1}$. $\bPhi_i = \phi(\bx_i)$ is the final layer with dimension $D_L$, which is represented by (1) a fixed weight matrix $\bW_{L, D_L \times D_{L-1}}$ whose entries are sampled i.i.d from $N(0, 1)$, and (2) a fixed bias term $\bb_{L, D_L \times 1}$ whose entries are sampled i.i.d from $Uniform(0, 2\pi)$. We can therefore write the logits using the \gls{RFF} expansion to the \gls{GP} prior in (\ref{eq:gp_prior}) as a neural network layer consisting of fixed hidden weights $\bW$ and learnable output weights $\bbeta$:
\begin{align}
    g(h_i) &= \sqrt{2 \sigma^2 /D_L} * \cos(-\bW_L h_i + \bb_L)^\top \bbeta, 
    \quad \mbox{with prior} \quad 
    &\bbeta_{D_L \times 1} \sim N(0, \tau * \bI_{D_L \times D_L}).
    \label{eq:rff_lr}
\end{align}

\paragraph{Laplace Approximation for Posterior Uncertainty.} Notice that conditional on $h$, the output weights $\beta$ are the only learnable parameters in the model. As a result, the \glsfirst{RFF} approximation in  (\ref{eq:rff_lr}) reduces an infinite-dimensional \gls{GP} to a standard Bayesian linear model, for which many posterior approximation methods (e.g., \gls{EP}) can be applied \citep{minka_family_2001}. In this work, we choose the Laplace method due to its simplicity and the fact that its posterior variance has a convenient closed form \citep{rasmussen_gaussian_2006}.
Briefly, the Laplace method approximates the model posterior $p(\beta|\Dsc)$ using a Gaussian distribution that is centered around the \gls{MAP} estimate $\hat{\beta} = \mbox{argmax}_{\beta}\,  p(\beta|\Dsc)$, such that 
\begin{align}
p(\beta | \Dsc) \approx MVN(\hat{\beta}, \hat{\bSigma}=\widehat{\bH}^{-1}), 
\quad \mbox{where} \quad 
\widehat{\bH}_{(i,j)} = -\frac{\partial^2}{\partial \beta_i \partial \beta_j} \log \, p(\beta|\Dsc)|_{\beta=\hat{\beta}}
\label{eq:laplace}
\end{align}
is the $D_L \times D_L$ Hessian matrix of the log posterior likelihood evaluated at the \gls{MAP} estimates. For a binary classification task, the posterior precision matrix (i.e., the inverse covariance matrix) adopts a simple expression $\hat{\bSigma}^{-1}_k = \bI + \sum_{i=1}^N \hat{p}_{i}(1-\hat{p}_{i}) \Phi_i \Phi_i^\top$, 
where $p_{i}$ is the model prediction $p_i = sigmoid(\hat{g}_i)$ under the \gls{MAP} estimates $\hat{g}_i = \Phi_i^\top\hat{\beta}$ \citep{rasmussen_gaussian_2006}. We introduce the extensions to regression and multi-class classification in Appendix \ref{sec:multiclass_app}. 

To summarize, for a classification task, the Laplace posterior for an approximate \gls{GP} under the \gls{RFF} expansion is:
\begin{align}
    \beta | \Dsc \sim 
    MVN(\hat{\beta}, \hat{\bSigma}), 
    \quad \mbox{where} \quad 
    \hat{\bSigma}^{-1} = \tau * \bI + \sum_{i=1}^N \hat{p}_{i}(1-\hat{p}_{i}) \Phi_i \Phi_i^\top,
\label{eq:gp_posterior_approx}
\end{align}
where $\hat{\bbeta}$ is the model's \gls{MAP} estimate conditioned on the \gls{RFF} hidden representation $\Phi$ in (\ref{eq:rff_gp}), and $\tau$ is the prior variance. To obtain the posterior distribution (\ref{eq:gp_posterior_approx}), one only need to first obtain the \gls{MAP} estimate by training the entire network with respect to the \gls{MAP} objective using \gls{SGD}:
\begin{align}
-\log p(\beta, \{\bW_l, \bb_l\}_{l=1}^{L-1}|\Dsc) = -\log p(\Dsc|\beta, \{\bW_l, \bb_l\}_{l=1}^{L-1}) + \frac{1}{2 \tau}||\bbeta||^2,   
\label{eq:gp_objective}
\end{align}
where $\{\bW_l, \bb_l\}_{l=1}^{L-1}$ are the hidden weights of the network and $-\log p(\Dsc|\beta, \{\bW_l, \bb_l\}_{l=1}^{L-1})$ is the negative log likelihood for the task (e.g., the cross entropy loss for a classification task). 
Then, in the final epoch, we can compute this covariance in an incremental fashion by initializing covariance with $I$ and accumulating covariance contributions from each batch as in (\ref{eq:gp_posterior_approx})
\footnote{i.e., to update the posterior precision matrix as $\hat{\bSigma}^{-1}_{t}=\hat{\bSigma}^{-1}_{t-1} + \sum_{i=1}^M \hat{p}_{i}(1-\hat{p}_{i})\Phi_i \Phi_i^\top$ for minibatches of size $M$. Notice that to obtain a MAP estimate $\hat{\beta}$ that is strictly conditioned on the hidden representations $\Phi$, one should freeze the hidden representations in the final epoch and only update the output weights $\bbeta$. However in practice, we find this has no significant impact on the final performance. This is most likely due to the fact that the hidden representation has already become stabilized in the final epochs.} 
\footnote{In the online learning setting where the model is processing a infinite stream of data, this expression can be modified to $\hat{\bSigma}^{-1}_{t}=(1-m)*\hat{\bSigma}^{-1}_{t-1} + m*\sum_{i=1}^M \hat{p}_{i}(1-\hat{p}_{i})\Phi_i \Phi_i^\top$ where $m$ is a small scaling coefficient. This is analogous to the exponential moving average estimator for batch variance as used in batch normalization \citep{ioffe2015batch}}. 
As a result, the approximate \gls{GP} posterior (\ref{eq:gp_posterior_approx}) can be learned scalably and in closed-form with minimal modification to the training pipeline of a deterministic \gls{DNN}.
It is worth noting that under the \gls{RFF} posterior, the Laplace approximation is in fact asymptotically exact by the virtue of the \gls{BvM} theorem and the fact that (\ref{eq:rff_lr}) is a finite-rank model \citep{freedman_wald_1999, lecam_convergence_1973, panov_finite_2015, dehaene_deterministic_2019}.

\subsection{Approximately Distance-preserving Hidden Mapping via Spectral Normalization}
\label{sec:sn}

Replacing the output layer $g$ with an approximate Gaussian process only allows the model $\logit(\bx) = g \circ h(\bx)$ to be aware of the distance in the hidden space $||h(\bx_1) - h(\bx_2)||_H$.
It is also important to ensure the hidden mapping $h$ is \textit{approximately distance preserving} so that the distance in the hidden space $||h(\bx) - h(\bx')||_H$ has a meaningful correspondence to the distance in the data manifold $d_X(\bx, \bx')$. To this end, we notice that modern deep learning models (e.g., ResNets, Transformers) are commonly composed of residual blocks, i.e., $h(\bx)=h_{L-1} \circ \dots  \circ h_2  \circ h_1(\bx)$ where $h_l(\bx)= \bx + g_l(\bx)$. For such models, there exists a simple method to ensure  $h$ is \textit{distance preserving}: by bounding the Lipschitz constants of all residual mappings $\{g_l\}_{l=1}^{L-1}$ in the residual blocks $h_l(\bx) = \bx + g_l(\bx)$.
Specifically, it can be shown that, under suitable regularity conditions, a \textit{strict} distance preservation can be guaranteed by restricting the Lipschitz constants to be less than 1.
We state this result formally below: 
\begin{proposition}[Lipschitz-bounded residual block is distance preserving \citep{bartlett_representing_2018}]
Consider a hidden mapping $h: \Xsc \rightarrow \Hsc$ with residual architecture $h=h_{L-1} \circ \dots h_2 \circ h_1$ where $h_l(\bx) = \bx + g_l(\bx)$ and $\Xsc$ and $\Hsc$ are of equal dimension. If for  $0 < \alpha \leq 1$, all $g_l$'s are $\alpha$-Lipschitz, i.e., $||g_l(\bx) - g_l(\bx')||_H \leq \alpha d_X(\bx, \bx') \quad \forall (\bx, \bx') \in \Xsc$. Then:
$$L_1 \times d_X(\bx, \bx') \leq || h(\bx) - h(\bx') ||_H \leq L_2 \times d_X(\bx, \bx'), $$
where $L_1 = (1-\alpha)^{L-1}$ and $L_2=(1+\alpha)^{L-1}$, i.e., $h$ is \textit{distance preserving}.
\label{thm:resnet_lipschitz}
\end{proposition}
Proof is in Appendix \ref{sec:resnet_lipschitz_proof}. The ability of a residual network to construct a geometry-preserving metric transform between the input space $\Xsc$ and the hidden space $\Hsc$ is well-established in learning theory and generative modeling literature, but the application of these results in the context of uncertainty estimation for \gls{DNN} appears to be new  \citep{bartlett_representing_2018, behrmann_invertible_2019, hauser_principles_2017,  rousseau_residual_2020}. 

However in practice, a \textit{strict} preservation of distance is both impossible and likely undesirable. The reason is that per common neural network practice, the model often projects a high-dimensional example into a lower-dimensional representation (i.e., $|\Hsc| < |\Xsc|$ for the hidden mapping $h:\Xsc \rightarrow \Hsc$). This necessitates information loss and precludes the possibility of exact isometry (i.e., invertibility) \citep{smith2021can}. Furthermore, an overly strict bound on the Lipschitz constant of $g_l$ can push the model toward identity mapping, greatly restricting the expressiveness of the network and leading to suboptimal generalization \citep{behrmann_invertible_2019}. 
To this end, a more prudent strategy would be to pursue \textit{approximate} distance preservation, i.e., by finetuning the magnitude of Lipschitz constant to a more relaxed extent, so as to balance the tradeoff between the expressiveness of the network and its distance-preservation ability (see Section \ref{sec:discussion} for further discussion). 

Algorithm-wise, to ensure the hidden mapping $h$ is approximately distance preserving, it is sufficient to ensure that the weight matrices for the nonlinear residual block $g_l(\bx)=a(\bW_l\bx + \bb_l)$ to have spectral norm (i.e., the largest singular value) to be upper-bounded, since $||g_l||_{Lip} \leq ||\bW_l\bx + \bb_l||_{Lip} \leq ||\bW_l||_2$. In this work, we enforce the aforementioned Lipschitz constraint on $g_l$'s by applying the \textit{ \gls{SN}} on the weight matrices $\{\bW_l\}_{l=1}^{L-1}$ as recommended in \cite{behrmann_invertible_2019}. Briefly, at every training step, the \gls{SN} method first estimates the spectral norm $\hat{\lambda} \approx ||\bW_l||_2$ using the power iteration method \citep{gouk_regularisation_2018, miyato_spectral_2018}, and then normalizes the weights as:\\
\scalebox{0.9}{
\parbox{\linewidth}{%
\begin{align}
    \bW_l = 
    \begin{cases} 
    c * \bW_l / \hat{\lambda}  & 
    \mbox{if } c < \hat{\lambda}\\ 
    \bW_l & \mbox{otherwise}
    \end{cases}
    \label{eq:spec_norm}
\end{align}
}
}\\
where $c>0$ is a hyperparameter used to adjust the exact spectral norm upper bound on $||\bW_l||_2$ (so that $||\bW_l||_2 \leq c$). 
Therefore, (\ref{eq:spec_norm}) allows us more flexibility in controlling the spectral norm of the neural network weights so it is the most compatible with the architecture at hand.
{\color{black} In this work, we treat $c$ as a tunable hyperparameter which can be selected based on the validation data (\Cref{sec:hyper_app}).}

\begin{center}
\scalebox{.85}{%
\begin{minipage}{0.5\textwidth}
\begin{algorithm}[H]
   \caption{\gls{SNGP} Training}
   \label{alg:training}
\begin{algorithmic}[1]
   \STATE {\bfseries Input:} \\
   Minibatches $\{D_i\}_{i=1}^N$ for $D_i=\{y_m, \bx_m\}_{m=1}^M$. 
   \vspace{0.2em}
   \STATE {\bfseries Initialize:} \\
    \vspace{-1.2em}
    $$\hat{\bSigma}=\tau * \bI, \bW_L \stackrel{iid}{\sim} N(0, 1), \bb_L \stackrel{iid}{\sim} U(0, 2\pi)$$.\\
     \vspace{-1.5em}
   \FOR{$\mathsf{train\_step}=1$ {\bfseries to} $\mathsf{max\_step}$}
   \STATE 
   \gls{SGD} update $\Big\{ \bbeta, \{\bW_l\}_{l=1}^{L-1}, \{\bb_l\}_{l=1}^{L-1} \Big\}$ (\ref{eq:gp_objective})
   \IF{$\mathsf{final\_epoch}$}
   \STATE 
   Update precision matrix $\hat{\bSigma}^{-1}$ 
   (\ref{eq:gp_posterior_approx}).
   \ENDIF
   \ENDFOR
   \STATE Compute posterior covariance $\hat{\bSigma}=\mathrm{inv}(\hat{\bSigma}^{-1})$.
\end{algorithmic}
\end{algorithm}
\end{minipage}

\hspace{1em}

\begin{minipage}{0.5\textwidth}
\begin{algorithm}[H]
   \caption{\gls{SNGP} Prediction}
   \label{alg:prediction}
\begin{algorithmic}[1]
   \STATE {\bfseries Input:} Testing example $\bx$.
   \STATE 
   Compute Features: 
   \vspace{-0.5em}
   $$\bPhi_{D_L \times 1} = \sqrt{2 \sigma^2 /D_L} * \cos(\bW_L h(\bx) + \bb_L)$$ 
   \vspace{-1.75em}
   \STATE 
   Compute Posterior Mean: 
   \vspace{-0.5em}
   $$\logit(\bx)=\Phi^\top\bbeta$$ 
   \vspace{-1.75em}
   \STATE 
   Compute Posterior Variance: 
   \vspace{-0.5em}
   $$\var(\bx)=\Phi^\top\hat{\Sigma} \Phi.$$ 
   \vspace{-1.75em}
      \STATE 
      Compute Predictive posterior distribution:
    \vspace{-0.5em}
   $$p(y|\bx)=\int_{g \sim N(\logit(\bx), \var(\bx))}{\sigmoid(g)} dg$$ 
   \vspace{-1em}
\end{algorithmic}
\end{algorithm}
\end{minipage}
}
\end{center}

\vspace{1em}

\paragraph{Method Summary $\&$ Practical Implementation.}
We summarize the method in Algorithms  \ref{alg:training}-\ref{alg:prediction}. 
As shown, during training, the model updates the hidden-layer weights $\{\bW_l, \bb_l\}_{l=1}^{L-1}$ and the trainable output weights $\beta$ via minibatch \gls{SGD} (i.e., exactly the same way as a deterministic DNN). Then, in the final epoch, it also performs an update of the precision matrix using Equation (\ref{eq:gp_posterior_approx}).
During inference, the model first performs the conventional forward pass to compute the final hidden features $\phi(\bx)_{D_L \times 1}$, and then compute the posterior mean  $m(\bx)=\phi(\bx)^\top\bbeta$ (time complexity $\mathcal{O}(D_L)$) and the predictive variance $v(\bx)^2=\phi(\bx)^\top\hat{\Sigma}\phi(\bx)$ (time complexity $\mathcal{O}(D^2_L)$)\footnote{
To extend Algorithm 1-2 to regression or multi-class classification task, one just need to use the appropriate precision matrix update as introduced in \ref{sec:multiclass_app}, and replace the sigmoid output activation to identity or softmax.}. To see how this predictive distribution of \gls{SNGP} reflects the distance-awareness property, notice that conditional on the penultimate embedding $h$ and assuming squared loss, the predictive logit of the \gls{SNGP} follows a Gaussian process:
\begin{align}
\label{eq:sngp_summary_primal}
    g(\bx) &\sim N \big(\logit(\bx), \var(\bx) \big) \quad \mbox{where}\\
    \logit(\bx) &= \phi(\bx)^\top (\bPhi^\top \bPhi + \tau \, \bI )^{-1} \bPhi^\top \by, \nonumber\\
    \var(\bx) &= \phi(\bx)^\top (\bPhi^\top \bPhi + \tau \, \bI )^{-1} \phi(\bx), \nonumber
\end{align}
where $\bPhi$ is the $N \times D_L$ random feature embedding of the training data. Then, straightforward application of Woodbury matrix identity reveals that (\ref{eq:sngp_summary_primal}) can equivalently be expressed in the familiar dual form of the Gaussian process \citep{rasmussen_gaussian_2006}:
\begin{align}
    \logit(\bx) &= \bk^{*}(\bx)^\top (\bK + \tau \, \bI )^{-1} \by, \\
    \var(\bx) &= \tau^{-1} * \big[ 
    k(\bx, \bx) - \bk^{*}(\bx)^\top (\bK + \tau \, \bI )^{-1} \bk^*(\bx) \big],
    \label{eq:sngp_summary_dual}
\end{align}
where $k(\bx, \bx)_{1 \times 1} = \phi(\bx)^\top\phi(\bx)$, $\bk^*(\bx)_{N \times 1} = \phi(\bx)^\top\bPhi^\top$ and $\bK_{N \times N} = \bPhi \bPhi^\top$ are kernel matrices approximating those under the \gls{RBF} kernel $k(\bx, \bx') \propto \exp(-||h(\bx) - h(\bx')||_2^2)$. Consequently, under a distance-preserving mapping $h$, for a test example $\bx$ that is moving away from the training-data manifold, its predictive kernel matrix $\bk^*(\bx) = [k(\bx, \bx_i)]_{i=1}^N$ systematically approaches 0, while the testing-data kernel $k(\bx, \bx)$ and the training-data kernel $\bK$ remain in a constant range. This causes the predictive mean $m(\bx)$ to approach zero, and the predictive variance $v(\bx)$ to approach its maximum
\footnote{This conclusion also holds for non-Gaussian outcome. Where the predictive mean and variance can be expressed as $m(\bx)=\bk^{*}(\bx)^\top (\bK + \tau \, \bI )^{-1} \hat{\by}$ and $v(\bx) = \tau^{-1} k(\bx, \bx) - \bk^{*}(\bx)^\top \hat{\bV} \bk^*(\bx)$, where $\hat{\by}$ is the model prediction on the training data, and $\hat{\bV}$ is the inverse covariance matrix calculated using the Laplace method (i.e., via Equation (\ref{eq:gp_posterior_approx})). (\cite{rasmussen_gaussian_2006}, Chapter 3.6)}. As a result, the SNGP model achieves the behavior as motivated in (\ref{eq:minimax_solution}), i.e., generating a maximum-entropy predictive distribution for OOD inputs that are far outside the training domain. {\color{black}We include a proof of this behavior in Appendix \ref{sec:optimal_sngp}.}

To estimate the predictive distribution $p(y|\bx)=\int_{g \sim N(\logit(\bx), \var(\bx))}{\softmax(g)} dg$, we can use either of two approaches: Monte Carlo averaging or mean-field approximation. Specifically, Monte Carlo averaging randomly generates $K$ i.i.d samples from the distribution $N(\logit(\bx), \var(\bx))$, compute the softmax, and take the average over the $K$ samples, 
$p(y|\bx) = \frac{1}{K} \sum_{k=1}^K \softmax\big(g_k\big)$. Note that this Monte Carlo approximation is applied just in last layer, so it is very fast and does not require multiple forward passes of the network. 
However, in ML systems where Monte Carlo sampling is difficult or considered too expensive (e.g., the on-device settings), we can alternatively consider mean-field approximation to the softmax Gaussian posterior  \citep{daunizeau_semi-analytical_2017, lu2020mean}: 
\begin{align}
    p(y|\bx) = \softmax\big(\frac{\logit(\bx)}{\sqrt{1 + \lambda * \var(\bx)}}\big),
    \label{eq:mean_field_approx}
\end{align}
where $\lambda$ is a scaling factor that is commonly set to $\pi/8$ \citep{lu2020mean}. 
In our experiments, we have found the two approaches to perform similarly. Therefore we report the results from  mean-field approximation in our experiments due to its wider applicability.

Interestingly, Equation (\ref{eq:mean_field_approx}) highlights the role of the kernel amplitude $\sigma$ in improving a classifier's calibration performance: as the predictive variance $var(\bx)$ is proportional to $\sigma$ (see Equation (\ref{eq:sngp_summary_dual})), it effectively re-scales the magnitude of the  logits and functions similarly to the temperature parameter in the temperature scaling technique \citep{guo_calibration_2017}. We also note that in our experiment, the value of $\sigma$ generally does not have a significant impact on model performance in OOD detection. This is likely due to the fact that scaling the model uncertainty by a constant $\sigma$ does not change the ranking of the uncertainty scores. 
In practice, we recommend estimating $\sigma$ on a small amount of in-distribution validation data by minimizing with respect to a proper scoring rule. This is because an overparameterized model such as a \gls{DNN} is known to overfit the training data, rendering the in-sample estimate of the second-order statistics not reliable \citep{wahba_spline_1990}. In all experiments, we estimate $\sigma$ by minimizing it against the logarithm score (i.e., the marginalized negative log likelihood \citep{gneiting_probabilistic_2007}) on a small held-out validation dataset. We summarize all other model hyperparameters (e.g., the spectral norm upper bound $c$) and provide practical recommendations in Appendix \ref{sec:hyper_app}.




\section{Where does SNGP fit in the current landscape of methods?}
\label{sec:method_landscape}

\begin{figure}[!ht]
    \centering
    \hspace{-5em}
    \includegraphics[width=0.8\columnwidth]{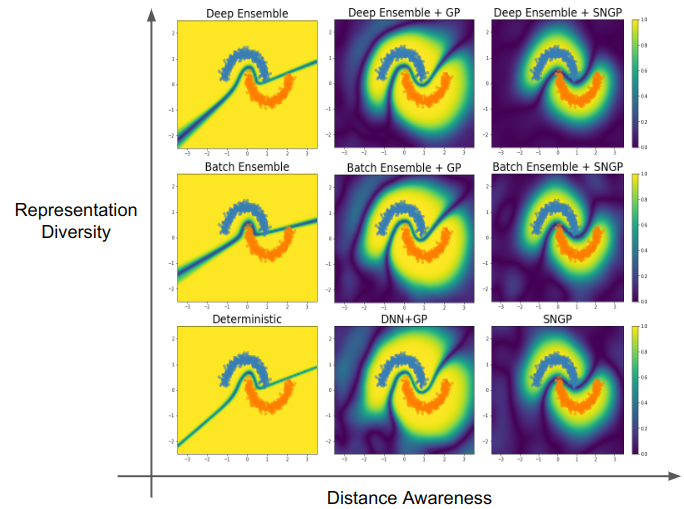}
    \caption{We highlight complementary ways of improving uncertainty of a vanilla DNN. \textbf{The x-axis highlights improvements to single model uncertainty  $p_{\theta}(y|x)$ via distance-awareness.} Note that in this setting, we're focused on improving quality of uncertainty for a fixed deterministic representation. \textbf{The y-axis shows an orthogonal direction of improvement  by combining multiple models $\theta_1, \theta_2,\ldots,\theta_M$.} Note that ensembles work primarily by leveraging    multiple representations (as opposed to the deterministic uncertainty quantification setting). This reveals some interesting differences: for instance, the left most column which uses vanilla DNN as the base model to construct deep ensembles (top-left) and efficient ensembles (top-middle), shows that ensembling strategies provide qualitatively different benefits as they primarily add uncertainty near the boundary and do not necessarily change the confidence landscape far away from the data. On the other hand, the bottom row shows that methods such as DNN-GP (bottom-middle) and SNGP (bottom-right) improve distance awareness in the classification boundary. \textbf{Finally, these two axes are not really competing approaches; they actually provide complementary benefits} and can be combined to further improve performance. See Section~\ref{sec:building_block_ensemble}. 
    }
    \label{fig:ensemble_vs_basemodel}
\end{figure}

We have introduced SNGP as a method to improve single model uncertainty. Before describing the experimental setup, we first explain where SNGP fits in the current landscape of uncertainty methods in deep learning. 

SNGP is a method for \textit{deterministic} uncertainty quantification, that is, given a single, non-random representation, SNGP improves the base model by enhancing its distance-awareness property. This is a orthogonal direction to what was taken by popular ensemble-based approaches (e.g., Monte-Carlo dropout, BatchEnsemble \citep{wen_batchensemble_2020} or Deep Ensemble), who improve performance primarily through integrating over multiple diverse representations \citep{fort_deep_2019}. 
However, when base models  consistently make overconfident predictions far away from the data, their ensemble can inherit such behavior as well. As a result, while ensembling vanilla DNNs are effective in improving accuracy and calibration under shift, they can be lacking in providing as significant a boost in OOD detection (e.g., due to the lack of distance awareness). 

To see this point visually, in Figure~\ref{fig:ensemble_vs_basemodel} where we highlight the two orthogonal axes of improvement.
The y-axis denotes better ensembling strategies: the left column shows that ensembling vanilla DNNs does not necessarily change the confidence landscape far away from the inputs. Even though efficient ensembles reduce compute and memory, they are still solving a different problem from SNGP by improving representation diversity but in an efficient manner. The x-axis denotes better single model uncertainty; the bottom row shows how DNN-GP and SNGP improve deterministic uncertainty quantification by improving distance preservation and distance awareness. 

Finally, we note that these two axes provide complementary benefits, so they can be combined to further improve performance (see SNGP ensemble in top right).  Since SNGP primarily improves single model uncertainty, we focus our comparisons with other methods for deterministic uncertainty quantification in the experiments (Section \ref{sec:experiments}). 
Whenever possible, we recommend to ensemble SNGP models build toward a strong probabilistic DNN model that simultaneously quantifies representation uncertainty and ensures distance awareness.  We explore ensembles of SNGP in Section~\ref{sec:building_block_ensemble}. As an aside, data augmentation is another orthogonal axis of improvement \citep{hendrycks_augmix_2020}. We also explore SNGP as a building block for data augmentation methods in Section~\ref{sec:building_block_data_aug}.



\section{Related Work} 
\label{sec:related_app}

\textbf{Single-model approaches to deep classifier uncertainty.}
Recent work examines uncertainty methods that add few additional parameters or runtime cost to the base model. The state-of-the-art on large-scale tasks are efficient ensemble methods \citep{wen_batchensemble_2020,dusenberry_efficient_2020}, which cast a set of models under a single one, encouraging independent member predictions using low-rank perturbations.  These methods are parameter-efficient but still require multiple forward passes from the model. SNGP investigates an orthogonal approach that improves the  uncertainty quantification by imposing suitable regularization on a single model, and therefore requires only a single forward pass during inference.
There exists other runtime-efficient, single-model approaches to estimate predictive uncertainty, achieved by either replacing the loss function \citep{hein_why_2019, malinin_predictive_2018, malinin_prior_2018, sensoy_evidential_2018, shu_doc_2017}, the output layer \citep{bendale_towards_2016, tagasovska_single-model_2019, calandra_manifold_2016, macedo_isotropic_2020, macedo2021enhanced, padhy2020revisiting}, computing a closed-form posterior for the output layer \citep{riquelme_deep_2018, snoek_scalable_2015, kristiadi_being_2020} {\color{black}, or predicting a-priori uncertainty away from training data \citep{skafte2019reliable}.} \gls{SNGP} builds on these approaches by also considering the intermediate representations which are necessary for good uncertainty estimation, and proposes a simple method (spectral normalization) to achieve it. A recent method named Deterministic Uncertainty Quantification (\textbf{DUQ}) also regulates the neural network mapping but uses a two-sided gradient penalty \citep{van2020uncertainty}. 
However, empirically, the  
two-sided gradient penalty can lead to unstable training dynamics for a deep residual network, 
and is observed to over-constrain the model capacity in more difficulty tasks (e.g., CIFAR-100) in our experiments (Section \ref{sec:cifar}). 
Following the initial conference publication of this work \citep{liu2020simple} and in a similar vein, some later work also investigated building other class of probabilistic models on top of a spectral-regularized network, e.g., variational Gaussian process, Gaussian mixture model, or stochastic differential equations \citep{mukhoti2021deterministic, van2021feature, cui2021accurate}. While obtaining encouraging results on small-scale benchmarks (e.g., CIFAR), these approaches are still computationally prohibitive for large-scale tasks with high number of output classes (e.g., ImageNet) which SNGP can easily scale to (Section \ref{sec:imagenet}).

\textbf{Laplace approximation and GP inference with \gls{DNN}.} Laplace approximation has a long history in \gls{GP} and (Bayesian) NN literature \citep{tierney_approximate_1989, denker_transforming_1991, rasmussen_gaussian_2006, mackay_practical_1992, ritter_scalable_2018, hobbhahn2020fast,  DBLP:journals/corr/abs-2106-10065, pmlr-v161-kristiadi21a, DBLP:journals/corr/abs-2111-03577, daxberger2021laplace}, and the theoretical connection between a Laplace-approximated \gls{DNN} and \gls{GP} has being explored recently \citep{khan_approximate_2019}. Differing from these works, \gls{SNGP} applies the Laplace approximation to the posterior of a neural GP, rather than to a shallow GP or a dense-output-layer \gls{DNN}. Earlier works that combine a \gls{GP} with a \gls{DNN} learns the hidden representation separately \citet{salakhutdinov07}, or performs end-to-end learning via MAP estimation  \citep{calandra_manifold_2016} or structured \gls{VI} \citep{hensman2015scalable, bradshaw_adversarial_2017, wilson_stochastic_2016}. These approaches were shown to lead to poor calibration by recent work \citep{tran_calibrating_2019}, which proposed a simple fix by combing \gls{MCD} with random Fourier features, {\color{black} which we term MC Dropout Gaurssian Process (\textbf{MCD-GP})\glsunset{MCDGP}.
\gls{SNGP} differs from \gls{MCDGP} in that it considered a different regularization approach (spectral normalization) and can compute its posterior uncertainty more efficiently in a single forward pass. We compare with \gls{MCDGP} in our experiments (i.e., the DNN-GP + Dropout method in Section \ref{sec:building_block}). }

\textbf{Distance-preserving neural networks and bi-Lipschitz condition.}
The theoretic connection between distance preservation and the bi-Lipschitz condition is well-established \citep{searcod_metric_2006}, and learning an approximately isometric, distance-preserving transform has been an important goal in the fields of dimensionality reduction \citep{blum_random_2006, perrault-joncas_metric_2017},  generative modeling \citep{lawrence_local_2006, dinh_nice:_2014, dinh_density_2016, jacobsen_i-revnet_2018}, and adversarial robustness \citep{jacobsen_excessive_2018, ruan_reachability_2018, sokolic_robust_2017, tsuzuku_lipschitz-margin_2018, weng_evaluating_2018}. This work is a novel application of the distance preservation property for uncertainty quantification. There are several methods for controlling the Lipschitz constant of a \gls{DNN} (e.g., gradient penalty or norm-preserving activation \citep{an_how_2015, anil_sorting_2019, chernodub_norm-preserving_2017, gulrajani_improved_2017}), and we chose spectral normalization in this work due to its simplicity and its minimal impact on a \gls{DNN}'s architecture and the optimization dynamics \citep{bartlett_representing_2018, behrmann_invertible_2019, rousseau_residual_2020, behrmann2021understanding}. Finally, \cite{smith2021can} studied the effect of spectral regularization on the residual networks in the context of image recognition, and concluded that, under mild assumptions, residual networks will be approximately distance preserving on the low-passed portion of their input.

\textbf{Open Set Classification.} The uncertainty risk minimization problem in Section \ref{sec:theory} assumes a data-generation mechanism similar to the \textit{open set recognition} problem \citep{scheirer_probability_2014}, where the whole input space is partitioned into known and unknown domains. However, our analysis is unique in that it focuses on measuring a model's behavior in uncertainty quantification and takes a rigorous, decision-theoretic approach to the problem. As a result, our analysis works with a special family of risk functions (i.e., \textit{the strictly proper scoring rule}) that measure a model's performance in uncertainty calibration. Furthermore, it handles the existence of unknown domain via a minimax formulation, and derives the solution by using a generalized version of maximum entropy theorem for the Bregman scores \citep{grunwald_game_2004, landes_probabilism_2015}. The form of the optimal solution we derived in (\ref{eq:minimax_solution}) takes an intuitive form, and has been used by many empirical work as a training objective to leverage adversarial training and generative modeling to detect \gls{OOD} examples \citep{hafner_reliable_2018, harang_principled_2018, lee_training_2018, malinin_prior_2018, meinke_towards_2020, hendrycks_deep_2018}. Our analysis provides theoretical support for these practices in verifying rigorously the uniqueness and optimality of this solution, and also provides a conceptual unification of the notion of calibration and the notion of \gls{OOD} generalization. Furthermore, it is used in this work to motivate a design principle (\textit{distance awareness}) that enables strong \gls{OOD} performance in discriminative classifiers without the need of explicit generative modeling.

\section{Benchmarking Experiments}
\label{sec:experiments}
In this section, we benchmark the performance of the SNGP model by applying it on a variety of both toy datasets and real-world tasks across different modalities
Specifically, we first benchmark the behavior of the approximate GP layer versus an exact GP, and illustrate the impact of spectral normalization on toy regression and classification tasks (Section \ref{sec:exp_low_d}). We then conduct a thorough benchmark study to compare the performance of \gls{SNGP} against the other state-of-the-art methods on popular benchmarks such as CIFAR-10 and CIFAR-100  (Section \ref{sec:cifar}. Finally, we illustrate the scalability of and the generality of the SNGP approach by applying it to a large-scale image recognition task (ImageNet, Section \ref{sec:imagenet}), and highlight the broad usefulness by applying SNGP to uncertainty tasks in two other data modalities, namely  conversational intent understanding  and genomics sequence identification (Section \ref{sec:exp_other}).

\subsection{Low-dimensional Experiments}
\label{sec:exp_low_d}
\subsubsection{Regression}

\begin{figure}[th!]
    \centering
    \subcaptionbox{Exact GP\label{fig:1d}}{
    \includegraphics[width=0.32\columnwidth]{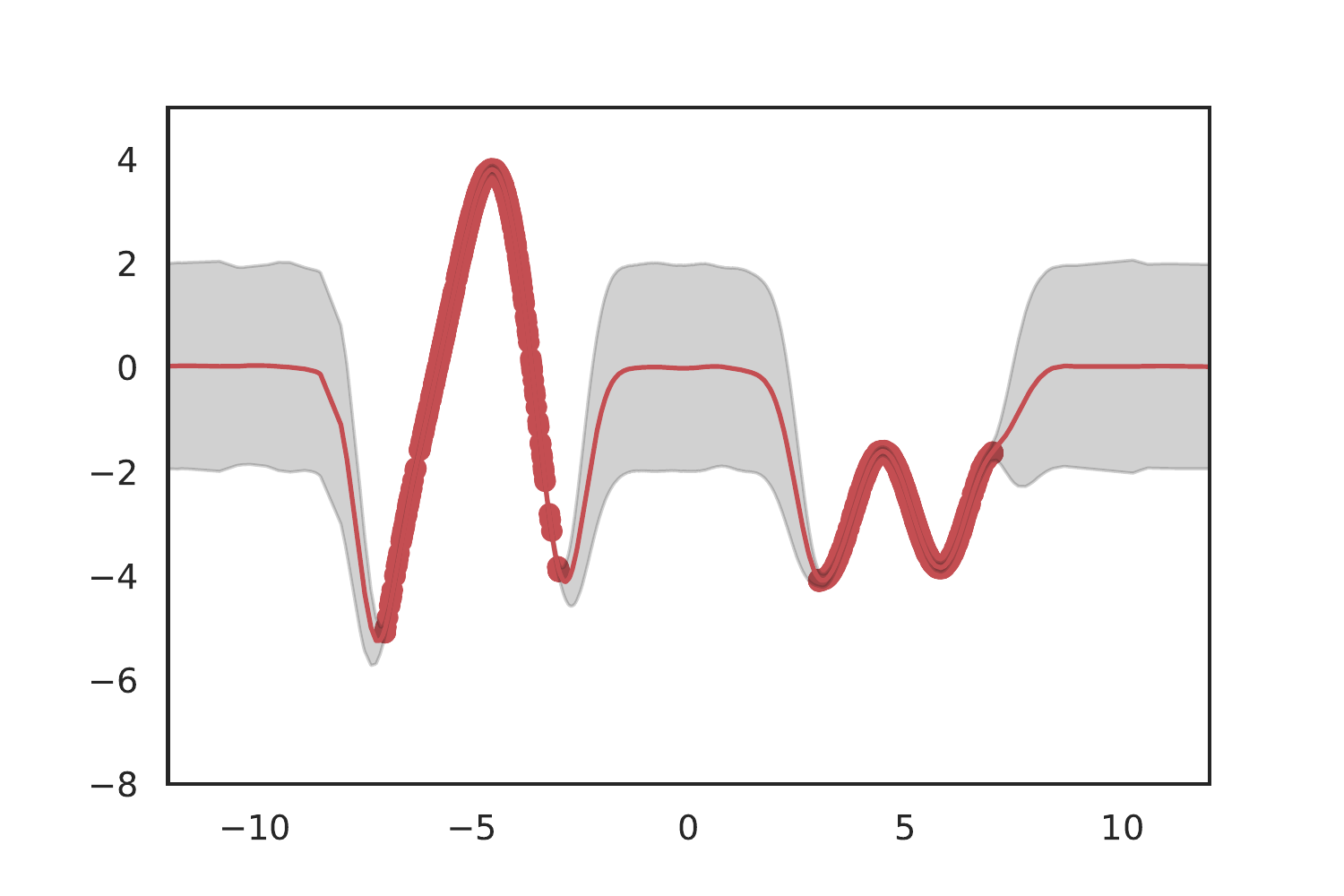}}
    \subcaptionbox{Variational GP
    }{
    \includegraphics[width=0.32\columnwidth]{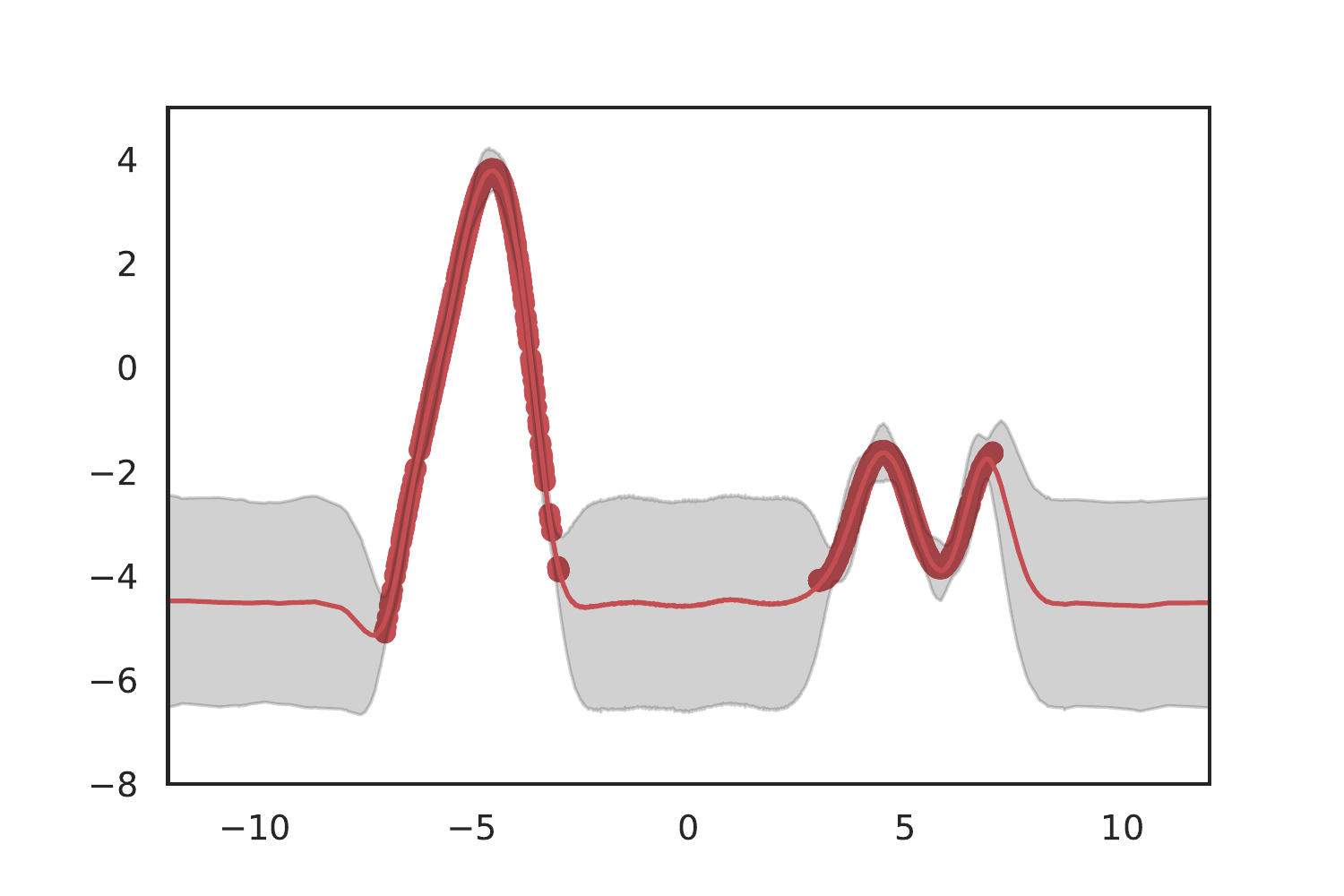}}
    \subcaptionbox{Random Fourier Features
    }{
    \includegraphics[width=0.32\columnwidth]{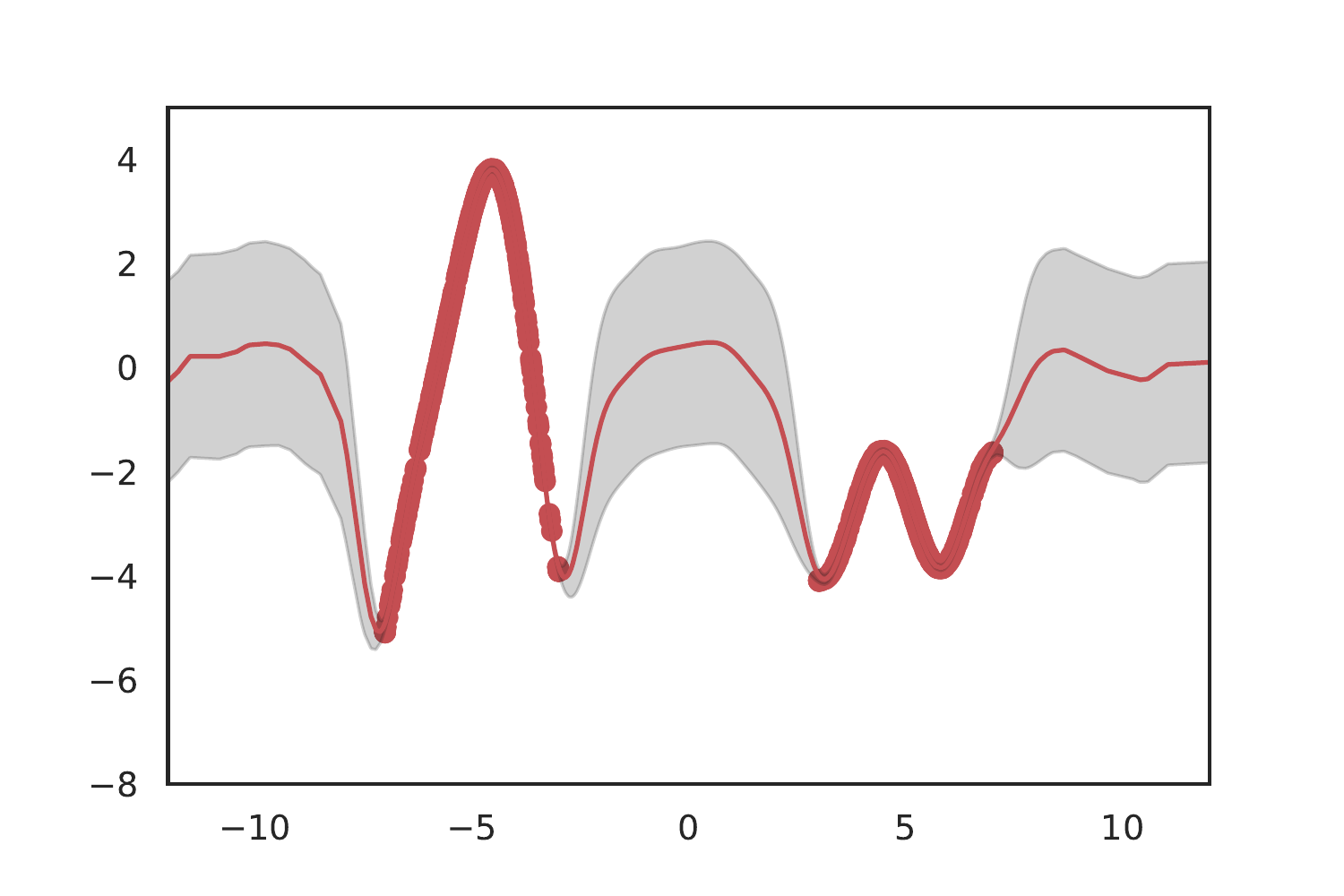}}
   \vspace{-0.5em}
    \caption{
    \small{
    1D toy Regression Task: The models are trained using the training datapoints sampled from a bimodal Gaussian distribution (shown as blue dots), and then return both a posterior mean and variance for the entire test domain $[-12, 12]$. Exact GP (left) returns predictions with zero mean and high posterior variance on test points away from the training data distribution, and low posterior variance on the training points, whereas Variational GP (middle) predicts a mean that is negative on unseen points. RFF-GP (right) closely mimics the behavior of Exact GP, with the addition of some periodic fluctuations arising due to the use of Fourier features to approximate the covariance matrix.
    }}
    \label{fig:1d_exp}
\end{figure}

We first benchmark the performance of the \textit{Random-Feature Gaussian process} layer (RFGP), which form a key component of the SNGP model, versus both an exact GP formulation and a variational Gaussian process (VGP) \citep{titsias_variational_2009} on a 1D toy regression task. We consider the bimodal toy regression example from \cite{van2021feature}, and use use the \gls{RBF} kernel for all three models
. As seen from Figure \ref{fig:1d}, the GP posterior mean for the exact GP tends to zero in the absence of training data, with low variance near training points and high variance otherwise. The RFGP mimics this behavior, with the addition of some periodic artifacts, which arises naturally due to the periodic nature of the approximation. In comparison, the prediction from the VGP model at locations that are further away from data tend to depend on the location of the inducing points, and does not revert back to the original zero-mean Gaussian process prior (e.g, between the two  modes and at two ends of Figure \ref{fig:1d_exp}). Although a more modern treatment of VGP may address this issue \citep{dutordoir2020sparse} \footnote{Note that the goal of this toy example is not to claim RFGP outperforms VGP, but just to illustrate that the RFGP, despite its simplicity, indeed returns a valid uncertainty surface that is distance-aware. To this end, \cite{van2021feature} studied the extension of SNGP method under VGP models, and obtained promising result on toy datasets.
}.

\subsubsection{Classification}
\label{sec:exp_2d}

After having benchmarked the predictive and uncertainty behavior of the RFF-GP algorithm compared to an exact GP formulation, in the subsequent subsections we use it as a subcomponent in the SNGP algorithm. 

\begin{figure}[th!]
    \centering
    \subcaptionbox{Gaussian Process\label{fig:2d_exp_oval_gp}}{
    \includegraphics[width=0.18\columnwidth]{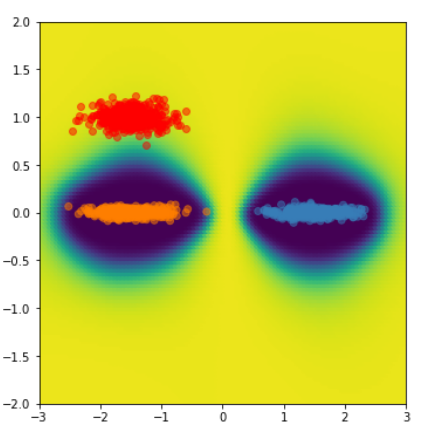}}
    \subcaptionbox{Deep Ensemble\label{fig:2d_exp_oval_deepens}}{
    \includegraphics[width=0.18\columnwidth]{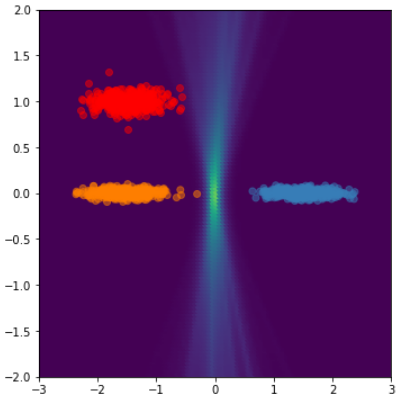}}
    \subcaptionbox{MC Dropout\label{fig:2d_exp_oval_mc}}{
    \includegraphics[width=0.18\columnwidth]{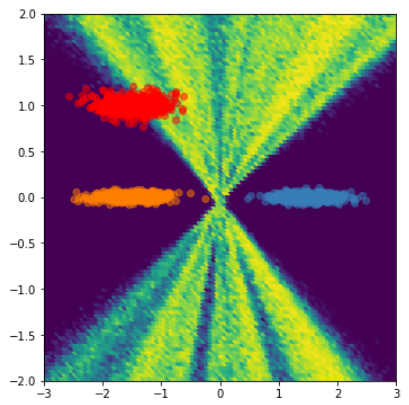}}
    \subcaptionbox{DNN-GP \label{fig:2d_exp_oval_deepgp}}{
    \includegraphics[width=0.18\columnwidth]{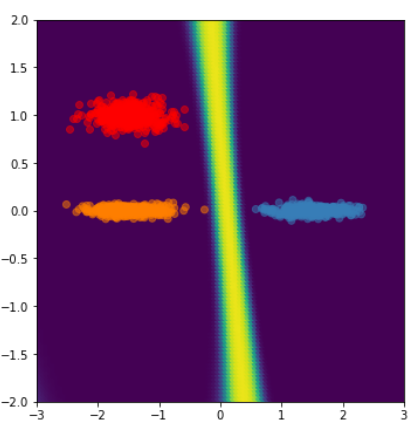}}
    \subcaptionbox{\gls{SNGP} (Ours)\label{fig:2d_exp_oval_sngp}}{
    \includegraphics[width=0.21\columnwidth]{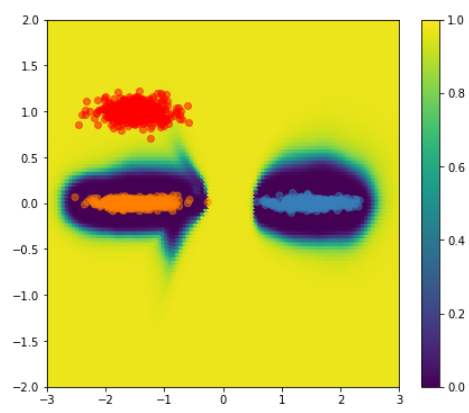}}

    \subcaptionbox{Gaussian Process\label{fig:2d_exp_moon_gp}}{
    \includegraphics[width=0.18\columnwidth]{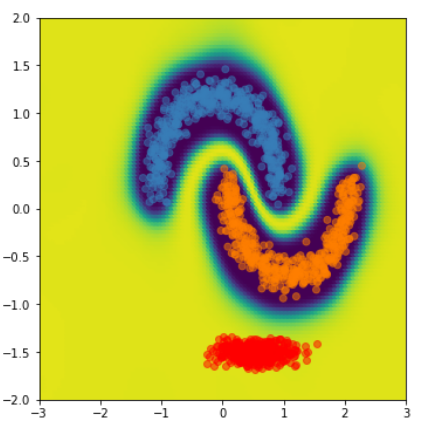}}
    \subcaptionbox{Deep Ensemble\label{fig:2d_exp_moon_deepens}}{
    \includegraphics[width=0.18\columnwidth]{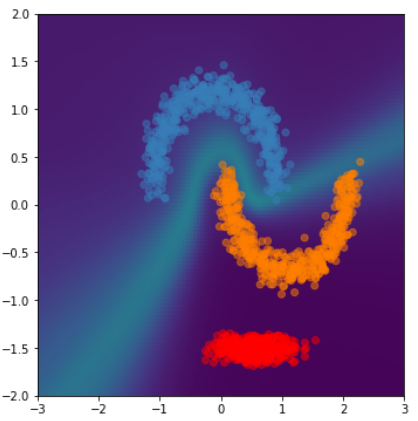}}
    \subcaptionbox{MC Dropout\label{fig:2d_exp_moon_mc}}{
    \includegraphics[width=0.18\columnwidth]{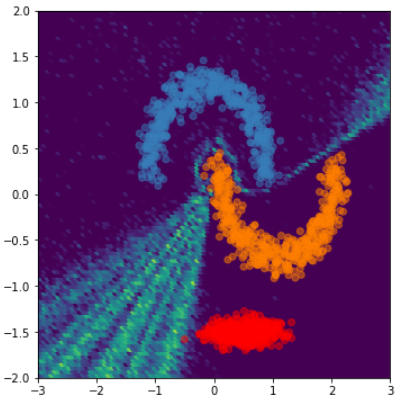}}
    \subcaptionbox{DNN-GP \label{fig:2d_exp_moon_deepgp}}{
    \includegraphics[width=0.18\columnwidth]{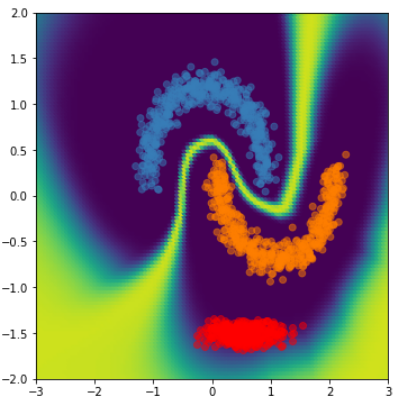}}
    \subcaptionbox{\gls{SNGP} (Ours)\label{fig:2d_exp_moon_sngp}}{
    \includegraphics[width=0.21\columnwidth]{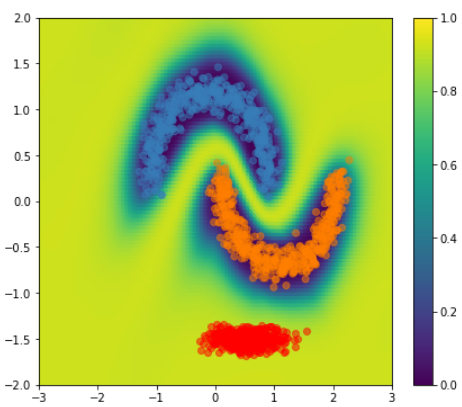}}
   \vspace{-0.5em}
    \caption{
    \small{
    The uncertainty surface of a \gls{GP} and different DNN approaches on the \textit{two ovals} (Top Row) and \textit{two moons} (Bottom Row) 2D classification benchmarks. SNGP is the only \gls{DNN}-based approach achieving a distance-aware uncertainty similar to the gold-standard \gls{GP}.
    Training data for positive ({\color{BurntOrange}{\textbf{Orange}}}) and negative classes ({\color{black}\textbf{Blue}}). OOD data ({\color{red}\textbf{Red}}) not observed during training. Background color represents the  estimated model uncertainty (See \ref{fig:2d_exp_oval_sngp} and \ref{fig:2d_exp_moon_sngp} for color map).
    See Section 5.1 for details. Visualization of additional methods (e.g., a single deterministic \gls{DNN} and ensemble of \gls{SNGP}s) is shown in Figure \ref{fig:ensemble_vs_basemodel}.
    }}
     \vspace{-0.9em}
    \label{fig:2d_exp}
\end{figure}

In this subsection, we compare the behavior of SNGP on a suite of 2D classification tasks. Specifically, we consider the \textit{two ovals} benchmark (Figure \ref{fig:2d_exp}, row 1) and the \textit{two moons} benchmark (Figure \ref{fig:2d_exp}, row 2). The \textit{two ovals} benchmark consists of two near-flat Gaussian distributions, which represent the two in-domain classes (orange and blue) that are separable by a linear decision boundary. There also exists an \gls{OOD} distribution (red) that the model doesn't observe during training. Similarly, the \textit{two moons} dataset consists of two moon-shaped distributions separable by a non-linear decision boundary.
For both benchmarks, we sample 500 observations $\bx_i=(x_{1i}, x_{2i})$ from each of the two in-domain classes (orange and blue), and consider a deep architecture ResFFN-12-128, which contains 12 residual feedforward layers with 128 hidden units and dropout rate 0.01. The input dimension is projected from 2 dimensions to the 128 dimensions using a dense layer.

In addition to \gls{SNGP}, we also visualize the uncertainty surface of the below approaches: \textbf{\glsfirst{GP}} is a standard Gaussian process directly taking $\bx_i$ as input and was trained with \gls{HMC}. In low-dimensional datasets, \gls{GP} is often considered the gold standard for uncertainty quantification. \textbf{Deep Ensemble} is an ensemble of 10 ResFFN-12-128 models with dense output layers. \textbf{\gls{MCD}} uses a single ResFFN-12-128 model with dense output layer and 10 dropout samples at test time. \textbf{DNN-GP} uses a single ResFFN-12-128 model with the \gls{GP} Layer (described in Section \ref{sec:gp}) without spectral normalization. Finally, \textbf{\gls{SNGP}} uses a single ResFFN-12-128 model with the \gls{GP} layer and with spectral normalization. The full experimental details for these algorihms are in Appendix \ref{sec:exp_app}.

Figure \ref{fig:2d_exp} shows the results of training these algorithms to achieve high (close to $100 ~\%$ accuracy) on the test data, and visualizes the uncertainty surface output by each model. Each algorithm returns a predictive distribution of the form $p(y|\bx)$, following which the confidence of the prediction can be written as $\hat{p} = \max_y p(y|\bx)$. We plot the uncertainty surface by defining $u(x) = \hat{p} (1 - \hat{p})$, and normalize it to the range of $[0, 1]$ by dividing it by 0.25. Firstly, we notice that the exact Gaussian process models (without using a \gls{DNN}, Figures \ref{fig:2d_exp_oval_gp}, \ref{fig:2d_exp_moon_gp})  exhibit the expected behavior for high-quality predictive uncertainty: 
they predict low uncertainty in the region $\Xsc_{\texttt{IND}}$ supported by the training data (blue color), and predict high uncertainty when $\bx$ is far from $\Xsc_{\texttt{IND}}$ (yellow color), i.e., \textit{distance-awareness}. As a result, the exact \gls{GP} model is able to assign low confidence to the \gls{OOD} data (colored in red), indicating reliable uncertainty quantification. On the other hand, Deep Ensemble (Figures \ref{fig:2d_exp_oval_deepens},  \ref{fig:2d_exp_moon_deepens}) and MC Dropout (Figures \ref{fig:2d_exp_oval_mc}, \ref{fig:2d_exp_moon_mc}) are based on dense output layers that are not \textit{distance aware}. As a result, both methods quantify their predictive uncertainty based on the distance from the decision boundaries, assigning low uncertainty to \gls{OOD} examples even if they are far from the data, due to a pathological behavior where the uncertainty is only high near the decision boundary (linear for \textit{two ovals} and non-linear for \textit{two moons}).
Finally, the DNN-GP (Figures \ref{fig:2d_exp_oval_deepgp} and \ref{fig:2d_exp_moon_deepgp}) and \gls{SNGP} (Figures \ref{fig:2d_exp_oval_sngp} and\ref{fig:2d_exp_moon_sngp}) both use \gls{GP} as their output layers, but with \gls{SNGP} additionally imposing the spectral normalization on its hidden mapping $h(.)$. As a result, the DNN-GP's uncertainty surfaces are still strongly impacted by the distance from decision boundary, likely caused by the fact that the un-regularized hidden mapping $h(\bx)$ is free to discard information that is not relevant for prediction. On the other hand, \gls{SNGP} is able to maintain the \textit{distance-awareness} property via its bi-Lipschitz constraint, and exhibits a uncertainty surface that is analogous to the gold-standard model (exact \gls{GP}) despite the fact that \gls{SNGP} is based on a deep 12-layer network.

\subsection{Image Classification}
\paragraph{Baseline Methods} All methods included in the vision and language understanding experiments are summarized in Table \ref{tab:methods}. 
We compare SNGP against a suite of algorithms; a deterministic baseline, two single-model approaches: 
\textbf{\gls{DUQ}}, and \textbf{\gls{IsoMax+}}.
The original DUQ consists of two novel components, one is a RBF kernel based loss function which computes the distance between the feature vector and the class centroids, and the other is adding gradient penalty to avoid feature collapse. 
As pointed in the paper, RBF networks prove difficult to optimize and scale to large number of output classes (e.g., CIFAR-100), so we instead replace the RBF layer with our GP layer. Thus we call this variant as \textbf{\gls{DUQ-GP}}.
\textbf{IsoMax+} replaces the softmax cross-entropy loss with a IsoMax+ loss which minimizes the distance to the correct class prototype based on the normalized feature embeddings in the last layer \citep{macedo2021enhanced}.
We also include two ablations of \gls{SNGP}: \textbf{DNN-SN} which uses spectral normalization on its hidden weights and a dense output layer (i.e. distance preserving hidden mapping without distance-aware output layer), and \textbf{DNN-GP} which uses the \gls{GP} as output layer but without spectral normalization on its hidden layers (i.e., distance-aware output layer without distance-preserving hidden mapping). Finally, to evaluate the SNGP's efficacy as a base model for ensemble approaches, we also train ensemble of SNGP and compare it with two popular approaches that quantifies representation diversity:
\textbf{\gls{MCD}} (with 10 dropout samples) and \textbf{Deep Ensemble} (with 10 models), both are trained with a dense output layer and no spectral regularization. Section \ref{sec:building_block} further explores the interplay between SNGP and various ensemble approaches.

For all models that use \gls{GP} layer, we keep $D_L=1024$ and compute predictive distribution by performing mean-field approximation to the softmax Gaussian posterior. Further experiment details and recommendations for practical implementation are in Appendix \ref{sec:exp_app}. 
All experiments are built on the \href{https://github.com/google/uncertainty-baselines}{\texttt{uncertainty\_baselines}} framework \footnote{\url{https://github.com/google/uncertainty-baselines}}  \citep{nado2021uncertainty}. We open-source our code there.
\begin{table}[ht]
    \centering
    \scalebox{0.85}{
    \begin{tabular}{c|cccc}
    \toprule
      & Additional & Output & Representation & Multi-pass \\
     Methods & Regularization & Layer & Uncertainty & Inference \\
     \midrule
     DNN & - & Dense & - & - \\
     \midrule
     DNN-IsoMax+ & - & Dense & - & - \\
     \gls{DUQ-GP} & Gradient Penalty & GP & - & - \\
     \midrule
     {DNN-SN} & Spec Norm & Dense & - & - \\
     {DNN-GP} & - & GP & - & - \\
     \gls{SNGP} & Spec Norm & GP & - & - \\
     \midrule
     \midrule
     \gls{MCD} & Dropout & Dense & Yes & Yes \\
     Deep Ensemble & - & Dense & Yes & Yes \\
     SNGP Ensemble & Spec Norm & GP & Yes & Yes \\
     \bottomrule
    \end{tabular}
    }
    \captionsetup{justification=centering}
    \caption{Summary of methods used in experiments. 
    Multi-pass Inference refers to whether the method needs to perform multiple forward passes to generate the predictive distribution.
    }
    \label{tab:methods}
\end{table}

\subsubsection{CIFAR-10 and CIFAR-100}
\label{sec:cifar}


For the CIFAR-10 and CIFAR-100 image classification benchmarks, we use a Wide ResNet 28-10 model as the base for all methods \citep{zagoruyko_wide_2017}. 
Following the benchmarking setup first suggested in \cite{ovadia_can_2019},
we evaluate the model's predictive accuracy, negative log-likelihood (NLL), and  expected calibration error (ECE) under both clean CIFAR testing data and its corrupted versions termed CIFAR-*-C 
\citep{hendrycks_benchmarking_2019}. To evaluate the model's \gls{OOD} detection performance, we consider two tasks: a  standard \textit{far}-\gls{OOD} task using \glsunset{SVHN}\gls{SVHN} as the \gls{OOD} dataset for a model trained on CIFAR-10/-100, and a difficult \textit{near}-\gls{OOD} task using CIFAR-100 as the \gls{OOD} dataset for a model trained on CIFAR-10, and vice versa. 
We use the in-distribution training data statistics to preprocess and normalize the OOD datasets.
In Tables \ref{tab:cifar10} and \ref{tab:cifar100}, we use the maximum softmax probability (MSP) as the uncertainty score while performing OOD evaluation, and report the Area Under the Reciever Operator Curve (AUROC) metric.

Tables \ref{tab:cifar10}-\ref{tab:cifar100} report the results. As shown, for predictive accuracy, \gls{SNGP} is competitive with other single-model approaches.
For calibration error, \gls{GP}-based models clearly outperform the other single-model approaches and are competitive with Deep Ensemble and \gls{MCD} approaches. We also observe that under MSP metric, \gls{IsoMax+} method produces underconfident predictions, leading to the highest calibration error. Furthermore, its performance degrades quickly on more complex tasks (e.g., CIFAR-100) both in terms of accuracy and in uncertainty quality\footnote{Contemporary to our work, \cite{macedo2022distinction} shows the performance of IsoMax++ in OOD detection can be improved by using more specialized metrics (e.g., minimum distance score (MDS)) instead of MSP.}.
That is consistent with the findings of  \citet{padhy2020revisiting}. 
For \gls{OOD} detection, \gls{SNGP} mostly outperforms other single-model approaches that are based on a dense output layer (except for CIFAR-10 vs. CIFAR-100 where \gls{IsoMax+} does best),
and is competitive with deep ensembles, \gls{MCD} approaches and \gls{DUQ-GP}. Finally, ensemble using SNGP as base model strongly outperforms all the other approaches, illustrating the importance of the \textit{distance-awareness} property for high-quality performance in uncertainty quantification and the composability of SNGP as a building block toward the state-of-the-art probabilistic deep models.

\begin{table}[ht]
\centering
\resizebox{\textwidth}{!}{  
\begin{tabular}{ccc|cc|cc|cc}
\toprule
& \multicolumn{2}{c|}{Accuracy ($\uparrow$)} & 
\multicolumn{2}{c|}{ECE ($\downarrow$)} &
\multicolumn{2}{c|}{NLL ($\downarrow$)} &
\multicolumn{2}{c}{OOD AUROC ($\uparrow$)} 
\\
Method  & Clean & Corrupted & Clean & Corrupted & Clean & Corrupted & 
SVHN & CIFAR-100 
\\
\midrule
\multicolumn{9}{c}{\textbf{Single Model}} \\
\midrule
DNN & 95.8 $\pm$ 0.190 & 79.0 $\pm$ 0.350 & 0.028 $\pm$ 0.002 & 0.153 $\pm$ 0.005 & 0.183 $\pm$ 0.007 & 1.042 $\pm$ 0.038 & 0.946 $\pm$ 0.005 & 0.893 $\pm$ 0.001 
\\
\midrule
DNN-IsoMax+ & 95.6 $\pm$ 0.190 & 79.1 $\pm$ 0.230 & 0.525 $\pm$ 0.003 & 0.439 $\pm$ 0.003 & 0.214 $\pm$ 0.008 & 1.276 $\pm$ 0.041 & \textbf{0.959 $\pm$ 0.006} & \textbf{0.918 $\pm$ 0.002}
\\
\glsunset{DUQ}\gls{DUQ}-GP  & 95.8 $\pm$ 0.120 & 78.2 $\pm$ 0.460 & 0.027 $\pm$ 0.001 & 0.149 $\pm$ 0.005 & 0.186 $\pm$ 0.008 & 1.160 $\pm$ 0.049 & 
0.939 $\pm$ 0.007 & 0.900 $\pm$ 0.003
\\
\midrule
DNN-SN & \textbf{96.0 $\pm$ 0.170} & 79.1 $\pm$ 0.290 & 0.027 $\pm$ 0.002  & 0.150 $\pm$ 0.004 &  0.179 $\pm$ 0.006 & 1.019 $\pm$ 0.027 & {0.945 $\pm$ 0.005} & 0.894 $\pm$ 0.002 
\\
DNN-GP & 95.8 $\pm$ 0.150 & 79.1 $\pm$ 0.430 & \textbf{0.017 $\pm$ 0.003} & 0.100 $\pm$ 0.009 &  \textbf{0.149 $\pm$ 0.006} & 0.761 $\pm$ 0.039 & \textbf{0.964 $\pm$ 0.006} & 0.902 $\pm$ 0.002 
\\
\gls{SNGP} (Ours) & 95.7 $\pm$ 0.140 & \textbf{79.3 $\pm$ 0.340} & \textbf{0.017 $\pm$ 0.003} & \textbf{0.099$\pm$ 0.008} & \textbf{0.149 $\pm$ 0.005} & \textbf{0.745 $\pm$ 0.026} & \textbf{0.960 $\pm$ 0.004} & {0.902 $\pm$ 0.003} 
\\
\midrule \midrule
\multicolumn{9}{c}{\textbf{Ensemble Model}} 
\\
\midrule
\gls{MCD} & 95.7 $\pm$ 0.130 & 78.6 $\pm$ 0.430 & 0.013 $\pm$ 0.002 & 0.099 $\pm$ 0.005 & 0.145 $\pm$ 0.004 & 0.829 $\pm$ 0.021 & 0.934 $\pm$ 0.004 & 0.903 $\pm$ 0.001 
\\
Deep Ensemble & \textbf{96.4 $\pm$ 0.090} & 80.4 $\pm$ 0.150 & 0.011 $\pm$ 0.001  & 0.092 $\pm$ 0.003  & 0.124 $\pm$ 0.001 & 0.768 $\pm$ 0.009 & 0.947 $\pm$ 0.002 & 0.914 $\pm$ 0.000
\\
SNGP Ensemble (Ours)  &   \textbf{96.4 $\pm$ 0.040} & \textbf{81.1 $\pm$ 0.130} & \textbf{0.009 $\pm$ 0.001}  & \textbf{0.042 $\pm$ 0.002}  & \textbf{0.114 $\pm$ 0.001} & \textbf{0.590 $\pm$ 0.005} & \textbf{0.967 $\pm$ 0.002} & \textbf{0.920 $\pm$ 0.001} 
\\
\bottomrule
\end{tabular}
}
\caption{
Results for Wide ResNet-28-10 on CIFAR-10, averaged over 10 seeds. 
Bold indicates the on-average best-performing methods among the single-model methods and the ensemble methods. 
SNGP almost always outperform other single model variants, and SNGP ensemble outperform other ensemble methods. 
}
\label{tab:cifar10}
\end{table}

\begin{table}[ht]
\centering
\resizebox{\textwidth}{!}{  
\begin{tabular}{ccc|cc|cc|cc}
\toprule
& \multicolumn{2}{c|}{Accuracy ($\uparrow$)} & 
\multicolumn{2}{c|}{ECE ($\downarrow$)} &
\multicolumn{2}{c|}{NLL ($\downarrow$)} &
\multicolumn{2}{c}{OOD AUROC ($\uparrow$)} 
\\
Method  & Clean & Corrupted & Clean & Corrupted & Clean & Corrupted & 
SVHN & CIFAR-10 
\\
\midrule
\multicolumn{9}{c}{\textbf{Single Model}} \\
\midrule
DNN & 80.4 $\pm$ 0.290 & {55.0 $\pm$ 0.180} & 0.107 $\pm$ 0.004 & 0.258 $\pm$ 0.004 & 0.941 $\pm$ 0.016 & 2.663 $\pm$ 0.045 & 0.799 $\pm$ 0.020 & {0.795 $\pm$ 0.001} 
\\
\midrule
DNN+IsoMax+ & 77.3 $\pm$ 0.330 & 52.9 $\pm$ 0.190 & 0.663 $\pm$ 0.004 & 0.455 $\pm$ 0.002 & 1.289 $\pm$ 0.014 & 3.800 $\pm$ 0.069 & 0.770 $\pm$ 0.026 & 0.786 $\pm$ 0.002 
\\
\gls{DUQ}-GP  & 79.7 $\pm$ 0.200 & 54.0 $\pm$ 0.280 & 0.112 $\pm$ 0.002 & {0.269 $\pm$ 0.006} &  0.988 $\pm$ 0.013 & 2.998 $\pm$ 0.100 & {0.777 $\pm$ 0.026} & {0.789 $\pm$ 0.002} 
\\
\midrule
DNN-SN & \textbf{80.5 $\pm$ 0.300} & 55.0 $\pm$ 0.210 & 0.111$\pm$ 0.002 & 0.268$\pm$ 0.005 &  0.951 $\pm$ 0.008 & 2.727$\pm$ 0.055 & 0.798$\pm$ 0.023 & 0.793$\pm$ 0.003 
\\
DNN-GP & 80.3 $\pm$ 0.380 & \textbf{55.3 $\pm$ 0.250} & 0.034$\pm$ 0.005 & \textbf{0.059$\pm$ 0.002} & 0.767$\pm$ 0.008 & \textbf{1.918$\pm$ 0.016} &  0.835$\pm$ 0.021 & 0.797$\pm$ 0.001 
\\
\gls{SNGP} (Ours) & {80.3 $\pm$ 0.230} & \textbf{55.3 $\pm$ 0.190} & \textbf{0.030 $\pm$ 0.004} & {0.060 $\pm$ 0.004} &  \textbf{0.761 $\pm$ 0.007} & 1.919 $\pm$ 0.013 & \textbf{0.846 $\pm$ 0.019} & \textbf{0.798 $\pm$ 0.001} 
\\
\midrule\midrule
\multicolumn{9}{c}{\textbf{Ensemble Model}} 
\\
\midrule
\gls{MCD} & 80.2 $\pm$ 0.220 & 54.6 $\pm$ 0.002 & {0.031 $\pm$ 0.002} & 0.136 $\pm$ 0.005 & 0.762 $\pm$ 0.008 & 2.198 $\pm$ 0.027 & 0.800 $\pm$ 0.014 & 0.797 $\pm$ 0.002 
\\
Deep Ensemble & \textbf{82.5 $\pm$ 0.190} & 57.6 $\pm$ 0.110 & {0.041 $\pm$ 0.002} & {0.146$\pm$ 0.003} &  {0.674 $\pm$ 0.004} & {2.078 $\pm$ 0.015} & {0.812 $\pm$ 0.007} & {0.814 $\pm$ 0.001} 
\\
SNGP Ensemble (Ours) & \textbf{82.5 $\pm$ 0.160} &	\textbf{58.2 $\pm$ 0.130} & \textbf{0.028 $\pm$ 0.002} & \textbf{0.064 $\pm$ 0.001} &  \textbf{0.635 $\pm$ 0.003} & \textbf{1.752 $\pm$ 0.007} & \textbf{0.838 $\pm$ 0.008} & \textbf{0.819 $\pm$ 0.001} 
\\
\bottomrule
\end{tabular}
}
\caption{
Results for Wide ResNet-28-10 on CIFAR-100, averaged over 10 seeds.
Bold indicates the on-average best-performing methods among the single-model methods and the ensemble methods. 
SNGP almost always outperform other single model variants, and SNGP ensemble outperform other ensemble methods. 
}
\label{tab:cifar100}
\end{table}

In Appendix \ref{sec:ood_metrics} we studied the performance of other uncertainty scores including Dempster-Shafer \citep{sensoy2018evidential}, Mahalanobis distance \citep{lee_simple_2018-2}, and Relative Mahalanobis distance \citep{ren2021simple} for \gls{OOD} detection. We found in general the Dempster Shafer metric attains a better \gls{OOD} performance (e.g., $0.984$ AUROC for CIFAR-10 v.s. SVHN) than the widely-used MSP metric reported in the main text.



\subsubsection{ImageNet}
\label{sec:imagenet}

We illustrate the scalability of SNGP by experimenting on the large-scale ImageNet dataset \citep{russakovsky_imagenet_2015} using a ResNet-50 model as the base for all methods. Similar to the CIFAR benchmarking setup, we follow the procedure from~\citep{ovadia_can_2019} and evaluate the model's predictive accuracy and calibration error under both clean ImageNet testing data and its corrupted versions termed ImageNet-C. 
For this complex, large-scale task with high-dimensional output, we find it difficult to scale some of the other single-model methods (e.g., IsoMax or DUQ), which tend to over-constrain the model expressiveness and lead to lower accuracy than a baseline \gls{DNN} (a phenomenon we already observe in the CIFAR-100 experiment). Therefore we omit them from results.
Table \ref{tab:imagenet} reports the results of methods that achieve competitive performance on ImageNet. As shown, for predictive accuracy, \gls{SNGP} is competitive with that of a deterministic network, despite using a last-layer GP and spectral normalization on the weights of the residual network. For calibration error, \gls{SNGP} not only outperforms the other single-model approaches and also, interestingly, the Deep Ensemble. Finally, the ensemble using SNGP as the base model attains the strongest predictive accuracy and NLL across all methods. These results illustrate the importance of the \textit{distance-awareness} property for high-quality performance in uncertainty quantification, and the proposed components (i.e., spectral normalization and the GP-layer) perform well in complex, large-scale tasks by maintaining model accuracy while improving the quality of its uncertainty quantification.


\begin{table}[ht]
\centering
\resizebox{0.8\textwidth}{!}{%
\begin{tabular}{ccc|cc|cc}
\toprule
\textbf{} & \multicolumn{2}{c|}{\textbf{Accuracy ($\uparrow$)}} & \multicolumn{2}{c|}{ECE ($\downarrow$)} & \multicolumn{2}{c}{NLL ($\downarrow$)} 
\\
Method & Clean & Corrupted & Clean & Corrupted & Clean & Corrupted 
\\ 
\midrule
\multicolumn{7}{c}{\textbf{Single Model}} \\
\midrule
DNN & 76.2 $\pm$ 0.01 & 40.5 $\pm$ 0.01 & 0.032 $\pm$ 0.002 & 0.103 $\pm$ 0.011 & 0.939 $\pm$ 0.01 & 3.21 $\pm$ 0.02 
\\ 
\hline
DNN-SN & \textbf{76.4 $\pm$ 0.01} & 40.6 $\pm$ 0.01 & 0.079 $\pm$ 0.001 & 0.074 $\pm$ 0.001 & 0.96 $\pm$ 0.01 & 3.14 $\pm$ 0.02 
\\
DNN-GP & 76.0 $\pm$ 0.01 & \textbf{41.3 $\pm$ 0.01} & 0.017 $\pm$ 0.001 & 0.049 $\pm$ 0.001 & 0.93 $\pm$ 0.01 & 3.06 $\pm$ 0.02 
\\
SNGP (Ours) & 76.1 $\pm$ 0.01 & 41.1 $\pm$ 0.01 & \textbf{0.013 $\pm$ 0.001} & \textbf{0.045 $\pm$ 0.012} & \textbf{0.93 $\pm$ 0.01} & \textbf{3.03 $\pm$ 0.01} 
\\
\midrule\midrule
\multicolumn{7}{c}{\textbf{Ensemble Model}} \\
\midrule
MC Dropout & 76.6 $\pm$ 0.01 & 42.4 $\pm$ 0.02 & 0.026 $\pm$ 0.002 & \textbf{0.046 $\pm$ 0.009} & 0.919 $\pm$ 0.01 & 2.96 $\pm$ 0.01 
\\
Deep Ensemble & 77.9 $\pm$ 0.01 & \textbf{44.9 $\pm$ 0.01} & \textbf{0.017 $\pm$ 0.001} & 0.047 $\pm$ 0.004 & {0.857 $\pm$ 0.01} & {2.82 $\pm$ 0.01} 
\\ 
SNGP Ensemble (Ours) & \textbf{78.1 $\pm$ 0.01} & \textbf{44.9 $\pm$ 0.01} & {0.039 $\pm$ 0.001} & {0.050 $\pm$ 0.002} &  \textbf{0.851 $\pm$ 0.01} & \textbf{2.77 $\pm$ 0.01} 
\\
\bottomrule
\end{tabular}%
}
\caption{Results for ResNet 50 on ImageNet, averaged over 10 seeds. 
Bold indicates the on-average best-performing methods among the single-model methods and the ensemble methods. 
SNGP outperforms other single model variants, and SNGP ensemble outperform other ensemble methods in generalization and NLL. 
}
\label{tab:imagenet}
\end{table}

\subsection{Generalization to other data modalities}
\label{sec:exp_other}

\subsubsection{Conversational Language Understanding}

To validate the hypothesis that \textit{distance awareness} is a crucial component for good predictive uncertainty on data modalities beyond images, we also evaluate \gls{SNGP} on a practical language understanding task where uncertainty quantification is of natural importance: dialog intent detection 
\citep{larson_evaluation_2019, vedula_towards_2019, yaghoub-zadeh-fard_user_2020, zheng_out--domain_2020}.
In a goal-oriented dialog system (e.g. chatbot) built for a collection of in-domain services, it is important for the model to understand if an input natural utterance from a user is in-scope (so it can activate one of the in-domain services) or out-of-scope (where the model should abstain). To this end, we consider training an intent understanding model using the CLINC \gls{OOS} intent detection benchmark dataset \citep{larson_evaluation_2019}. Briefly, the \gls{OOS} dataset contains data for 150 in-domain services with 150 training sentences in each domain, and also 1500 natural out-of-domain utterances. 
We train a $BERT_{\tt{ base}}$ model only on in-domain data, and evaluate their predictive accuracy on the in-domain test data, their calibration and \gls{OOD} detection performance on the combined in-domain and out-of-domain data. 
Due the lack of empirically validated implementations of other single-model methods for non-image modalities, we focus on comparing ablated versions of SNGP and the two general-purpose methods: \gls{MCD} and Deep Ensemble.
The results are in Table \ref{tab:clinc}. As shown, consistent with the previous vision experiments, \gls{SNGP} outperforms other single model approaches. It is competitive with Deep Ensemble in predictive accuracy and calibration, and outperforms all the approaches in \gls{OOD} detection.

\begin{table}[!h]
\centering
\scalebox{0.7}{
\begin{tabular}{cc|c|c|cc}
\toprule
& \multicolumn{1}{c}{Accuracy ($\uparrow$)} & 
\multicolumn{1}{c}{ECE ($\downarrow$)} &
\multicolumn{1}{c}{NLL ($\downarrow$)} &
\multicolumn{2}{c}{OOD} 
\\
Method  &   &   &  & 
AUROC ($\uparrow$) & AUPR ($\uparrow$) 
\\
\midrule
\multicolumn{6}{c}{\textbf{Single Model}} \\
\midrule
DNN & 96.5 $\pm$ 0.11 & 0.024 $\pm$ 0.002 & 3.559 $\pm$ 0.11 & 0.897 $\pm$ 0.01 & 0.757 $\pm$ 0.02 
\\
\midrule
DNN-SN & 95.4 $\pm$ 0.10 & 0.037 $\pm$ 0.004 & 3.565 $\pm$ 0.03 & 0.922 $\pm$ 0.02 & 0.733 $\pm$ 0.01 
\\
DNN-GP & 95.9 $\pm$ 0.07 & 0.075 $\pm$ 0.003 & 3.594 $\pm$ 0.02 & 0.941 $\pm$ 0.01 & 0.831 $\pm$ 0.01 
\\
\midrule
\gls{SNGP} (Ours) & \textbf{96.6 $\pm$ 0.05} & \textbf{0.014 $\pm$ 0.005} & \textbf{1.218 $\pm$ 0.03} & \textbf{0.969 $\pm$ 0.01} & \textbf{0.880 $\pm$ 0.01} 
\\ 
\midrule\midrule
\multicolumn{6}{c}{\textbf{Ensemble Model}} \\
\midrule
\gls{MCD} & 96.1 $\pm$ 0.10 & 0.021 $\pm$ 0.001  & 1.658 $\pm$ 0.05 & 0.938 $\pm$ 0.01 & 0.799 $\pm$ 0.01 
\\
Deep Ensemble & \textbf{97.5 $\pm$ 0.03} & {0.013 $\pm$ 0.002} & {1.062 $\pm$ 0.02} & {0.964 $\pm$ 0.01} & {0.862 $\pm$ 0.01} 
\\
SNGP Ensemble (Ours) & {97.4 $\pm$ 0.02} & \textbf{0.009 $\pm$ 0.002} & \textbf{0.918 $\pm$ 0.03} & \textbf{0.973 $\pm$ 0.02} & \textbf{0.910 $\pm$ 0.01} 
\\
\bottomrule
\end{tabular}
}
\caption{\small{
Results for BERT$_{\texttt{Base}}$ on CLINC \gls{OOS}, averaged over 10 seeds. 
Bold indicates the on-average best-performing methods among the single-model methods and the ensemble methods. 
SNGP outperforms other single model variants, and SNGP ensemble outperform other ensemble methods in uncertainty performance. 
}}
\vspace{-1em}
\label{tab:clinc}
\end{table}

\subsubsection{Bacteria Genomics Sequence Identification}

Here we apply SNGP to a genomic sequence prediction task as another data modality.
\cite{ren2019likelihood} proposed the genomics \gls{OOD} benchmark dataset motivated by the real-world problem of bacteria identification based on genomic sequences, which can be useful for diagnosis and treatment of infectious diseases. 
A classification model can be trained for classifying known bacteria species with decent test accuracy. However, when deploying the model to real data, the model will be inevitably exposed to the genomic sequences from unknown bacteria species. 
In fact, the real data can contain approximately $60-80\%$ of sequences from unknown classes that have not been studied before.
The model needs to be able to detect those out-of-distribution inputs from unknown species, and abstain from making predictions for them. 
The genomic \gls{OOD} benchmark dataset contains 10 bacteria classes as in-distribution for training, and 60 bacteria classes as out-of-distribution for testing. 
The in-distribution and out-of-distribution bacteria classes were chosen and separated naturally by the year of discovery, to mimic the real scenario that new bacteria species are discovered gradually over the years. 
Following \citep{ren2019likelihood}, we train a 1D CNN but with SN and GP components added on the 10 in-domain classes and evaluate the models' accuracy, \gls{ECE} and NLL on the in-domain test data, and also evaluate the model's OOD performance for detecting the 60 \gls{OOD} classes.
In addition to SNGP, we also consider Monte Carlo dropout and Deep Ensemble as another two baselines for comparison. We also study the effect of each component of SN and GP on the model's calibration and \gls{OOD} detection performance. Section \ref{sec:exp_detail} contains further model detail.

\begin{table}[ht]
\centering
\scalebox{0.8}{%
\resizebox{\columnwidth}{!}{%
\begin{tabular}{cc|c|c|cc}
\toprule
\multirow{2}{*}{Method} & \multirow{2}{*}{Accuracy ($\uparrow$)} & \multirow{2}{*}{ECE ($\downarrow$)} & \multirow{2}{*}{NLL ($\downarrow$)} & \multicolumn{2}{c}{OOD} \\
 &  &  &  &  \multicolumn{1}{c}{AUROC ($\uparrow$)} & \multicolumn{1}{c}{AUPR ($\uparrow$)} \\ \midrule
\multicolumn{6}{c}{\textbf{Single Model}} 
\\
\midrule
DNN & 84.40 $\pm$ 0.390 & 0.049 $\pm$ 0.007 & 0.487 $\pm$ 0.007 & 0.640 $\pm$ 0.005 & 0.609 $\pm$ 0.005 \\ \midrule
DNN-SN & 85.56 $\pm$ 0.150	& 0.025 $\pm$ 0.003 & 0.422 $\pm$ 0.004 & 0.658 $\pm$ 0.006 &	0.625 $\pm$ 0.005 \\
DNN-GP & 85.23 $\pm$ 0.200 &	0.041 $\pm$ 0.006 & 0.457 $\pm$ 0.005	& 0.654 $\pm$ 0.006 &	0.629 $\pm$ 0.003 \\  \midrule
SNGP (Ours) & \textbf{85.71 $\pm$ 0.100} &	\textbf{0.019 $\pm$ 0.004} &	\textbf{0.417 $\pm$ 0.004} &	\textbf{0.672 $\pm$ 0.011} &	\textbf{0.637 $\pm$ 0.009} \\ \midrule \midrule
\multicolumn{6}{c}{\textbf{Ensemble Model}} 
\\
\midrule
\gls{MCD} & 84.16 $\pm$ 0.370	& 0.033 $\pm$ 0.003 & 0.480 $\pm$ 0.003 & 0.641 $\pm$ 0.004 &	0.609 $\pm$ 0.004 \\
Deep Ensemble & {87.21 $\pm$ 0.630} &	\textbf{0.014 $\pm$ 0.006} & {0.373 $\pm$ 0.012} & {0.671 $\pm$ 0.005} & {0.640 $\pm$ 0.005} \\ 
SNGP Ensemble (Ours) & \textbf{88.19 $\pm$ 0.560} & 0.049 $\pm$ 0.004 &	\textbf{0.357 $\pm$ 0.012} &	\textbf{0.687 $\pm$ 0.008} &	\textbf{0.656 $\pm$ 0.007} \\
\bottomrule
\end{tabular}%
}
}
\caption{Results for Genomics sequence prediction and OOD detection, average over 10 seeds. Bold indicates the on-average best-performing methods among the single-model methods and the ensemble methods. }
\label{genomics}
\end{table}

The results are shown in Table \ref{genomics}. Comparing with all the other baseline models, SNGP model achieves the best OOD performance and best in-distribution accuracy, ECE, and NLL, among the single models. Each of the two components, \gls{SN} and \gls{GP} helps to reduce ECE and NLL, and improve OOD detection over the baseline DNN model. 
SNGP Ensemble is better than DNN Ensemble and MC Dropout in terms of in-distribution accuracy and NLL, and OOD detection.

\section{SNGP as a Building Block for Probabilistic Deep Learning}
\label{sec:building_block}

From a probabilistic machine learning perspective, the SNGP algorithm provides an efficient way to learn a single high-quality deterministic model $p_{\theta}(y|\bx)$ by, 1) improving the representation learning quality of the underlying neural network model through spectral normalization, and 2) introducing \textit{distance awareness} in conjunction with an approximate GP random feature layer. However, in the literature, there exist other competitive and state-of-the-art probabilistic methods that improve a deep learning model's predictive uncertainty via different approaches. 

The first class of such approaches is \textit{\textbf{ensembling}} that quantifies a model's uncertainty in the hidden layers by marginalizing over an (implicit) probabilistic distribution of model parameters. The most notable examples of this class is Deep Ensemble that averages parallel-trained deep learning models that are initialized from distinct random seeds \citep{lakshminarayanan_simple_2017,fort_deep_2019}, and also MC Dropout \citep{gal_dropout_2016} that samples from a functional-space model distribution that is generated by perturbing the model's dropout masks. As these approaches are uniquely capable of quantifying hidden-representation uncertainty, we hypothesize that they would synergize very well with SNGP, which is a single-model, last-layer-oriented approach. 

 \textit{\textbf{Data augmentation}} \citep{thulasidasan_mixup_2019, hendrycks_augmix_2020, cubuk2020randaugment, chen2020simple, chen2020improved} represents a second class of approach that improves the representation learning of neural networks by explicitly injecting expert knowledge {\color{black} about the types of surface-form perturbations that a model should be sensitive to or invariance against, thereby encouraging the model to learn a meaningful distance in its representation space.}
 Compared to the spectral normalization technique which provides a \textit{global} guarantee in distance preservation that applies to the entire input space (Proposition \ref{thm:resnet_lipschitz}), the data augmentation methods provides a \textit{local} guarantee 
 in the neighborhood of the training data. However, this guarantee is also expected to be \textit{stronger} (i.e., a better correspondence between the hidden-space distance $||.||_H$ and a suitable metric $d_X$ for the input data manifold) since we have explicitly instructed the model representation to be sensitive to or invariant against the semantically meaningful or meaningless directions as specified by human experts, respectively (see Section \ref{sec:discussion} for further discussion). Consequently, in the case where the coverage of the training data is sufficiently large, and a well-designed augmentation library is available for the task, data augmentation is expected to complement the spectral-normalization technique to provide a even stronger guarantee in distance-awareness.



In this section, we investigate the efficacy of the SNGP model as a building block for other uncertainty estimation techniques, and show that it leads to improvements that are complementary to those achieved by the other methods. Specifically, we consider the following methods from literature:

\begin{itemize}[leftmargin=2em]
    \item \textbf{Ensemble Methods}: 
    \begin{itemize}
        \item MC Dropout \citep{gal_dropout_2016} uses dropout regularization at test-time as a way to obtain samples from a predictive posterior distribution. SNGP can be easily combined with \gls{MCD} by enabling dropout at the inference time.
        \item Deep Ensemble \citep{lakshminarayanan_simple_2017} train multiple SNGP models with different random seed initializations and then average the predictions during test-time. 
    \end{itemize}
    \item \textbf{Data Augmentation} We consider AugMix \citep{hendrycks_augmix_2020}, a data augmentation algorithm that is known to improve the robustness and uncertainty estimation properties of a network by improving the learned representations of data through a combination of expert-designed data augmentations.
    It is easy to augment the SNGP training procedure by simply plugging in the Augmix augmentations into the data processing pipeline before feeding the data into the SNGP model \footnote{For ease of composability, we did not add the further consistency regularization through the Jensen-Shannon divergence as done in the original AugMix paper, as we do not notice further significant improvements on adding the loss on top of the Augmix augmentations.}.
\end{itemize}

\begin{table}[ht]
\centering
\scalebox{0.75}{%
\begin{tabular}{cccc}
\toprule
\multicolumn{1}{c|}{\multirow{2}{*}{Method}} & \multicolumn{3}{c}{CIFAR-10} \\
\multicolumn{1}{c|}{} & \multicolumn{1}{c|}{Acc / cAcc ($\uparrow$)} & \multicolumn{1}{c|}{ECE / cECE ($\downarrow$)} & \multicolumn{1}{c}{AUROC SVHN / CIFAR-100 ($\uparrow$)}  \\ \midrule

\multicolumn{1}{c|}{DNN} & \multicolumn{1}{c|}{95.8 $\pm$ 0.190 / 79.0 $\pm$ 0.350} & \multicolumn{1}{c|}{0.029 $\pm$ 0.002	/ 0.153 $\pm$ 0.005} & \multicolumn{1}{c}{0.946 $\pm$ 0.005 / 0.893 $\pm$ 0.001} \\
\multicolumn{1}{c|}{DNN-SN} & \multicolumn{1}{c|}{\textbf{96.0 $\pm$ 0.170} / 79.1 $\pm$ 0.290} & \multicolumn{1}{c|}{0.027 $\pm$ 0.002 / 0.150 $\pm$ 0.004} & \multicolumn{1}{c}{0.945 $\pm$ 0.005 / 0.894 $\pm$ 0.002} \\
\multicolumn{1}{c|}{DNN-GP} & \multicolumn{1}{c|}{95.8 $\pm$ 0.150	/ 79.1 $\pm$ 0.430} & \multicolumn{1}{c|}{\textbf{0.017 $\pm$ 0.003} / 0.100 $\pm$ 0.009} & \multicolumn{1}{c}{\textbf{0.964 $\pm$ 0.006 / 0.902 $\pm$ 0.002}} \\
\multicolumn{1}{c|}{SNGP} & \multicolumn{1}{c|}{95.7 $\pm$ 0.140 / \textbf{79.3 $\pm$ 0.340}} & \multicolumn{1}{c|}{\textbf{0.017 $\pm$ 0.003} / \textbf{0.099 $\pm$ 0.008}} & \multicolumn{1}{c}{0.960 $\pm$ 0.004 / \textbf{0.902 $\pm$ 0.003}}\\ \midrule\midrule

\multicolumn{1}{c|}{DNN + Dropout} & \multicolumn{1}{c|}{\textbf{95.7 $\pm$ 0.130} /	78.6 $\pm$ 0.430} & \multicolumn{1}{c|}{0.013 $\pm$ 0.002 / 0.099 $\pm$ 0.005} & \multicolumn{1}{c}{0.934 $\pm$ 0.004 / 0.903 $\pm$ 0.001}\\
\multicolumn{1}{c|}{DNN-SN + Dropout} & \multicolumn{1}{c|}{\textbf{95.7 $\pm$ 0.100 }/	78.6 $\pm$ 0.200} & \multicolumn{1}{c|}{0.014 $\pm$ 0.001 /	0.097 $\pm$ 0.003} & \multicolumn{1}{c}{0.935 $\pm$ 0.006 / 0.902 $\pm$ 0.001}\\
\multicolumn{1}{c|}{DNN-GP + Dropout} & \multicolumn{1}{c|}{\textbf{95.7 $\pm$ 0.140} /	78.8 $\pm$ 0.330} & \multicolumn{1}{c|}{\textbf{0.009 $\pm$ 0.002} /	\textbf{0.048 $\pm$ 0.004}} & \multicolumn{1}{c}{	\textbf{0.960 $\pm$ 0.006 / 0.909 $\pm$ 0.002}} \\
\multicolumn{1}{c|}{SNGP + Dropout} & \multicolumn{1}{c|}{95.5 $\pm$ 0.130 /	\textbf{78.9 $\pm$ 0.260}} & \multicolumn{1}{c|}{\textbf{0.009 $\pm$ 0.003}	/ 0.055 $\pm$ 0.007} & \multicolumn{1}{c}{0.953 $\pm$ 0.007 / 0.908 $\pm$ 0.002} \\ \midrule

\multicolumn{1}{c|}{DNN + Ensemble} & \multicolumn{1}{c|}{96.4 $\pm$ 0.090 /	80.4 $\pm$ 0.150} & \multicolumn{1}{c|}{0.011 $\pm$ 0.001	/ 0.092 $\pm$ 0.003} & \multicolumn{1}{c}{0.947 $\pm$ 0.002 / 0.914 $\pm$ 0.000}\\
\multicolumn{1}{c|}{DNN-SN + Ensemble} & \multicolumn{1}{c|}{\textbf{96.5 $\pm$ 0.050} /	80.5 $\pm$ 0.090 } & \multicolumn{1}{c|}{0.011 $\pm$ 0.001	/ 0.090 $\pm$ 0.002} & \multicolumn{1}{c}{0.952 $\pm$ 0.002 / 0.914 $\pm$ 0.001}\\
\multicolumn{1}{c|}{DNN-GP + Ensemble} & \multicolumn{1}{c|}{96.4 $\pm$ 0.060 /	80.8 $\pm$ 0.110} & \multicolumn{1}{c|}{0.009 $\pm$ 0.001 /	\textbf{0.041 $\pm$ 0.002}} & \multicolumn{1}{c}{\textbf{0.970 $\pm$ 0.002 / 0.920 $\pm$ 0.001}}\\
\multicolumn{1}{c|}{SNGP + Ensemble} & \multicolumn{1}{c|}{96.4 $\pm$ 0.040	/ \textbf{81.1 $\pm$ 0.130}} & \multicolumn{1}{c|}{\textbf{0.008 $\pm$ 0.001} / 0.042 $\pm$ 0.002} & \multicolumn{1}{c}{0.967 $\pm$ 0.002 / \textbf{0.920 $\pm$ 0.000}} \\ \midrule\midrule

\multicolumn{1}{c|}{DNN + AugMix} & \multicolumn{1}{c|}{\textbf{96.6 $\pm$ 0.150} /	\textbf{87.5 $\pm$ 0.003}} & \multicolumn{1}{c|}{0.014 $\pm$ 0.002 /	0.035 $\pm$ 0.002} & \multicolumn{1}{c}{0.958 $\pm$ 0.007 / 0.923 $\pm$ 0.002}\\
\multicolumn{1}{c|}{DNN-SN + AugMix} & \multicolumn{1}{c|}{\textbf{96.6 $\pm$ 0.110} /	\textbf{87.5 $\pm$ 0.300}} & \multicolumn{1}{c|}{0.013 $\pm$ 0.001 /	 0.035 $\pm$ 0.002} & \multicolumn{1}{c}{0.958 $\pm$ 0.005 / 0.923 $\pm$ 0.001}\\
\multicolumn{1}{c|}{DNN-GP + AugMix} & \multicolumn{1}{c|}{96.4 $\pm$ 0.240	/ 87.2 $\pm$ 0.330} & \multicolumn{1}{c|}{0.010 $\pm$ 0.002	/ 0.028 $\pm$ 0.004} &  \multicolumn{1}{c}{\textbf{0.978 $\pm$ 0.004 / 0.927 $\pm$ 0.002}} \\
\multicolumn{1}{c|}{SNGP + AugMix} & \multicolumn{1}{c|}{96.2 $\pm$ 0.200 /	 87.1 $\pm$ 0.330} & \multicolumn{1}{c|}{\textbf{0.009 $\pm$ 0.003}	/ \textbf{0.026 $\pm$ 0.005}} & \multicolumn{1}{c}{0.976 $\pm$ 0.005 / 0.925 $\pm$ 0.003} \\
\bottomrule
\end{tabular}%
}
\caption{
Results for CIFAR-10 when SNGP is used as a building block combined with MC Dropout, Ensemble, and Augmix. Best result in each class of method highlighted in bold.}
\label{tab:cifar10-abl}
\end{table}

\begin{table}[ht]
\centering
\scalebox{0.75}{%
\begin{tabular}{cccc}
\toprule
\multicolumn{1}{c|}{\multirow{2}{*}{Method}} & \multicolumn{3}{c}{CIFAR-100} \\
\multicolumn{1}{c|}{} & \multicolumn{1}{c|}{Acc / cAcc ($\uparrow$)} & \multicolumn{1}{c|}{ECE / cECE ($\downarrow$)} & \multicolumn{1}{c}{AUROC SVHN / CIFAR-10 ($\uparrow$)}  \\ \midrule

\multicolumn{1}{c|}{DNN} & \multicolumn{1}{c|}{80.4 $\pm$ 0.290	/ 55.0 $\pm$ 0.180} & \multicolumn{1}{c|}{0.107 $\pm$ 0.004 / 0.258 $\pm$ 0.004} & \multicolumn{1}{c}{0.799 $\pm$ 0.020 / 0.795 $\pm$ 0.001} \\
\multicolumn{1}{c|}{DNN-SN} & \multicolumn{1}{c|}{\textbf{80.5 $\pm$ 0.300} / 55.0 $\pm$ 0.210} & \multicolumn{1}{c|}{0.111 $\pm$ 0.002 / 0.268 $\pm$ 0.005} & \multicolumn{1}{c}{0.798 $\pm$ 0.022 / 0.793 $\pm$ 0.003 } \\
\multicolumn{1}{c|}{DNN-GP} & \multicolumn{1}{c|}{80.3 $\pm$ 0.400 / \textbf{55.3 $\pm$ 0.300} } & \multicolumn{1}{c|}{0.034 $\pm$ 0.005 /	\textbf{0.059 $\pm$ 0.002}} & \multicolumn{1}{c}{0.835 $\pm$ 0.021 / 0.797 $\pm$ 0.001}\\
\multicolumn{1}{c|}{SNGP} & \multicolumn{1}{c|}{80.3 $\pm$ 0.230 / \textbf{55.3 $\pm$ 0.190}} & \multicolumn{1}{c|}{\textbf{0.030 $\pm$ 0.004} / 0.060 $\pm$ 0.004} & \multicolumn{1}{c}{\textbf{0.846$\pm$ 0.019 / 0.798 $\pm$ 0.001}}\\ \midrule\midrule

\multicolumn{1}{c|}{DNN + Dropout} & \multicolumn{1}{c|}{80.2 $\pm$ 0.220 / 54.6 $\pm$ 0.160} & \multicolumn{1}{c|}{\textbf{0.031 $\pm$ 0.002} /	0.136 $\pm$ 0.005} & \multicolumn{1}{c}{0.800 $\pm$ 0.014 / 0.797 $\pm$ 0.002}\\
\multicolumn{1}{c|}{DNN-SN + Dropout} & \multicolumn{1}{c|}{\textbf{80.5 $\pm$ 0.340}	/ 54.6 $\pm$ 0.260} & \multicolumn{1}{c|}{0.035 $\pm$ 0.003 /	0.143 $\pm$ 0.004} & \multicolumn{1}{c}{0.781 $\pm$ 0.018 / 0.797 $\pm$ 0.001	} \\
\multicolumn{1}{c|}{DNN-GP + Dropout} & \multicolumn{1}{c|}{80.2 $\pm$ 0.360 / \textbf{54.9 $\pm$ 0.230}} & \multicolumn{1}{c|}{0.126 $\pm$ 0.007 /	\textbf{0.116 $\pm$ 0.004	}} & \multicolumn{1}{c}{0.825 $\pm$ 0.020 / 0.802 $\pm$ 0.003} \\
\multicolumn{1}{c|}{SNGP + Dropout} & \multicolumn{1}{c|}{80.3 $\pm$ 0.370 / \textbf{54.9 $\pm$ 0.240}} & \multicolumn{1}{c|}{0.128 $\pm$ 0.008 /	\textbf{0.116 $\pm$ 0.006}} & \multicolumn{1}{c}{\textbf{0.840 $\pm$ 0.021 / 0.804 $\pm$ 0.003}}\\ \midrule

\multicolumn{1}{c|}{DNN + Ensemble} & \multicolumn{1}{c|}{82.5 $\pm$ 0.190 / 57.6 $\pm$ 0.110} & \multicolumn{1}{c|}{0.041 $\pm$ 0.002 / 0.146 $\pm$ 0.003} & \multicolumn{1}{c}{0.812 $\pm$ 0.007 / 0.814 $\pm$ 0.001} \\
\multicolumn{1}{c|}{DNN-SN + Ensemble} & \multicolumn{1}{c|}{\textbf{82.7 $\pm$ 0.120} / 57.7 $\pm$ 0.120} & \multicolumn{1}{c|}{0.044 $\pm$ 0.002 / 0.151 $\pm$ 0.003} & \multicolumn{1}{c}{0.804 $\pm$ 0.005 / 0.814 $\pm$ 0.001} \\
\multicolumn{1}{c|}{DNN-GP + Ensemble} & \multicolumn{1}{c|}{82.4 $\pm$ 0.120 / \textbf{58.2 $\pm$ 0.110}} & \multicolumn{1}{c|}{0.030 $\pm$ 0.002 /	0.065 $\pm$ 0.001} & \multicolumn{1}{c}{0.828 $\pm$ 0.008 / 0.817 $\pm$ 0.001} \\
\multicolumn{1}{c|}{SNGP + Ensemble} & \multicolumn{1}{c|}{82.5 $\pm$ 0.160 /	\textbf{58.2 $\pm$ 0.130}} & \multicolumn{1}{c|}{\textbf{0.028 $\pm$ 0.002	/ 0.064 $\pm$ 0.001}} & \multicolumn{1}{c}{\textbf{0.838 $\pm$ 0.008 / 0.819 $\pm$ 0.001}} \\ \midrule\midrule

\multicolumn{1}{c|}{DNN + AugMix} & \multicolumn{1}{c|}{81.6 $\pm$ 0.003 /	\textbf{66.4 $\pm$ 0.280}} & \multicolumn{1}{c|}{0.082 $\pm$ 0.003	/ 0.131 $\pm$ 0.005} & \multicolumn{1}{c}{0.814 $\pm$ 0.025 / \textbf{0.798 $\pm$ 0.003}} \\
\multicolumn{1}{c|}{DNN-SN + AugMix} & \multicolumn{1}{c|}{81.9 $\pm$ 0.280 / \textbf{66.4 $\pm$ 0.240}} & \multicolumn{1}{c|}{0.080 $\pm$ 0.002	/ 0.133 $\pm$ 0.004} & \multicolumn{1}{c}{0.824 $\pm$ 0.021 / 0.796 $\pm$ 0.002} \\
\multicolumn{1}{c|}{DNN-GP + AugMix} & \multicolumn{1}{c|}{81.6 $\pm$ 0.350 /	66.2 $\pm$ 0.220	} & \multicolumn{1}{c|}{\textbf{0.042 $\pm$ 0.004}	/ 0.066 $\pm$ 0.002} & \multicolumn{1}{c}{0.855 $\pm$ 0.019 / 0.797 $\pm$ 0.002} \\
\multicolumn{1}{c|}{SNGP + AugMix} & \multicolumn{1}{c|}{81.6 $\pm$ 0.240	/ \textbf{66.4 $\pm$ 0.190}} & \multicolumn{1}{c|}{\textbf{0.042 $\pm$ 0.004 /	0.064 $\pm$ 0.002}} & \multicolumn{1}{c}{\textbf{0.870 $\pm$ 0.024 / 0.798 $\pm$ 0.001}} \\
\bottomrule
\end{tabular}%
}
\caption{
Results for CIFAR-100 when SNGP is used as a building block combined with MC Dropout, Ensemble, and Augmix. Best result in each class of method highlighted in bold.}
\label{tab:cifar100-abl}
\end{table}

\subsection{Ensembling Approaches for Hidden Representation Uncertainty}
\label{sec:building_block_ensemble}
We find that the SNGP approach complements extremely well with the ensemble methods in practice. From Tables~\ref{tab:cifar10-abl} and \ref{tab:cifar100-abl}, we can see that ensembling SNGP members either through MC Dropout or through Deep Ensemble results in improved performance on all metrics when compared to a single model, and in particular improves both the in-distribution calibration and the \gls{OOD} detection for both CIFAR-10 and CIFAR-100 datasets. Though there are some benefits to adding either spectral normalization or a last-layer GP separately to ensemble members, the combined effect of both parts of the SNGP algorithm consistently results in the best calibration and \gls{OOD} performance, especially in CIFAR-100 (which is a harder task). This behavior suggests that ensembling models does not inherently address the caveats in the representations of the underlying members, and further improvement can be obtained by imposing spectral normalization and the GP layer to the base models to improve their distance awareness property. To put it another way, SNGP offers an orthogonal improvement to the model's uncertainty quality that cannot be obtained from ensembling, and combining these two approaches can result in the best overall performance. Therefore, wherever resource permits, we suggest using an ensemble of SNGP base models to compound the benefits of \textit{distance awareness} and \textit{representation diversity} in improving the quality of uncertainty quantification (Section \ref{sec:method_landscape}). 
To conclude, even though ensembling different neural network instantiations improves epistemic uncertainty quantification by marginalizing from different points in the posterior, improving the quality of each member of the ensemble is crucial to further improve the predictive uncertainty.

\subsection{Data Augmentation for Improved Distance Awareness}
\label{sec:building_block_data_aug}

As shown in the introduction (Figure \ref{fig:sngp_overview}), applying spectral normalization to residual neural networks encourages the model representation to be distance preserving
and prevents it from feature collapse. Interestingly, this overlaps with the design goal of another well-known uncertainty technique: \textit{data augmentation} (e.g., AugMix \citep{hendrycks_augmix_2020}). Specifically, data augmentation improves the distance preservation ability of the representation by forcing the model to be invariant against the semantic-preserving perturbations (e.g., rotating an image), and also being sensitive to semantic-modifying perturbations (e.g., adding a word ``not" to a natural language sentence). To empirically investigate the composability of these two techniques, we perform ablation experiments of training SNGP models on the CIFAR-10 and CIFAR-100 datasets with AugMix, and show that the improvements in representations learnt with AugMix complement SNGP. Results are shown in Table \ref{tab:cifar10-abl}-\ref{tab:cifar100-abl}. If data augmentation improves distance awareness, we would expect to see DNN-GP + AugMix significantly improve over DNN-GP; this is indeed the case, which confirms our hypothesis that increasing distance awareness (either through smoothness or data augmentation) improve DNN-GP. 
It is interesting to observe that in this task, while the SNGP + AugMix on average outperforms its ablated counterparts, the gap between DNN-GP + AugMix  vs. SNGP + AugMix is lower than DNN-GP vs. \gls{SNGP}, illustrating the effectiveness of AugMix in improving model uncertainty via enhancing the distance awareness.
Therefore, for domains where a suite of well-designed augmentation is available for the task, and when the test data is in a neighborhood of the augmented training examples (which is the case for CIFAR-C), the data augmentation method can provide a strong guarantee in preserving a meaningful distance in the input space. 

In summary, these experiments illustrate the importance of the representation's distance-awareness quality in improving the model's uncertainty quality. Consequently, for datasets and modalities where a well-designed augmentation library is available (e.g., AugMix for image modality), it is advantageous to complement SNGP with the data augmentation approaches to obtain a stronger guarantee in preserving a semantically meaningful distance in its representation space. However, when it is difficult to obtain high-quality augmentations (e.g. the genomics example), or it is computationally intractable to sufficiently augment the training data, the SNGP alone provides a cheap, modality-invariant approach to improve trained representations that provides a \textit{global} guarantee for distance preservation 
(see Section \ref{sec:discussion} for further discussion).

\section{Conclusions and Discussion}
\label{sec:conclusion}

We propose SNGP, a simple approach to improve the predictive uncertainty estimation of a single deterministic \gls{DNN}. It makes minimal changes to the architecture and training/prediction pipeline of a deterministic \gls{DNN}, only adding spectral normalization to the hidden mapping, and replacing the dense output layer with a random feature layer that approximates a \gls{GP}. We theoretically motivate \textit{distance awareness}, the key design principle behind \gls{SNGP}, via a decision-theoretic analysis of the uncertainty estimation problem. We also propose a closed-form approximation method to make the GP posterior end-to-end trainable in linear time with the rest of the neural network. On a suite of vision and language understanding tasks and on modern architectures (ResNet and BERT), 
\gls{SNGP} is competitive 
in prediction, calibration and out-of-domain detection, outperforms other single-model approaches, and combines well with other state-of-the-art uncertainty techniques. 

\subsection{Further Discussions}
\label{sec:discussion}

{\color{black} 
A central goal of this work is to provide theoretical and empirical evidences for the importance of incorporating \textit{distance awareness} (i.e., the distance of a test example from the training data) into a model's uncertainty estimate. This provides a complementary view to the classic approaches to deep learning uncertainty, where the model uncertainty is primarily quantified by a test example's distance from the decision boundary (e.g., Figure \ref{fig:sngp_overview}). Indeed, both the \textit{distance to training data} and the \textit{distance to decision boundary} are reasonable quantifiers of model uncertainty, and should be treated as equally important components of a practitioner's uncertainty quantification toolbox. In practice, the choice between the two distances should be made based on the nature of the data generating mechanism and the optimality criteria that the practitioner wish to pursue. For example, under the classic i.i.d. assumption where the test examples always stays in-domain (i.e., identically distributed as the training data) and one wish to use model uncertainty to detect ambiguous examples, then \textit{distance to the decision boundary} is a suitable choice. On the other hand, in a safe-critical setting where it is important to guard against worst-case risk (i.e., \Cref{sec:minimax}) in the presence of likely distributional shift, \textit{distance to the training data} would be a more appropriate choice. Alternatively, one may consider selecting a uncertainty metric that incorporates both types of distances. For example, as revealed by \Cref{eq:mean_field_approx}, the posterior predictive mean of SNGP in fact incorporates both the magnitude of the predictive logit (i.e., distance to the decision boundary) and predictive variance (i.e., distance to the training data).
}

Furthermore, an important observation we made in this work is that \textit{learning smooth  representations  is important for good uncertainty quantification}. In particular, we highlighted  \textit{bi-Lipschitz} (Equation (\ref{eq:dp})) as an important condition for the learned representation of a \gls{DNN} to attain high-quality uncertainty performance. We proposed spectral normalization as a simple approach to ensure such property in practice, and illustrated its practical effectiveness across a wide range of data modalities. Perhaps surprisingly, the improvement is observed even in the high-dimensional regime (e.g., image and text), where the true ``semantic" distance seemingly diverges from that based on the surface-form representation (Section \ref{sec:experiments}), and the benefit of SNGP seems to be not fully overlapping with those provided by the other state-of-the-art uncertainty techniques (Section \ref{sec:building_block}). To this end, we find it theoretically relevant to initiate a discussion about the role that the bi-Lipschitz condition may play in high-dimensional and overparameterized learning, and how it  compares with those of the other state-of-the-art approaches. The following discussion is in no way comprehensive, and is intended to serve as potential starting points for future work.

\paragraph{Preservation of ``semantic" distance.} At a first glance, the technique proposed in Section \ref{sec:sn} seems to act on a naive, surface-level distance in the input space (e.g., $L_2$ distance in the pixel space), hence raising the question whether the ``semantic" distance $d_X$ {\color{black} (i.e., a meaningful distance metric that is appropriate for the input data manifold)}
is being preserved as well. Indeed, in common machine learning applications, the data points reside on an underlying low-dimensional manifold behind their high-dimensional surface representation \footnote{
{\color{black}\Cref{sec:semantic_distance_formalization} provides an example mathematical  formalization that further elaborates this idea.}}. Here, the correspondence between the {\color{black} true manifold distance $d_X$ (i.e., the ``semantic distance")} and the surface-level distance (e.g., $L_2$ distance in the pixel space) is complex and dynamic, and can be described by the concept of \textit{metric distortion} from metric embedding theory \citep{abraham2011advances, matouvsek2013lecture, chennuru2018measures}. Formally, consider $m: (\Xsc, d_S) \rightarrow (\Xsc, d_X)$ a mapping between the metric space $(\Xsc, d_S)$ equipped with surface-level distance to that equipped with the {\color{black} true} distance $(\Xsc, d_X)$. Then,  $d_X(\bx_1, \bx_2)$ can be understood as a \textit{distorted} version of the surface-level distance $d_S(\bx_1, \bx_2)$ as implemented by $m$, with the type and degree of the distortion changes depending on the location in the product feature space $(\bx_1, \bx_2) \in \Xsc \times \Xsc$. More specifically, defining $\rho(\bx_1, \bx_2) = \frac{d_X(\bx_1, \bx_2)}{d_S(\bx_1, \bx_2)}$ the distance ratio that measures the degree of local distortion at $(\bx_1, \bx_2)$, two types of distortions may occur:

\begin{definition}[Distance Distortion] 
$ $

\begin{itemize}[leftmargin=2em]
\item \textbf{(Distance Contraction)} $d_X(\bx_1, \bx_2) < d_S(\bx_1, \bx_2)$, such that the magnitude of $\rho(x_1, x_2)$ is high.
\item \textbf{(Distance Expansion)} $d_X(\bx_1, \bx_2) > d_S(\bx_1, \bx_2)$, such that the magnitude of $1/\rho(x_1, x_2)$ is high.
\end{itemize}
\label{def:distance_distortion}
\end{definition}

The first scenario (distance contraction) is common in both image and text modalities, where the underlying content of an image / text is invariant to the noisy movements in the pixel / token space. The second scenario (distance expansion) occurs often in language understanding, where the addition of a single token (e.g., ``not") drastically changes the meaning of a sentence. Furthermore, the type and degree of distortion is location- and direction-dependent. For example, in the language domain, a sentence's meaning is generally invariant to its surface-level syntactic forms (i.e., distance contraction), however some sentences can be drastically changed by the modification of a few key tokens (i.e., distance expansion).

Interestingly, despite this complicated and dynamic distortion of distance, it is possible to show that, as long as the ``semantic" distance $d_X$ is well defined (i.e., $0 \leq d_X(\bx_1, \bx_2) < \infty, \forall \bx_1, \bx_2 \in \Xsc$), the global bound on the surface-level distance still provides a basic guarantee in preserving the local true distances at every $(\bx_1, \bx_2)$ (albeit at a looser degree):

\begin{proposition}[Preserving ``semantic" distance under metric distortion]

Assume that the mapping $h: \Xsc \rightarrow \Hsc$ preserves the surface-form distance $d_S$, i.e., there exist non-negative constants $L_1 < L_2$ such that:
\begin{align}
L_1 \times d_S(x_1, x_2)  \leq ||h(x_1) - h(x_2)||_H  \leq L_2 \times d_S(x_1, x_2),
\label{eq:surface_level_bound}
\end{align}
then, for a well-defined surface-form distance $d_S \in [0, \infty)$ that is a distortion of $d_X$, we have:

\begin{itemize}[leftmargin=2em]
\item \textbf{(Distance Contraction)} For a local region $(\bx_1, \bx_2)$ where the distance contraction occurs with magnitude $\rho(\bx_1, \bx_2) = l'$, there exists a $L'_2$ such that $L'_2 > l' * L_2$ and
\begin{align*}
L_1 \times d_X(x_1, x_2)  \leq ||h(x_1) - h(x_2)||_H  \leq L_2 \times d_S(x_1, x_2) \leq L'_2 \times d_X(x_1, x_2),
\end{align*}

\item \textbf{(Distance Expansion)} For a local region $(\bx_1, \bx_2)$ where the distance expansion occurs with magnitude $1/\rho(\bx_1, \bx_2) = l'$, there exists a $L'_1$ such that $L'_1 \leq L_1 / l'$ and
\begin{align*}
L'_1 \times d_X(x_1, x_2)  \leq L_1 \times d_S(x_1, x_2)  \leq ||h(x_1) - h(x_2)||_H  \leq L_2 \times d_X(x_1, x_2).
\end{align*}
\end{itemize}

\label{thm:semantic_distance_preservation}
\end{proposition}
As a result, by bounding the neural network model from warping the representation space distance $||.||_H$ to an extreme degree, the Lipschitz bound (\ref{eq:surface_level_bound}) provides a basic guarantee in the preservation of the {\color{black} ``semantic" distance that is appropriate for the data manifold}.
Consequently, the SNGP approach concretely improves the unregularized training of the overparameterized network by helping it to reach a better balance between \textit{dimension reduction} and \textit{information preservation}, which we discuss the next.

\paragraph{Balancing the information tradeoff in high-dimensional learning.}
As the information of machine learning data tends to concentrate on a low-dimensional manifold, a high-dimensional learning model is often confronted with a tradeoff between the need of \textit{dimension reduction} and the need of \textit{information preservation}: dimension reduction is necessary for reducing the statistical complexity of the learning problem and for ensuring superior generalization under finite data. However, an excessive reduction of information leads the model to ignore semantically meaningful features that are less correlated with the training label, creating a challenge for uncertainty quantification.

In the context of modern overparameterized networks, naive training without regularization is often observed to lead to one extreme end of this tradeoff (i.e., excessive dimension reduction). Specifically, by minimizing the training cross entropy toward zero, the model pushes its logits $logit(\bx) = h(\bx)^T\bbeta$ to a large magnitude at the location of training labels $y$, leading to a high alignment between training logits and the corresponding labels. Viewing from the representation space $h(\bx) \in H$, this manifests into the training data representations $h(x)$ being pushed far away from the decision boundary $\bbeta$, which is often achieved by stretching the hidden-space distance $||.||_H$ along the task-relevant directions (i.e., those orthogonal to the decision boundaries) to an extreme degree. This essentially leads to a warped hidden-space geometry with extremely high weighting on a handful of task-relevant principal directions (see, e.g., Figure \ref{fig:sngp_overview} top), implying low effective degrees of freedom \citep{papyan2020traces, NIPS2017_0ebcc77d}. Although not detrimental to the in-domain generalization, this warped geometry leads the model to be overly sensitive to the hidden-space movements along the label-relevant direction, while being insensitive to semantic change in the directions that are less correlated with the training labels, thereby creating challenges for uncertainty tasks such as out-of-domain detection \citep{hein2019relu}.

To this end, the SNGP approach helps the model to strike a better balance between dimension reduction v.s. information preservation by modulating the degree of geometry distortion via the Lipschitz bound. As a result, the model is still able to disentangle input features through a cascade of layer-wise transformations, but is protected from extreme distortion in $H$ to a degree that causes feature collapse (see, e.g., Figure \ref{fig:2d_exp} bottom). As a result, the model is still capable of learning abstract and task-relevant features by exploiting the expressive power of overparameterization, while not blind to semantically meaningful movements that are less correlated with training labels, thereby leading to a better balance between generalization and uncertainty performance.

\paragraph{Alternative approaches, limitations, and future work.} 

It is worth noting that there exist other representation learning techniques, e.g., data augmentation, constrastive learning or unsupervised pretraining, that are known to also improve a network's uncertainty performance \citep{hendrycks_using_2019, hendrycks_augmix_2020}. From the perspective of distance awareness, these methods help the model representation to preserve a semantically meaningful distance by injecting external knowledge into the learning pipeline. For example, data augmentation and constrastive learning instructs the model to learn distance contraction or expansion by creating semantically similar or dissimilar pairs $(\bx_1, \bx_2)$ using expert knowledge, so that the \gls{DNN} representation is invariant against the semantic-preserving perturbations (e.g., rotating an image), and is sensitive to semantic-modifying perturbations (e.g., adding a word ``not" to a natural language sentence). On the other hand, pre-training injects the model with a prior about the similarity or dissimilarity between examples that was learned (e.g., via contrastive masked language modeling) from a large external corpus. Compared to these data-intensive approaches, spectral normalization provides a global guarantee in distance preservation and does not require external resources such as pretraining datasets or an understanding of which augmentations are semantically meaningful and useful. However, the guarantee is also looser since it only places a uniform upper and lower bound on $||.||_H$ along all directions of the perturbation, rather than explicitly training $||.||_H$ to be contractive or expansive with respect to the direction of semantic perturbation. Interestingly, as shown in Sections \ref{sec:experiments} - \ref{sec:building_block}, SNGP combines well with these additional techniques, indicating that the benefits they provide to the model do not conflict with each other.

Finally, we note that the Gaussian process algorithm presented this work is optimized for practicality (i.e., scalability to extremely large datasets, and seamless integration into minibatch SGD-based neural training pipeline). Such design goal invariably necessitates several approximations for the exact Gaussian process posterior: (1) random-feature expansion for the kernel function, (2) Laplace approximation for the posterior variance, and (3) mean-field approximation for the softmax posterior mean. In our preliminary experiments, we found that a higher-quality approximation to the kernel function (e.g., using orthogonal random feature \citep{yu_orthogonal_2016} rather than random Fourier feature) and an near-exact Monte Carlo approximation to the softmax posterior mean did not lead to a meaningful improvement to model performance. However, this does not preclude the theoretical possibility that an exact Gaussian process posterior can further improve uncertainty quality when compared to the current algorithm choice. Therefore, identifying GP algorithms that can better implement the distance-awareness principle without sacrificing practicality is an important direction for future work.

\clearpage

\acks{
We 
would like to 
thank Rodolphe Jenatton, D. Sculley, Kevin Murphy, Deepak Ramachandran for the insightful comments and fruitful discussions.} 

\vskip 0.2in
\bibliography{references}

\begin{thebibliography}{160}
\providecommand{\natexlab}[1]{#1}
\providecommand{\url}[1]{\texttt{#1}}
\expandafter\ifx\csname urlstyle\endcsname\relax
  \providecommand{\doi}[1]{doi: #1}\else
  \providecommand{\doi}{doi: \begingroup \urlstyle{rm}\Url}\fi

\bibitem[Abraham et~al.(2011)Abraham, Bartal, and Neiman]{abraham2011advances}
Ittai Abraham, Yair Bartal, and Ofer Neiman.
\newblock Advances in metric embedding theory.
\newblock \emph{Advances in Mathematics}, 228\penalty0 (6):\penalty0
  3026--3126, 2011.

\bibitem[Amodei et~al.(2016)Amodei, Olah, Steinhardt, Christiano, Schulman, and
  Man{\'e}]{amodei_concrete_2016}
Dario Amodei, Chris Olah, Jacob Steinhardt, Paul Christiano, John Schulman, and
  Dan Man{\'e}.
\newblock Concrete {Problems} in {AI} {Safety}.
\newblock \emph{arXiv:1606.06565 [cs]}, June 2016.

\bibitem[An et~al.(2015)An, Boussaid, and Bennamoun]{an_how_2015}
Senjian An, Farid Boussaid, and Mohammed Bennamoun.
\newblock How {Can} {Deep} {Rectifier} {Networks} {Achieve} {Linear}
  {Separability} and {Preserve} {Distances}?
\newblock In \emph{International {Conference} on {Machine} {Learning}}, pages
  514--523, June 2015.
\newblock ISSN: 1938-7228 Section: Machine Learning.

\bibitem[Anil et~al.(2019)Anil, Lucas, and Grosse]{anil_sorting_2019}
Cem Anil, James Lucas, and Roger Grosse.
\newblock Sorting {Out} {Lipschitz} {Function} {Approximation}.
\newblock In \emph{International {Conference} on {Machine} {Learning}}, pages
  291--301, May 2019.
\newblock ISSN: 1938-7228 Section: Machine Learning.

\bibitem[Bach(2017)]{bach_breaking_2017}
Francis Bach.
\newblock Breaking the {Curse} of {Dimensionality} with {Convex} {Neural}
  {Networks}.
\newblock \emph{Journal of Machine Learning Research}, 18\penalty0
  (19):\penalty0 1--53, 2017.
\newblock ISSN 1533-7928.

\bibitem[Bartlett et~al.(2018)Bartlett, Evans, and
  Long]{bartlett_representing_2018}
Peter Bartlett, Steven Evans, and Phil Long.
\newblock Representing smooth functions as compositions of near-identity
  functions with implications for deep network optimization.
\newblock \emph{arXiv}, 2018.

\bibitem[Behrmann et~al.(2019)Behrmann, Grathwohl, Chen, Duvenaud, and
  Jacobsen]{behrmann_invertible_2019}
Jens Behrmann, Will Grathwohl, Ricky T.~Q. Chen, David Duvenaud, and
  Joern-Henrik Jacobsen.
\newblock Invertible {Residual} {Networks}.
\newblock In \emph{International {Conference} on {Machine} {Learning}}, pages
  573--582, May 2019.
\newblock ISSN: 1938-7228 Section: Machine Learning.

\bibitem[Behrmann et~al.(2021)Behrmann, Vicol, Wang, Grosse, and
  Jacobsen]{behrmann2021understanding}
Jens Behrmann, Paul Vicol, Kuan-Chieh Wang, Roger Grosse, and Joern-Henrik
  Jacobsen.
\newblock Understanding and mitigating exploding inverses in invertible neural
  networks.
\newblock In \emph{International Conference on Artificial Intelligence and
  Statistics}, pages 1792--1800. PMLR, 2021.

\bibitem[Bendale and Boult(2016)]{bendale_towards_2016}
Abhijit Bendale and Terrance~E. Boult.
\newblock Towards {Open} {Set} {Deep} {Networks}.
\newblock \emph{2016 IEEE Conference on Computer Vision and Pattern Recognition
  (CVPR)}, 2016.

\bibitem[Berger(1985)]{berger_statistical_1985}
James~O. Berger.
\newblock \emph{Statistical {Decision} {Theory} and {Bayesian} {Analysis}}.
\newblock Springer {Series} in {Statistics}. Springer-Verlag, New York, 2
  edition, 1985.
\newblock ISBN 978-0-387-96098-2.

\bibitem[Bishop(2011)]{bishop_pattern_2011}
Christopher~M. Bishop.
\newblock \emph{Pattern {Recognition} and {Machine} {Learning}}.
\newblock Springer, New York, April 2011.
\newblock ISBN 978-0-387-31073-2.

\bibitem[Blum(2006)]{blum_random_2006}
Avrim Blum.
\newblock Random {Projection}, {Margins}, {Kernels}, and {Feature}-{Selection}.
\newblock In \emph{Subspace, {Latent} {Structure} and {Feature} {Selection}},
  Lecture {Notes} in {Computer} {Science}, pages 52--68, Berlin, Heidelberg,
  2006. Springer.
\newblock ISBN 978-3-540-34138-3.
\newblock \doi{10.1007/11752790_3}.

\bibitem[Blundell et~al.(2010)Blundell, Teh, and
  Heller]{blundell_bayesian_2012}
Charles Blundell, Yee~Whye Teh, and Katherine~A Heller.
\newblock Bayesian rose trees.
\newblock In \emph{Proceedings of the Twenty-Sixth Conference on Uncertainty in
  Artificial Intelligence}, pages 65--72, 2010.

\bibitem[Bradshaw et~al.(2017)Bradshaw, Matthews, and
  Ghahramani]{bradshaw_adversarial_2017}
John Bradshaw, Alexander G. de~G. Matthews, and Zoubin Ghahramani.
\newblock Adversarial {Examples}, {Uncertainty}, and {Transfer} {Testing}
  {Robustness} in {Gaussian} {Process} {Hybrid} {Deep} {Networks}.
\newblock \emph{arXiv:1707.02476 [stat]}, July 2017.
\newblock arXiv: 1707.02476.

\bibitem[Br{\"o}cker(2009)]{brocker2009reliability}
Jochen Br{\"o}cker.
\newblock Reliability, sufficiency, and the decomposition of proper scores.
\newblock \emph{Quarterly Journal of the Royal Meteorological Society: A
  journal of the atmospheric sciences, applied meteorology and physical
  oceanography}, 135\penalty0 (643):\penalty0 1512--1519, 2009.

\bibitem[Calandra et~al.(2016)Calandra, Peters, Rasmussen, and
  Deisenroth]{calandra_manifold_2016}
Roberto Calandra, Jan Peters, Carl~E. Rasmussen, and Marc~Peter Deisenroth.
\newblock Manifold {Gaussian} {Processes} for regression.
\newblock \emph{2016 International Joint Conference on Neural Networks
  (IJCNN)}, 2016.
\newblock \doi{10.1109/IJCNN.2016.7727626}.

\bibitem[Cer et~al.(2017)Cer, Diab, Agirre, Lopez-Gazpio, and
  Specia]{cer_semeval-2017_2017}
Daniel Cer, Mona Diab, Eneko Agirre, Inigo Lopez-Gazpio, and Lucia Specia.
\newblock {SemEval}-2017 {Task} 1: {Semantic} {Textual} {Similarity}
  {Multilingual} and {Crosslingual} {Focused} {Evaluation}.
\newblock In \emph{Proceedings of the 11th {International} {Workshop} on
  {Semantic} {Evaluation} ({SemEval}-2017)}, pages 1--14, Vancouver, Canada,
  August 2017. Association for Computational Linguistics.
\newblock \doi{10.18653/v1/S17-2001}.

\bibitem[Cer et~al.(2018)Cer, Yang, Kong, Hua, Limtiaco, John, Constant,
  Guajardo-Cespedes, Yuan, Tar, et~al.]{cer2018universal}
Daniel Cer, Yinfei Yang, Sheng-yi Kong, Nan Hua, Nicole Limtiaco, Rhomni~St
  John, Noah Constant, Mario Guajardo-Cespedes, Steve Yuan, Chris Tar, et~al.
\newblock Universal sentence encoder for english.
\newblock In \emph{Proceedings of the 2018 conference on empirical methods in
  natural language processing: system demonstrations}, pages 169--174, 2018.

\bibitem[Chandrasekaran and Mago(2021)]{chandrasekaran2021evolution}
Dhivya Chandrasekaran and Vijay Mago.
\newblock Evolution of semantic similarity—a survey.
\newblock \emph{ACM Computing Surveys (CSUR)}, 54\penalty0 (2):\penalty0 1--37,
  2021.

\bibitem[Chen et~al.(2020{\natexlab{a}})Chen, Kornblith, Norouzi, and
  Hinton]{chen2020simple}
Ting Chen, Simon Kornblith, Mohammad Norouzi, and Geoffrey Hinton.
\newblock A simple framework for contrastive learning of visual
  representations.
\newblock In \emph{International conference on machine learning}, pages
  1597--1607. PMLR, 2020{\natexlab{a}}.

\bibitem[Chen et~al.(2020{\natexlab{b}})Chen, Fan, Girshick, and
  He]{chen2020improved}
Xinlei Chen, Haoqi Fan, Ross Girshick, and Kaiming He.
\newblock Improved baselines with momentum contrastive learning.
\newblock \emph{arXiv preprint arXiv:2003.04297}, 2020{\natexlab{b}}.

\bibitem[Chennuru~Vankadara and von Luxburg(2018)]{chennuru2018measures}
Leena Chennuru~Vankadara and Ulrike von Luxburg.
\newblock Measures of distortion for machine learning.
\newblock \emph{Advances in Neural Information Processing Systems}, 31, 2018.

\bibitem[Chernodub and Nowicki(2017)]{chernodub_norm-preserving_2017}
Artem Chernodub and Dimitri Nowicki.
\newblock Norm-preserving {Orthogonal} {Permutation} {Linear} {Unit}
  {Activation} {Functions} ({OPLU}).
\newblock \emph{arXiv:1604.02313 [cs]}, January 2017.
\newblock arXiv: 1604.02313.

\bibitem[Choromanski et~al.(2018)Choromanski, Rowland, Sarlos, Sindhwani,
  Turner, and Weller]{choromanski_geometry_2018}
Krzysztof Choromanski, Mark Rowland, Tamas Sarlos, Vikas Sindhwani, Richard
  Turner, and Adrian Weller.
\newblock The {Geometry} of {Random} {Features}.
\newblock In \emph{International {Conference} on {Artificial} {Intelligence}
  and {Statistics}}, pages 1--9, March 2018.

\bibitem[Cubuk et~al.(2020)Cubuk, Zoph, Shlens, and Le]{cubuk2020randaugment}
Ekin~D Cubuk, Barret Zoph, Jonathon Shlens, and Quoc~V Le.
\newblock Randaugment: Practical automated data augmentation with a reduced
  search space.
\newblock In \emph{Proceedings of the IEEE/CVF Conference on Computer Vision
  and Pattern Recognition Workshops}, pages 702--703, 2020.

\bibitem[Cui et~al.(2021)Cui, Deng, Hu, and Zhu]{cui2021accurate}
Peng Cui, Zhijie Deng, Wenbo Hu, and Jun Zhu.
\newblock Accurate and reliable forecasting using stochastic differential
  equations.
\newblock \emph{arXiv preprint arXiv:2103.15041}, 2021.

\bibitem[Damianou and Lawrence(2013)]{damianou_deep_2013}
Andreas Damianou and Neil Lawrence.
\newblock Deep {Gaussian} {Processes}.
\newblock In \emph{Artificial {Intelligence} and {Statistics}}, pages 207--215,
  April 2013.

\bibitem[Daunizeau(2017)]{daunizeau_semi-analytical_2017}
Jean Daunizeau.
\newblock Semi-analytical approximations to statistical moments of sigmoid and
  softmax mappings of normal variables.
\newblock \emph{arXiv preprint arXiv:1703.00091}, 2017.

\bibitem[Daxberger et~al.(2021)Daxberger, Kristiadi, Immer, Eschenhagen, Bauer,
  and Hennig]{daxberger2021laplace}
Erik Daxberger, Agustinus Kristiadi, Alexander Immer, Runa Eschenhagen,
  Matthias Bauer, and Philipp Hennig.
\newblock Laplace redux-effortless {B}ayesian deep learning.
\newblock \emph{Advances in Neural Information Processing Systems}, 34, 2021.

\bibitem[DeGroot and Schervish(2012)]{degroot2012probability}
Morris~H DeGroot and Mark~J Schervish.
\newblock \emph{Probability and statistics}.
\newblock Pearson Education, 2012.

\bibitem[Dehaene(2019)]{dehaene_deterministic_2019}
Guillaume~P. Dehaene.
\newblock A deterministic and computable {Bernstein}-von {Mises} theorem.
\newblock \emph{ArXiv}, 2019.

\bibitem[Denker and LeCun(1991)]{denker_transforming_1991}
John~S. Denker and Yann LeCun.
\newblock Transforming {Neural}-{Net} {Output} {Levels} to {Probability}
  {Distributions}.
\newblock In \emph{Advances in {Neural} {Information} {Processing} {Systems}
  3}, pages 853--859. Morgan-Kaufmann, 1991.

\bibitem[Deselaers and Ferrari(2011)]{deselaers2011visual}
Thomas Deselaers and Vittorio Ferrari.
\newblock Visual and semantic similarity in imagenet.
\newblock In \emph{CVPR 2011}, pages 1777--1784. IEEE, 2011.

\bibitem[Dinh et~al.(2014)Dinh, Krueger, and Bengio]{dinh_nice:_2014}
Laurent Dinh, David Krueger, and Yoshua Bengio.
\newblock {NICE}: {Non}-linear {Independent} {Components} {Estimation}.
\newblock \emph{arXiv:1410.8516 [cs]}, October 2014.
\newblock arXiv: 1410.8516.

\bibitem[Dinh et~al.(2016)Dinh, Sohl-Dickstein, and Bengio]{dinh_density_2016}
Laurent Dinh, Jascha Sohl-Dickstein, and Samy Bengio.
\newblock Density estimation using {Real} {NVP}.
\newblock \emph{arXiv:1605.08803 [cs, stat]}, May 2016.
\newblock arXiv: 1605.08803.

\bibitem[Dusenberry et~al.(2020)Dusenberry, Jerfel, Wen, Ma, Snoek, Heller,
  Lakshminarayanan, and Tran]{dusenberry_efficient_2020}
Mike Dusenberry, Ghassen Jerfel, Yeming Wen, Yian Ma, Jasper Snoek, Katherine
  Heller, Balaji Lakshminarayanan, and Dustin Tran.
\newblock Efficient and {Scalable} {Bayesian} {Neural} {Nets} with {Rank}-1
  {Factors}.
\newblock \emph{Proceedings of the International Conference on Machine
  Learning}, 1, 2020.

\bibitem[Dutordoir et~al.(2020)Dutordoir, Durrande, and
  Hensman]{dutordoir2020sparse}
Vincent Dutordoir, Nicolas Durrande, and James Hensman.
\newblock Sparse gaussian processes with spherical harmonic features.
\newblock In \emph{International Conference on Machine Learning}, pages
  2793--2802. PMLR, 2020.

\bibitem[Eschenhagen et~al.(2021)Eschenhagen, Daxberger, Hennig, and
  Kristiadi]{DBLP:journals/corr/abs-2111-03577}
Runa Eschenhagen, Erik Daxberger, Philipp Hennig, and Agustinus Kristiadi.
\newblock Mixtures of laplace approximations for improved post-hoc uncertainty
  in deep learning.
\newblock \emph{CoRR}, abs/2111.03577, 2021.

\bibitem[Fort et~al.(2019)Fort, Hu, and Lakshminarayanan]{fort_deep_2019}
Stanislav Fort, Huiyi Hu, and Balaji Lakshminarayanan.
\newblock Deep {Ensembles}: {A} {Loss} {Landscape} {Perspective}.
\newblock \emph{arXiv:1912.02757 [cs, stat]}, December 2019.
\newblock arXiv: 1912.02757.

\bibitem[Freedman(1999)]{freedman_wald_1999}
David Freedman.
\newblock Wald {Lecture}: {On} the {Bernstein}-von {Mises} theorem with
  infinite-dimensional parameters.
\newblock \emph{The Annals of Statistics}, 27\penalty0 (4):\penalty0
  1119--1141, August 1999.
\newblock ISSN 0090-5364, 2168-8966.
\newblock \doi{10.1214/aos/1017938917}.

\bibitem[Gal and Ghahramani(2016)]{gal_dropout_2016}
Yarin Gal and Zoubin Ghahramani.
\newblock Dropout {As} a {Bayesian} {Approximation}: {Representing} {Model}
  {Uncertainty} in {Deep} {Learning}.
\newblock In \emph{Proceedings of the 33rd {International} {Conference} on
  {International} {Conference} on {Machine} {Learning} - {Volume} 48},
  {ICML}'16, pages 1050--1059, New York, NY, USA, 2016. JMLR.org.

\bibitem[Gelman et~al.(2013)Gelman, Carlin, Stern, Dunson, Vehtari, and
  Rubin]{gelman_bayesian_2013}
Andrew Gelman, John~B. Carlin, Hal~S. Stern, David~B. Dunson, Aki Vehtari, and
  Donald~B. Rubin.
\newblock \emph{Bayesian {Data} {Analysis}}.
\newblock Chapman and Hall/CRC, Boca Raton, 3 edition edition, November 2013.
\newblock ISBN 978-1-4398-4095-5.

\bibitem[Gneiting and Raftery(2007)]{gneiting_strictly_2007}
Tilmann Gneiting and Adrian~E Raftery.
\newblock Strictly {Proper} {Scoring} {Rules}, {Prediction}, and {Estimation}.
\newblock \emph{Journal of the American Statistical Association}, 102\penalty0
  (477):\penalty0 359--378, March 2007.
\newblock ISSN 0162-1459.
\newblock \doi{10.1198/016214506000001437}.

\bibitem[Gneiting et~al.(2007)Gneiting, Balabdaoui, and
  Raftery]{gneiting_probabilistic_2007}
Tilmann Gneiting, Fadoua Balabdaoui, and Adrian~E. Raftery.
\newblock Probabilistic forecasts, calibration and sharpness.
\newblock \emph{Journal of the Royal Statistical Society: Series B (Statistical
  Methodology)}, 69\penalty0 (2):\penalty0 243--268, April 2007.
\newblock ISSN 1467-9868.
\newblock \doi{10.1111/j.1467-9868.2007.00587.x}.

\bibitem[Gouk et~al.(2021)Gouk, Frank, Pfahringer, and
  Cree]{gouk_regularisation_2018}
Henry Gouk, Eibe Frank, Bernhard Pfahringer, and Michael~J Cree.
\newblock Regularisation of neural networks by enforcing lipschitz continuity.
\newblock \emph{Machine Learning}, 110\penalty0 (2):\penalty0 393--416, 2021.

\bibitem[Gr\"{u}nwald and Dawid(2004)]{grunwald_game_2004}
Peter~D. Gr\"{u}nwald and A.~Philip Dawid.
\newblock Game theory, maximum entropy, minimum discrepancy and robust
  {Bayesian} decision theory.
\newblock \emph{Annals of Statistics}, 32\penalty0 (4):\penalty0 1367--1433,
  August 2004.
\newblock ISSN 0090-5364, 2168-8966.
\newblock \doi{10.1214/009053604000000553}.
\newblock Publisher: Institute of Mathematical Statistics.

\bibitem[Gulrajani et~al.(2017)Gulrajani, Ahmed, Arjovsky, Dumoulin, and
  Courville]{gulrajani_improved_2017}
Ishaan Gulrajani, Faruk Ahmed, Martin Arjovsky, Vincent Dumoulin, and Aaron
  Courville.
\newblock Improved training of wasserstein {GANs}.
\newblock In \emph{Proceedings of the 31st {International} {Conference} on
  {Neural} {Information} {Processing} {Systems}}, {NeurIPS}'17, pages
  5769--5779, Long Beach, California, USA, December 2017. Curran Associates
  Inc.
\newblock ISBN 978-1-5108-6096-4.

\bibitem[Guo et~al.(2017)Guo, Pleiss, Sun, and
  Weinberger]{guo_calibration_2017}
Chuan Guo, Geoff Pleiss, Yu~Sun, and Kilian~Q. Weinberger.
\newblock On {Calibration} of {Modern} {Neural} {Networks}.
\newblock In \emph{International {Conference} on {Machine} {Learning}}, pages
  1321--1330, July 2017.
\newblock ISSN: 1938-7228 Section: Machine Learning.

\bibitem[Gupta et~al.(2021)Gupta, Anpalagan, Guan, and Khwaja]{GUPTA2021100057}
Abhishek Gupta, Alagan Anpalagan, Ling Guan, and Ahmed~Shaharyar Khwaja.
\newblock Deep learning for object detection and scene perception in
  self-driving cars: Survey, challenges, and open issues.
\newblock \emph{Array}, 10:\penalty0 100057, 2021.
\newblock ISSN 2590-0056.
\newblock \doi{https://doi.org/10.1016/j.array.2021.100057}.

\bibitem[Hafner et~al.(2020)Hafner, Tran, Lillicrap, Irpan, and
  Davidson]{hafner_reliable_2018}
Danijar Hafner, Dustin Tran, Timothy Lillicrap, Alex Irpan, and James Davidson.
\newblock Noise contrastive priors for functional uncertainty.
\newblock In \emph{Uncertainty in Artificial Intelligence}, pages 905--914.
  PMLR, 2020.

\bibitem[Han et~al.(2021)Han, Lakshminarayanan, and Liu]{han2021reliable}
Kehang Han, Balaji Lakshminarayanan, and Jeremiah~Zhe Liu.
\newblock Reliable graph neural networks for drug discovery under
  distributional shift.
\newblock In \emph{NeurIPS 2021 Workshop on Distribution Shifts: Connecting
  Methods and Applications}, 2021.

\bibitem[Harang and Rudd(2018)]{harang_principled_2018}
Richard Harang and Ethan~M Rudd.
\newblock Towards principled uncertainty estimation for deep neural networks.
\newblock \emph{arXiv preprint arXiv:1810.12278}, 2018.

\bibitem[Hashimoto et~al.(2016)Hashimoto, Alvarez-Melis, and
  Jaakkola]{hashimoto2016word}
Tatsunori~B Hashimoto, David Alvarez-Melis, and Tommi~S Jaakkola.
\newblock Word embeddings as metric recovery in semantic spaces.
\newblock \emph{Transactions of the Association for Computational Linguistics},
  4:\penalty0 273--286, 2016.

\bibitem[Hauser and Ray(2017{\natexlab{a}})]{NIPS2017_0ebcc77d}
Michael Hauser and Asok Ray.
\newblock Principles of {R}iemannian geometry in neural networks.
\newblock In \emph{Advances in Neural Information Processing Systems},
  volume~30. Curran Associates, Inc., 2017{\natexlab{a}}.

\bibitem[Hauser and Ray(2017{\natexlab{b}})]{hauser_principles_2017}
Michael Hauser and Asok Ray.
\newblock Principles of {Riemannian} {Geometry} in {Neural} {Networks}.
\newblock In \emph{Advances in {Neural} {Information} {Processing} {Systems}
  30}, pages 2807--2816. Curran Associates, Inc., 2017{\natexlab{b}}.

\bibitem[Hein et~al.(2019{\natexlab{a}})Hein, Andriushchenko, and
  Bitterwolf]{hein2019relu}
Matthias Hein, Maksym Andriushchenko, and Julian Bitterwolf.
\newblock Why relu networks yield high-confidence predictions far away from the
  training data and how to mitigate the problem.
\newblock In \emph{Proceedings of the IEEE/CVF Conference on Computer Vision
  and Pattern Recognition}, pages 41--50, 2019{\natexlab{a}}.

\bibitem[Hein et~al.(2019{\natexlab{b}})Hein, Andriushchenko, and
  Bitterwolf]{hein_why_2019}
Matthias Hein, Maksym Andriushchenko, and Julian Bitterwolf.
\newblock Why {ReLU} {Networks} {Yield} {High}-{Confidence} {Predictions} {Far}
  {Away} {From} the {Training} {Data} and {How} to {Mitigate} the {Problem}.
\newblock In \emph{Proceedings of the IEEE/CVF Conference on Computer Vision
  and Pattern Recognition}, pages 41--50, 2019{\natexlab{b}}.

\bibitem[Hendrycks and Dietterich(2018)]{hendrycks_benchmarking_2019}
Dan Hendrycks and Thomas Dietterich.
\newblock Benchmarking neural network robustness to common corruptions and
  perturbations.
\newblock In \emph{International Conference on Learning Representations}, 2018.

\bibitem[Hendrycks and Gimpel(2017)]{hendrycks_baseline_2017}
Dan Hendrycks and Kevin Gimpel.
\newblock A {Baseline} for {Detecting} {Misclassified} and
  {Out}-of-{Distribution} {Examples} in {Neural} {Networks}.
\newblock In \emph{International {Conference} on {Learning} {Representations}},
  Toulon, France, April 2017.

\bibitem[Hendrycks et~al.(2018)Hendrycks, Mazeika, and
  Dietterich]{hendrycks_deep_2018}
Dan Hendrycks, Mantas Mazeika, and Thomas Dietterich.
\newblock Deep anomaly detection with outlier exposure.
\newblock In \emph{International Conference on Learning Representations}, 2018.

\bibitem[Hendrycks et~al.(2019)Hendrycks, Lee, and
  Mazeika]{hendrycks_using_2019}
Dan Hendrycks, Kimin Lee, and Mantas Mazeika.
\newblock Using {Pre}-{Training} {Can} {Improve} {Model} {Robustness} and
  {Uncertainty}.
\newblock In \emph{International {Conference} on {Machine} {Learning}}, pages
  2712--2721, May 2019.
\newblock ISSN: 1938-7228 Section: Machine Learning.

\bibitem[Hendrycks* et~al.(2020)Hendrycks*, Mu*, Cubuk, Zoph, Gilmer, and
  Lakshminarayanan]{hendrycks_augmix_2020}
Dan Hendrycks*, Norman Mu*, Ekin~Dogus Cubuk, Barret Zoph, Justin Gilmer, and
  Balaji Lakshminarayanan.
\newblock {AugMix}: {A} {Simple} {Method} to {Improve} {Robustness} and
  {Uncertainty} under {Data} {Shift}.
\newblock In \emph{International {Conference} on {Learning} {Representations}},
  2020.

\bibitem[Hensman et~al.(2015)Hensman, Matthews, and
  Ghahramani]{hensman2015scalable}
James Hensman, Alexander Matthews, and Zoubin Ghahramani.
\newblock Scalable variational {G}aussian process classification.
\newblock In \emph{Artificial Intelligence and Statistics}, pages 351--360.
  PMLR, 2015.

\bibitem[Higgins et~al.(2018)Higgins, Amos, Pfau, Racaniere, Matthey, Rezende,
  and Lerchner]{higgins2018towards}
Irina Higgins, David Amos, David Pfau, Sebastien Racaniere, Loic Matthey,
  Danilo Rezende, and Alexander Lerchner.
\newblock Towards a definition of disentangled representations.
\newblock \emph{arXiv preprint arXiv:1812.02230}, 2018.

\bibitem[Hobbhahn et~al.(2022)Hobbhahn, Kristiadi, and
  Hennig]{hobbhahn2020fast}
Marius Hobbhahn, Agustinus Kristiadi, and Philipp Hennig.
\newblock Fast predictive uncertainty for classification with bayesian deep
  networks.
\newblock In \emph{Uncertainty in Artificial Intelligence}, pages 822--832.
  PMLR, 2022.

\bibitem[Ioffe and Szegedy(2015)]{ioffe2015batch}
Sergey Ioffe and Christian Szegedy.
\newblock Batch normalization: Accelerating deep network training by reducing
  internal covariate shift.
\newblock In \emph{International conference on machine learning}, pages
  448--456. PMLR, 2015.

\bibitem[Jacobsen et~al.(2019{\natexlab{a}})Jacobsen, Behrmann, Zemel, and
  Bethge]{jacobsen_excessive_2018}
Joern-Henrik Jacobsen, Jens Behrmann, Richard Zemel, and Matthias Bethge.
\newblock Excessive invariance causes adversarial vulnerability.
\newblock In \emph{International Conference on Learning Representations},
  2019{\natexlab{a}}.

\bibitem[Jacobsen et~al.(2019{\natexlab{b}})Jacobsen, Behrmannn, Carlini,
  Tramer, and Papernot]{jacobsen_exploiting_2019}
J{\"o}rn-Henrik Jacobsen, Jens Behrmannn, Nicholas Carlini, Florian Tramer, and
  Nicolas Papernot.
\newblock Exploiting excessive invariance caused by norm-bounded adversarial
  robustness.
\newblock \emph{arXiv preprint arXiv:1903.10484}, 2019{\natexlab{b}}.

\bibitem[Jacobsen et~al.(2018)Jacobsen, Smeulders, and
  Oyallon]{jacobsen_i-revnet_2018}
Jörn-Henrik Jacobsen, Arnold~W.M. Smeulders, and Edouard Oyallon.
\newblock {i-RevNet}: Deep invertible networks.
\newblock In \emph{International Conference on Learning Representations}, 2018.

\bibitem[Jumper et~al.(2021)Jumper, Evans, Pritzel, Green, Figurnov,
  Ronneberger, Tunyasuvunakool, Bates, {\v{Z}}{\'\i}dek, Potapenko,
  et~al.]{jumper2021highly}
John Jumper, Richard Evans, Alexander Pritzel, Tim Green, Michael Figurnov,
  Olaf Ronneberger, Kathryn Tunyasuvunakool, Russ Bates, Augustin
  {\v{Z}}{\'\i}dek, Anna Potapenko, et~al.
\newblock Highly accurate protein structure prediction with {AlphaFold}.
\newblock \emph{Nature}, 596\penalty0 (7873):\penalty0 583--589, 2021.

\bibitem[Khan et~al.(2019)Khan, Immer, Abedi, and
  Korzepa]{khan_approximate_2019}
Mohammad Emtiyaz~E Khan, Alexander Immer, Ehsan Abedi, and Maciej Korzepa.
\newblock Approximate {Inference} {Turns} {Deep} {Networks} into {Gaussian}
  {Processes}.
\newblock In \emph{Advances in {Neural} {Information} {Processing} {Systems}
  32}, pages 3094--3104. Curran Associates, Inc., 2019.

\bibitem[Khemakhem et~al.(2020)Khemakhem, Kingma, Monti, and
  Hyvarinen]{khemakhem2020variational}
Ilyes Khemakhem, Diederik Kingma, Ricardo Monti, and Aapo Hyvarinen.
\newblock Variational autoencoders and nonlinear ica: A unifying framework.
\newblock In \emph{International Conference on Artificial Intelligence and
  Statistics}, pages 2207--2217. PMLR, 2020.

\bibitem[Kivlichan et~al.(2021)Kivlichan, Lin, Liu, and
  Vasserman]{kivlichan-etal-2021-measuring}
Ian Kivlichan, Zi~Lin, Jeremiah Liu, and Lucy Vasserman.
\newblock Measuring and improving model-moderator collaboration using
  uncertainty estimation.
\newblock In \emph{Proceedings of the 5th Workshop on Online Abuse and Harms
  (WOAH 2021)}, pages 36--53, Online, August 2021. Association for
  Computational Linguistics.
\newblock \doi{10.18653/v1/2021.woah-1.5}.

\bibitem[Kristiadi et~al.(2020)Kristiadi, Hein, and
  Hennig]{kristiadi_being_2020}
Agustinus Kristiadi, Matthias Hein, and Philipp Hennig.
\newblock Being bayesian, even just a bit, fixes overconfidence in relu
  networks.
\newblock In \emph{International conference on machine learning}, pages
  5436--5446. PMLR, 2020.

\bibitem[Kristiadi et~al.(2021)Kristiadi, Hein, and
  Hennig]{pmlr-v161-kristiadi21a}
Agustinus Kristiadi, Matthias Hein, and Philipp Hennig.
\newblock Learnable uncertainty under {L}aplace approximations.
\newblock In \emph{Proceedings of the Thirty-Seventh Conference on Uncertainty
  in Artificial Intelligence}, volume 161 of \emph{Proceedings of Machine
  Learning Research}, pages 344--353. PMLR, 27--30 Jul 2021.

\bibitem[Kristiadi et~al.(2022)Kristiadi, Hein, and
  Hennig]{DBLP:journals/corr/abs-2106-10065}
Agustinus Kristiadi, Matthias Hein, and Philipp Hennig.
\newblock Being a bit frequentist improves bayesian neural networks.
\newblock In \emph{International Conference on Artificial Intelligence and
  Statistics}, pages 529--545. PMLR, 2022.

\bibitem[Kunstner et~al.(2019)Kunstner, Hennig, and
  Balles]{kunstner2019limitations}
Frederik Kunstner, Philipp Hennig, and Lukas Balles.
\newblock Limitations of the empirical fisher approximation for natural
  gradient descent.
\newblock \emph{Advances in neural information processing systems}, 32, 2019.

\bibitem[Lakshminarayanan et~al.(2017)Lakshminarayanan, Pritzel, and
  Blundell]{lakshminarayanan_simple_2017}
Balaji Lakshminarayanan, Alexander Pritzel, and Charles Blundell.
\newblock Simple and {Scalable} {Predictive} {Uncertainty} {Estimation} using
  {Deep} {Ensembles}.
\newblock In \emph{Advances in {Neural} {Information} {Processing} {Systems}
  30}, pages 6402--6413. Curran Associates, Inc., 2017.

\bibitem[Landes(2015)]{landes_probabilism_2015}
Jurgen Landes.
\newblock Probabilism, entropies and strictly proper scoring rules.
\newblock \emph{International Journal of Approximate Reasoning}, 63:\penalty0
  1--21, August 2015.
\newblock ISSN 0888-613X.

\bibitem[Larson et~al.(2019)Larson, Mahendran, Peper, Clarke, Lee, Hill,
  Kummerfeld, Leach, Laurenzano, Tang, et~al.]{larson_evaluation_2019}
Stefan Larson, Anish Mahendran, Joseph~J Peper, Christopher Clarke, Andrew Lee,
  Parker Hill, Jonathan~K Kummerfeld, Kevin Leach, Michael~A Laurenzano,
  Lingjia Tang, et~al.
\newblock An evaluation dataset for intent classification and out-of-scope
  prediction.
\newblock In \emph{Proceedings of the 2019 Conference on Empirical Methods in
  Natural Language Processing and the 9th International Joint Conference on
  Natural Language Processing (EMNLP-IJCNLP)}, pages 1311--1316, 2019.

\bibitem[Lawrence and Quinonero-Candela(2006)]{lawrence_local_2006}
Neil~D Lawrence and Joaquin Quinonero-Candela.
\newblock Local distance preservation in the {GP-LVM} through back constraints.
\newblock In \emph{Proceedings of the 23rd international conference on Machine
  learning}, pages 513--520, 2006.

\bibitem[LeCam(1973)]{lecam_convergence_1973}
L.~LeCam.
\newblock Convergence of {Estimates} {Under} {Dimensionality} {Restrictions}.
\newblock \emph{The Annals of Statistics}, 1\penalty0 (1):\penalty0 38--53,
  January 1973.
\newblock ISSN 0090-5364, 2168-8966.
\newblock \doi{10.1214/aos/1193342380}.

\bibitem[Lee et~al.(2018{\natexlab{a}})Lee, Lee, Lee, and
  Shin]{lee_training_2018}
Kimin Lee, Honglak Lee, Kibok Lee, and Jinwoo Shin.
\newblock Training {Confidence}-calibrated {Classifiers} for {Detecting}
  {Out}-of-{Distribution} {Samples}.
\newblock In \emph{International {Conference} on {Learning} {Representations}},
  2018{\natexlab{a}}.

\bibitem[Lee et~al.(2018{\natexlab{b}})Lee, Lee, Lee, and Shin]{lee2018simple}
Kimin Lee, Kibok Lee, Honglak Lee, and Jinwoo Shin.
\newblock A simple unified framework for detecting out-of-distribution samples
  and adversarial attacks.
\newblock \emph{Advances in neural information processing systems}, 31,
  2018{\natexlab{b}}.

\bibitem[Lee et~al.(2018{\natexlab{c}})Lee, Lee, Lee, and
  Shin]{lee_simple_2018-2}
Kimin Lee, Kibok Lee, Honglak Lee, and Jinwoo Shin.
\newblock A {Simple} {Unified} {Framework} for {Detecting}
  {Out}-of-{Distribution} {Samples} and {Adversarial} {Attacks}.
\newblock In \emph{Advances in {Neural} {Information} {Processing} {Systems}
  31}, pages 7167--7177. Curran Associates, Inc., 2018{\natexlab{c}}.

\bibitem[Liang et~al.(2018)Liang, Li, and Srikant]{liang2017enhancing}
Shiyu Liang, Yixuan Li, and R~Srikant.
\newblock Enhancing the reliability of out-of-distribution image detection in
  neural networks.
\newblock In \emph{International Conference on Learning Representations}, 2018.

\bibitem[Liu et~al.(2021)Liu, Huang, Chen, and Suykens]{liu2020random}
Fanghui Liu, Xiaolin Huang, Yudong Chen, and Johan~AK Suykens.
\newblock Random features for kernel approximation: A survey on algorithms,
  theory, and beyond.
\newblock \emph{IEEE Transactions on Pattern Analysis and Machine
  Intelligence}, 44\penalty0 (10):\penalty0 7128--7148, 2021.

\bibitem[Liu et~al.(2020{\natexlab{a}})Liu, Lin, Padhy, Tran, Bedrax~Weiss, and
  Lakshminarayanan]{liu2020simple}
Jeremiah Liu, Zi~Lin, Shreyas Padhy, Dustin Tran, Tania Bedrax~Weiss, and
  Balaji Lakshminarayanan.
\newblock Simple and principled uncertainty estimation with deterministic deep
  learning via distance awareness.
\newblock \emph{Advances in Neural Information Processing Systems},
  33:\penalty0 7498--7512, 2020{\natexlab{a}}.

\bibitem[Liu et~al.(2020{\natexlab{b}})Liu, Wang, Owens, and Li]{liu2020energy}
Weitang Liu, Xiaoyun Wang, John Owens, and Yixuan Li.
\newblock Energy-based out-of-distribution detection.
\newblock \emph{Advances in Neural Information Processing Systems},
  33:\penalty0 21464--21475, 2020{\natexlab{b}}.

\bibitem[Lu et~al.(2020)Lu, Ie, and Sha]{lu2020mean}
Zhiyun Lu, Eugene Ie, and Fei Sha.
\newblock Mean-field approximation to {G}aussian-softmax integral with
  application to uncertainty estimation.
\newblock \emph{arXiv preprint arXiv:2006.07584}, 2020.

\bibitem[Mac{\^e}do and Ludermir(2021)]{macedo2021enhanced}
David Mac{\^e}do and Teresa Ludermir.
\newblock Enhanced isotropy maximization loss: Seamless and high-performance
  out-of-distribution detection simply replacing the softmax loss.
\newblock \emph{arXiv preprint arXiv:2105.14399}, 2021.

\bibitem[Macedo et~al.(2020)Macedo, Ren, Zanchettin, Oliveira, Tapp, and
  Ludermir]{macedo_isotropic_2020}
David Macedo, Tsang~Ing Ren, Cleber Zanchettin, Adriano L.~I. Oliveira, Alain
  Tapp, and Teresa Ludermir.
\newblock Isotropic {Maximization} {Loss} and {Entropic} {Score}: {Fast},
  {Accurate}, {Scalable}, {Unexposed}, {Turnkey}, and {Native} {Neural}
  {Networks} {Out}-of-{Distribution} {Detection}.
\newblock \emph{arXiv:1908.05569 [cs, stat]}, February 2020.
\newblock arXiv: 1908.05569.

\bibitem[Mac{\^e}do et~al.(2022)Mac{\^e}do, Zanchettin, and
  Ludermir]{macedo2022distinction}
David Mac{\^e}do, Cleber Zanchettin, and Teresa Ludermir.
\newblock Distinction maximization loss: Efficiently improving classification
  accuracy, uncertainty estimation, and out-of-distribution detection simply
  replacing the loss and calibrating.
\newblock \emph{arXiv preprint arXiv:2205.05874}, 2022.

\bibitem[MacKay(1992)]{mackay_practical_1992}
David J.~C. MacKay.
\newblock A practical {Bayesian} framework for backpropagation networks.
\newblock \emph{Neural Computation}, 4\penalty0 (3):\penalty0 448--472, May
  1992.
\newblock ISSN 0899-7667.
\newblock Number: 3 Publisher: MIT Press.

\bibitem[Malinin and Gales(2018{\natexlab{a}})]{malinin_predictive_2018}
Andrey Malinin and Mark Gales.
\newblock Predictive {Uncertainty} {Estimation} via {Prior} {Networks}.
\newblock In \emph{Advances in {Neural} {Information} {Processing} {Systems}
  31}, pages 7047--7058. Curran Associates, Inc., 2018{\natexlab{a}}.

\bibitem[Malinin and Gales(2018{\natexlab{b}})]{malinin_prior_2018}
Andrey Malinin and Mark Gales.
\newblock Prior {Networks} for {Detection} of {Adversarial} {Attacks}.
\newblock \emph{arXiv:1812.02575 [cs, stat]}, December 2018{\natexlab{b}}.
\newblock arXiv: 1812.02575.

\bibitem[Matou{\v{s}}ek(2013)]{matouvsek2013lecture}
Jir{\i} Matou{\v{s}}ek.
\newblock Lecture notes on metric embeddings.
\newblock Technical report, Technical report, ETH Z{\"u}rich, 2013.

\bibitem[Meinke and Hein(2020)]{meinke_towards_2020}
Alexander Meinke and Matthias Hein.
\newblock Towards neural networks that provably know when they don't know.
\newblock In \emph{International {Conference} on {Learning} {Representations}},
  2020.

\bibitem[Minka(2001)]{minka_family_2001}
Thomas~P. Minka.
\newblock \emph{A family of algorithms for approximate {B}ayesian inference}.
\newblock phd, Massachusetts Institute of Technology, USA, 2001.
\newblock AAI0803033.

\bibitem[Miyato et~al.(2018)Miyato, Kataoka, Koyama, and
  Yoshida]{miyato_spectral_2018}
Takeru Miyato, Toshiki Kataoka, Masanori Koyama, and Yuichi Yoshida.
\newblock Spectral {Normalization} for {Generative} {Adversarial} {Networks}.
\newblock In \emph{International {Conference} on {Learning} {Representations}},
  2018.

\bibitem[Mohammad and Hirst(2006)]{mohammad2006distributional}
Saif Mohammad and Graeme Hirst.
\newblock Distributional measures as proxies for semantic distance: A survey.
\newblock \emph{Computational Linguistics}, 1\penalty0 (1), 2006.

\bibitem[Mukhoti et~al.(2021)Mukhoti, Kirsch, van Amersfoort, Torr, and
  Gal]{mukhoti2021deterministic}
Jishnu Mukhoti, Andreas Kirsch, Joost van Amersfoort, Philip~HS Torr, and Yarin
  Gal.
\newblock Deterministic neural networks with appropriate inductive biases
  capture epistemic and aleatoric uncertainty.
\newblock \emph{arXiv preprint arXiv:2102.11582}, 2021.

\bibitem[Nado et~al.(2021)Nado, Band, Collier, Djolonga, Dusenberry, Farquhar,
  Filos, Havasi, Jenatton, Jerfel, et~al.]{nado2021uncertainty}
Zachary Nado, Neil Band, Mark Collier, Josip Djolonga, Michael~W Dusenberry,
  Sebastian Farquhar, Angelos Filos, Marton Havasi, Rodolphe Jenatton, Ghassen
  Jerfel, et~al.
\newblock Uncertainty baselines: Benchmarks for uncertainty \& robustness in
  deep learning.
\newblock \emph{arXiv preprint arXiv:2106.04015}, 2021.

\bibitem[Netzer et~al.(2011)Netzer, Wang, Coates, Bissacco, Wu, and
  Ng]{netzer_reading_2011}
Yuval Netzer, Tao Wang, Adam Coates, Alessandro Bissacco, Bo~Wu, and Andrew~Y.
  Ng.
\newblock Reading {Digits} in {Natural} {Images} with {Unsupervised} {Feature}
  {Learning}.
\newblock In \emph{{NeurIPS} {Workshop} on {Deep} {Learning} and {Unsupervised}
  {Feature} {Learning} 2011}, 2011.

\bibitem[Nguyen et~al.(2015)Nguyen, Yosinski, and Clune]{nguyen2015deep}
Anh Nguyen, Jason Yosinski, and Jeff Clune.
\newblock Deep neural networks are easily fooled: High confidence predictions
  for unrecognizable images.
\newblock In \emph{Proceedings of the IEEE conference on computer vision and
  pattern recognition}, pages 427--436, 2015.

\bibitem[Nixon et~al.(2019)Nixon, Dusenberry, Zhang, Jerfel, and
  Tran]{nixon2019measuring}
Jeremy Nixon, Michael~W Dusenberry, Linchuan Zhang, Ghassen Jerfel, and Dustin
  Tran.
\newblock Measuring calibration in deep learning.
\newblock In \emph{CVPR Workshop}, 2019.

\bibitem[Osawa et~al.(2019)Osawa, Swaroop, Khan, Jain, Eschenhagen, Turner, and
  Yokota]{osawa_practical_2019}
Kazuki Osawa, Siddharth Swaroop, Mohammad Emtiyaz~E Khan, Anirudh Jain, Runa
  Eschenhagen, Richard~E Turner, and Rio Yokota.
\newblock Practical {Deep} {Learning} with {Bayesian} {Principles}.
\newblock In \emph{Advances in {Neural} {Information} {Processing} {Systems}
  32}, pages 4287--4299. Curran Associates, Inc., 2019.

\bibitem[Ovadia et~al.(2019)Ovadia, Fertig, Ren, Nado, Sculley, Nowozin,
  Dillon, Lakshminarayanan, and Snoek]{ovadia_can_2019}
Yaniv Ovadia, Emily Fertig, Jie Ren, Zachary Nado, D.~Sculley, Sebastian
  Nowozin, Joshua Dillon, Balaji Lakshminarayanan, and Jasper Snoek.
\newblock Can you trust your model' s uncertainty? evaluating predictive
  uncertainty under dataset shift.
\newblock In \emph{Advances in Neural Information Processing Systems},
  volume~32. Curran Associates, Inc., 2019.

\bibitem[Padhy et~al.(2020)Padhy, Nado, Ren, Liu, Snoek, and
  Lakshminarayanan]{padhy2020revisiting}
Shreyas Padhy, Zachary Nado, Jie Ren, Jeremiah Liu, Jasper Snoek, and Balaji
  Lakshminarayanan.
\newblock Revisiting one-vs-all classifiers for predictive uncertainty and
  out-of-distribution detection in neural networks.
\newblock \emph{arXiv preprint arXiv:2007.05134}, 2020.

\bibitem[Panov and Spokoiny(2015)]{panov_finite_2015}
Maxim Panov and Vladimir Spokoiny.
\newblock Finite {Sample} {Bernstein} von {Mises} {Theorem} for
  {Semiparametric} {Problems}.
\newblock \emph{Bayesian Analysis}, 10\penalty0 (3):\penalty0 665--710,
  September 2015.
\newblock ISSN 1936-0975, 1931-6690.
\newblock \doi{10.1214/14-BA926}.

\bibitem[Papyan(2020)]{papyan2020traces}
Vardan Papyan.
\newblock Traces of class/cross-class structure pervade deep learning spectra.
\newblock \emph{Journal of Machine Learning Research}, 21\penalty0
  (252):\penalty0 1--64, 2020.

\bibitem[Parry et~al.(2012)Parry, Dawid, and Lauritzen]{parry_proper_2012}
Matthew Parry, A.~Philip Dawid, and Steffen Lauritzen.
\newblock Proper local scoring rules.
\newblock \emph{Annals of Statistics}, 40\penalty0 (1):\penalty0 561--592,
  February 2012.
\newblock ISSN 0090-5364, 2168-8966.
\newblock \doi{10.1214/12-AOS971}.
\newblock Publisher: Institute of Mathematical Statistics.

\bibitem[Perrault-Joncas and Meila(2012)]{perrault-joncas_metric_2017}
Dominique Perrault-Joncas and Marina Meila.
\newblock Metric learning and manifolds: Preserving the intrinsic geometry.
\newblock \emph{Preprint Department of Statistics, University of Washington},
  2012.

\bibitem[Poplin et~al.(2018)Poplin, Chang, Alexander, Schwartz, Colthurst, Ku,
  Newburger, Dijamco, Nguyen, Afshar, et~al.]{poplin2018universal}
Ryan Poplin, Pi-Chuan Chang, David Alexander, Scott Schwartz, Thomas Colthurst,
  Alexander Ku, Dan Newburger, Jojo Dijamco, Nam Nguyen, Pegah~T Afshar, et~al.
\newblock A universal {SNP} and small-indel variant caller using deep neural
  networks.
\newblock \emph{Nature biotechnology}, 36\penalty0 (10):\penalty0 983--987,
  2018.

\bibitem[Rahimi and Recht(2008{\natexlab{a}})]{rahimi2008uniform}
Ali Rahimi and Benjamin Recht.
\newblock Uniform approximation of functions with random bases.
\newblock In \emph{2008 46th Annual Allerton Conference on Communication,
  Control, and Computing}, pages 555--561. IEEE, 2008{\natexlab{a}}.

\bibitem[Rahimi and Recht(2008{\natexlab{b}})]{rahimi_random_2008}
Ali Rahimi and Benjamin Recht.
\newblock Random {Features} for {Large}-{Scale} {Kernel} {Machines}.
\newblock In \emph{Advances in {Neural} {Information} {Processing} {Systems}
  20}, pages 1177--1184. Curran Associates, Inc., 2008{\natexlab{b}}.

\bibitem[Rasmussen and Williams(2006)]{rasmussen_gaussian_2006}
Carl~Edward Rasmussen and Christopher K.~I. Williams.
\newblock \emph{Gaussian {Processes} for {Machine} {Learning}}.
\newblock University Press Group Limited, January 2006.
\newblock ISBN 978-0-262-18253-9.
\newblock Google-Books-ID: vWtwQgAACAAJ.

\bibitem[Reilly et~al.(2022)Reilly, Zuckerman, Finley, Litovsky, and
  Kenett]{reilly2022semantic}
Jamie Reilly, Bonnie Zuckerman, Ann~Marie Finley, Celia~Paula Litovsky, and
  Yoed Kenett.
\newblock What is semantic distance? a review and proposed method for modeling
  conceptual transitions in natural language.
\newblock 2022.

\bibitem[Ren et~al.(2019)Ren, Liu, Fertig, Snoek, Poplin, Depristo, Dillon, and
  Lakshminarayanan]{ren2019likelihood}
Jie Ren, Peter~J Liu, Emily Fertig, Jasper Snoek, Ryan Poplin, Mark Depristo,
  Joshua Dillon, and Balaji Lakshminarayanan.
\newblock Likelihood ratios for out-of-distribution detection.
\newblock \emph{Advances in neural information processing systems}, 32, 2019.

\bibitem[Ren et~al.(2021)Ren, Fort, Liu, Roy, Padhy, and
  Lakshminarayanan]{ren2021simple}
Jie Ren, Stanislav Fort, Jeremiah Liu, Abhijit~Guha Roy, Shreyas Padhy, and
  Balaji Lakshminarayanan.
\newblock A simple fix to {M}ahalanobis distance for improving near-{OOD}
  detection.
\newblock \emph{arXiv preprint arXiv:2106.09022}, 2021.

\bibitem[Riquelme et~al.(2018)Riquelme, Tucker, and Snoek]{riquelme_deep_2018}
Carlos Riquelme, George Tucker, and Jasper Snoek.
\newblock Deep {Bayesian} {Bandits} {Showdown}: {An} {Empirical} {Comparison}
  of {Bayesian} {Deep} {Networks} for {Thompson} {Sampling}.
\newblock In \emph{International {Conference} on {Learning} {Representations}},
  2018.

\bibitem[Ritter et~al.(2018)Ritter, Botev, and Barber]{ritter_scalable_2018}
Hippolyt Ritter, Aleksandar Botev, and David Barber.
\newblock A {Scalable} {Laplace} {Approximation} for {Neural} {Networks}.
\newblock In \emph{International {Conference} on {Learning} {Representations}},
  2018.

\bibitem[Rousseau et~al.(2020)Rousseau, Drumetz, and
  Fablet]{rousseau_residual_2020}
Francois Rousseau, Lucas Drumetz, and Ronan Fablet.
\newblock Residual {Networks} as {Flows} of {Diffeomorphisms}.
\newblock \emph{Journal of Mathematical Imaging and Vision}, 62\penalty0
  (3):\penalty0 365--375, April 2020.
\newblock ISSN 1573-7683.
\newblock \doi{10.1007/s10851-019-00890-3}.

\bibitem[Roy et~al.(2022)Roy, Ren, Azizi, Loh, Natarajan, Mustafa, Pawlowski,
  Freyberg, Liu, Beaver, et~al.]{roy2022does}
Abhijit~Guha Roy, Jie Ren, Shekoofeh Azizi, Aaron Loh, Vivek Natarajan, Basil
  Mustafa, Nick Pawlowski, Jan Freyberg, Yuan Liu, Zach Beaver, et~al.
\newblock Does your dermatology classifier know what it doesn’t know?
  detecting the long-tail of unseen conditions.
\newblock \emph{Medical Image Analysis}, 75:\penalty0 102274, 2022.

\bibitem[Ruan et~al.(2018)Ruan, Huang, and Kwiatkowska]{ruan_reachability_2018}
Wenjie Ruan, Xiaowei Huang, and Marta Kwiatkowska.
\newblock Reachability analysis of deep neural networks with provable
  guarantees.
\newblock In \emph{Proceedings of the 27th {International} {Joint} {Conference}
  on {Artificial} {Intelligence}}, {IJCAI}'18, pages 2651--2659, Stockholm,
  Sweden, July 2018. AAAI Press.
\newblock ISBN 978-0-9992411-2-7.

\bibitem[Rudin et~al.(1976)]{rudin1976principles}
Walter Rudin et~al.
\newblock \emph{Principles of mathematical analysis}, volume~3.
\newblock McGraw-hill New York, 1976.

\bibitem[Russakovsky et~al.(2015)Russakovsky, Deng, Su, Krause, Satheesh, Ma,
  Huang, Karpathy, Khosla, Bernstein, Berg, and
  Fei-Fei]{russakovsky_imagenet_2015}
Olga Russakovsky, Jia Deng, Hao Su, Jonathan Krause, Sanjeev Satheesh, Sean Ma,
  Zhiheng Huang, Andrej Karpathy, Aditya Khosla, Michael Bernstein,
  Alexander~C. Berg, and Li~Fei-Fei.
\newblock {ImageNet} {Large} {Scale} {Visual} {Recognition} {Challenge}.
\newblock \emph{International Journal of Computer Vision}, 115\penalty0
  (3):\penalty0 211--252, December 2015.
\newblock ISSN 1573-1405.
\newblock \doi{10.1007/s11263-015-0816-y}.

\bibitem[Salakhutdinov and Mnih(2008)]{salakhutdinov_bayesian_2008}
Ruslan Salakhutdinov and Andriy Mnih.
\newblock Bayesian {Probabilistic} {Matrix} {Factorization} {Using} {Markov}
  {Chain} {Monte} {Carlo}.
\newblock In \emph{Proceedings of the 25th {International} {Conference} on
  {Machine} {Learning}}, {ICML} '08, pages 880--887, New York, NY, USA, 2008.
  ACM.
\newblock ISBN 978-1-60558-205-4.
\newblock \doi{10.1145/1390156.1390267}.

\bibitem[Salakhutdinov and Hinton(2007)]{salakhutdinov07}
Russ~R Salakhutdinov and Geoffrey~E Hinton.
\newblock Using deep belief nets to learn covariance kernels for gaussian
  processes.
\newblock In \emph{Advances in Neural Information Processing Systems}, 2007.

\bibitem[Salimbeni and Deisenroth(2017)]{salimbeni_doubly_2017}
Hugh Salimbeni and Marc Deisenroth.
\newblock Doubly stochastic variational inference for deep {G}aussian
  processes.
\newblock \emph{Advances in neural information processing systems}, 30, 2017.

\bibitem[Scheirer et~al.(2014)Scheirer, Jain, and
  Boult]{scheirer_probability_2014}
Walter~J. Scheirer, Lalit~P. Jain, and Terrance~E. Boult.
\newblock Probability {Models} for {Open} {Set} {Recognition}.
\newblock \emph{IEEE Transactions on Pattern Analysis and Machine
  Intelligence}, 36\penalty0 (11):\penalty0 2317--2324, November 2014.
\newblock ISSN 1939-3539.
\newblock \doi{10.1109/TPAMI.2014.2321392}.
\newblock Conference Name: IEEE Transactions on Pattern Analysis and Machine
  Intelligence.

\bibitem[Searcod(2006)]{searcod_metric_2006}
Micheal~O. Searcod.
\newblock \emph{Metric {Spaces}}.
\newblock Springer London, London, 2007 edition edition, August 2006.
\newblock ISBN 978-1-84628-369-7.

\bibitem[Sensoy et~al.(2018{\natexlab{a}})Sensoy, Kaplan, and
  Kandemir]{sensoy2018evidential}
Murat Sensoy, Lance Kaplan, and Melih Kandemir.
\newblock Evidential deep learning to quantify classification uncertainty.
\newblock \emph{Advances in Neural Information Processing Systems}, 31,
  2018{\natexlab{a}}.

\bibitem[Sensoy et~al.(2018{\natexlab{b}})Sensoy, Kaplan, and
  Kandemir]{sensoy_evidential_2018}
Murat Sensoy, Lance Kaplan, and Melih Kandemir.
\newblock Evidential {Deep} {Learning} to {Quantify} {Classification}
  {Uncertainty}.
\newblock In \emph{Advances in {Neural} {Information} {Processing} {Systems}
  31}, pages 3179--3189. Curran Associates, Inc., 2018{\natexlab{b}}.

\bibitem[Shu et~al.(2017)Shu, Xu, and Liu]{shu_doc_2017}
Lei Shu, Hu~Xu, and Bing Liu.
\newblock Doc: Deep open classification of text documents.
\newblock In \emph{Proceedings of the 2017 Conference on Empirical Methods in
  Natural Language Processing}, pages 2911--2916, 2017.

\bibitem[Skafte et~al.(2019)Skafte, J{\o}rgensen, and
  Hauberg]{skafte2019reliable}
Nicki Skafte, Martin J{\o}rgensen, and S{\o}ren Hauberg.
\newblock Reliable training and estimation of variance networks.
\newblock \emph{Advances in Neural Information Processing Systems}, 32, 2019.

\bibitem[Smith et~al.(2021)Smith, van Amersfoort, Huang, Roberts, and
  Gal]{smith2021can}
Lewis Smith, Joost van Amersfoort, Haiwen Huang, Stephen Roberts, and Yarin
  Gal.
\newblock Can convolutional resnets approximately preserve input distances? a
  frequency analysis perspective.
\newblock \emph{arXiv preprint arXiv:2106.02469}, 2021.

\bibitem[Snoek et~al.(2015)Snoek, Rippel, Swersky, Kiros, Satish, Sundaram,
  Patwary, Prabhat, and Adams]{snoek_scalable_2015}
Jasper Snoek, Oren Rippel, Kevin Swersky, Ryan Kiros, Nadathur Satish,
  Narayanan Sundaram, Mostofa Patwary, Mr~Prabhat, and Ryan Adams.
\newblock Scalable bayesian optimization using deep neural networks.
\newblock In \emph{International conference on machine learning}, pages
  2171--2180. PMLR, 2015.

\bibitem[Sokolic et~al.(2017)Sokolic, Giryes, Sapiro, and
  Rodrigues]{sokolic_robust_2017}
Jure Sokolic, Raja Giryes, Guillermo Sapiro, and Miguel R.~D. Rodrigues.
\newblock Robust {Large} {Margin} {Deep} {Neural} {Networks}.
\newblock \emph{IEEE Transactions on Signal Processing}, 2017.
\newblock \doi{10.1109/TSP.2017.2708039}.

\bibitem[Tagasovska and Lopez-Paz(2019)]{tagasovska_single-model_2019}
Natasa Tagasovska and David Lopez-Paz.
\newblock Single-{Model} {Uncertainties} for {Deep} {Learning}.
\newblock In \emph{Advances in {Neural} {Information} {Processing} {Systems}
  32}, pages 6417--6428. Curran Associates, Inc., 2019.

\bibitem[Tenenbaum et~al.(2000)Tenenbaum, Silva, and
  Langford]{tenenbaum2000global}
Joshua~B Tenenbaum, Vin~de Silva, and John~C Langford.
\newblock A global geometric framework for nonlinear dimensionality reduction.
\newblock \emph{Science}, 290\penalty0 (5500):\penalty0 2319--2323, 2000.

\bibitem[Thulasidasan et~al.(2019)Thulasidasan, Chennupati, Bilmes,
  Bhattacharya, and Michalak]{thulasidasan_mixup_2019}
Sunil Thulasidasan, Gopinath Chennupati, Jeff~A Bilmes, Tanmoy Bhattacharya,
  and Sarah Michalak.
\newblock On {Mixup} {Training}: {Improved} {Calibration} and {Predictive}
  {Uncertainty} for {Deep} {Neural} {Networks}.
\newblock In \emph{Advances in {Neural} {Information} {Processing} {Systems}
  32}, pages 13888--13899. Curran Associates, Inc., 2019.

\bibitem[Tierney et~al.(1989)Tierney, Kass, and
  Kadane]{tierney_approximate_1989}
Luke Tierney, Robert~E. Kass, and Joseph~B. Kadane.
\newblock Approximate {Marginal} {Densities} of {Nonlinear} {Functions}.
\newblock \emph{Biometrika}, 76\penalty0 (3):\penalty0 425--433, 1989.
\newblock ISSN 0006-3444.
\newblock \doi{10.2307/2336109}.
\newblock Publisher: [Oxford University Press, Biometrika Trust].

\bibitem[Titsias(2009)]{titsias_variational_2009}
Michalis Titsias.
\newblock Variational {Learning} of {Inducing} {Variables} in {Sparse}
  {Gaussian} {Processes}.
\newblock In \emph{Artificial {Intelligence} and {Statistics}}, pages 567--574,
  April 2009.

\bibitem[Tran et~al.(2019)Tran, Bonilla, Cunningham, Michiardi, and
  Filippone]{tran_calibrating_2019}
Gia-Lac Tran, Edwin~V. Bonilla, John Cunningham, Pietro Michiardi, and Maurizio
  Filippone.
\newblock Calibrating {Deep} {Convolutional} {Gaussian} {Processes}.
\newblock In \emph{The 22nd {International} {Conference} on {Artificial}
  {Intelligence} and {Statistics}}, pages 1554--1563, April 2019.
\newblock ISSN: 1938-7228 Section: Machine Learning.

\bibitem[Tsuzuku et~al.(2018)Tsuzuku, Sato, and
  Sugiyama]{tsuzuku_lipschitz-margin_2018}
Yusuke Tsuzuku, Issei Sato, and Masashi Sugiyama.
\newblock Lipschitz-{Margin} {Training}: {Scalable} {Certification} of
  {Perturbation} {Invariance} for {Deep} {Neural} {Networks}.
\newblock In \emph{Advances in {Neural} {Information} {Processing} {Systems}
  31}, pages 6541--6550. Curran Associates, Inc., 2018.

\bibitem[Van~Amersfoort et~al.(2020)Van~Amersfoort, Smith, Teh, and
  Gal]{van2020uncertainty}
Joost Van~Amersfoort, Lewis Smith, Yee~Whye Teh, and Yarin Gal.
\newblock Uncertainty estimation using a single deep deterministic neural
  network.
\newblock In \emph{International conference on machine learning}, pages
  9690--9700. PMLR, 2020.

\bibitem[van Amersfoort et~al.(2021)van Amersfoort, Smith, Jesson, Key, and
  Gal]{van2021feature}
Joost van Amersfoort, Lewis Smith, Andrew Jesson, Oscar Key, and Yarin Gal.
\newblock On feature collapse and deep kernel learning for single forward pass
  uncertainty.
\newblock \emph{arXiv preprint arXiv:2102.11409}, 2021.

\bibitem[Vedula et~al.(2019)Vedula, Lipka, Maneriker, and
  Parthasarathy]{vedula_towards_2019}
Nikhita Vedula, Nedim Lipka, Pranav Maneriker, and Srinivasan Parthasarathy.
\newblock Towards {Open} {Intent} {Discovery} for {Conversational} {Text}.
\newblock \emph{arXiv:1904.08524 [cs]}, April 2019.
\newblock arXiv: 1904.08524.

\bibitem[Wahba(1990)]{wahba_spline_1990}
Grace Wahba.
\newblock \emph{Spline {Models} for {Observational} {Data}}.
\newblock SIAM, September 1990.
\newblock ISBN 978-0-89871-244-5.

\bibitem[Wen et~al.(2020)Wen, Tran, and Ba]{wen_batchensemble_2020}
Yeming Wen, Dustin Tran, and Jimmy Ba.
\newblock {BatchEnsemble}: an {Alternative} {Approach} to {Efficient}
  {Ensemble} and {Lifelong} {Learning}.
\newblock In \emph{International {Conference} on {Learning} {Representations}},
  2020.

\bibitem[Weng et~al.(2018)Weng, Zhang, Chen, Yi, Su, Gao, Hsieh, and
  Daniel]{weng_evaluating_2018}
Tsui-Wei Weng, Huan Zhang, Pin-Yu Chen, Jinfeng Yi, Dong Su, Yupeng Gao,
  Cho-Jui Hsieh, and Luca Daniel.
\newblock Evaluating the {Robustness} of {Neural} {Networks}: {An} {Extreme}
  {Value} {Theory} {Approach}.
\newblock In \emph{International {Conference} on {Learning} {Representations}},
  2018.

\bibitem[Wenzel et~al.(2020)Wenzel, Roth, Veeling, Swiatkowski, Tran, Mandt,
  Snoek, Salimans, Jenatton, and Nowozin]{wenzel_how_2020}
Florian Wenzel, Kevin Roth, Bastiaan Veeling, Jakub Swiatkowski, Linh Tran,
  Stephan Mandt, Jasper Snoek, Tim Salimans, Rodolphe Jenatton, and Sebastian
  Nowozin.
\newblock How good is the {B}ayes posterior in deep neural networks really?
\newblock In \emph{Proceedings of the 37th International Conference on Machine
  Learning}, volume 119 of \emph{Proceedings of Machine Learning Research},
  pages 10248--10259. PMLR, 13--18 Jul 2020.

\bibitem[Wilson and Izmailov(2020)]{wilson2020bayesian}
Andrew~G Wilson and Pavel Izmailov.
\newblock Bayesian deep learning and a probabilistic perspective of
  generalization.
\newblock \emph{Advances in neural information processing systems},
  33:\penalty0 4697--4708, 2020.

\bibitem[Wilson et~al.(2016{\natexlab{a}})Wilson, Hu, Salakhutdinov, and
  Xing]{wilson_deep_2015}
Andrew~Gordon Wilson, Zhiting Hu, Ruslan Salakhutdinov, and Eric~P Xing.
\newblock Deep kernel learning.
\newblock In \emph{Artificial intelligence and statistics}, pages 370--378.
  PMLR, 2016{\natexlab{a}}.

\bibitem[Wilson et~al.(2016{\natexlab{b}})Wilson, Hu, Salakhutdinov, and
  Xing]{wilson_stochastic_2016}
Andrew~Gordon Wilson, Zhiting Hu, Ruslan Salakhutdinov, and Eric~P. Xing.
\newblock Stochastic {Variational} {Deep} {Kernel} {Learning}.
\newblock In \emph{Proceedings of the 30th {International} {Conference} on
  {Neural} {Information} {Processing} {Systems}}, {NeurIPS}'16, pages
  2594--2602, USA, 2016{\natexlab{b}}. Curran Associates Inc.
\newblock ISBN 978-1-5108-3881-9.

\bibitem[Yaghoub-Zadeh-Fard et~al.(2020)Yaghoub-Zadeh-Fard, Benatallah, Casati,
  Chai~Barukh, and Zamanirad]{yaghoub-zadeh-fard_user_2020}
Mohammad-Ali Yaghoub-Zadeh-Fard, Boualem Benatallah, Fabio Casati, Moshe
  Chai~Barukh, and Shayan Zamanirad.
\newblock User {Utterance} {Acquisition} for {Training} {Task}-{Oriented}
  {Bots}: {A} {Review} of {Challenges}, {Techniques} and {Opportunities}.
\newblock \emph{IEEE Internet Computing}, pages 1--1, 2020.
\newblock ISSN 1941-0131.
\newblock \doi{10.1109/MIC.2020.2978157}.
\newblock Conference Name: IEEE Internet Computing.

\bibitem[Yu et~al.(2016)Yu, Suresh, Choromanski, Holtmann-Rice, and
  Kumar]{yu_orthogonal_2016}
Felix Xinnan~X Yu, Ananda~Theertha Suresh, Krzysztof~M Choromanski, Daniel~N
  Holtmann-Rice, and Sanjiv Kumar.
\newblock Orthogonal {Random} {Features}.
\newblock In \emph{Advances in {Neural} {Information} {Processing} {Systems}
  29}, pages 1975--1983. Curran Associates, Inc., 2016.

\bibitem[Zagoruyko and Komodakis(2017)]{zagoruyko_wide_2017}
Sergey Zagoruyko and Nikos Komodakis.
\newblock Wide {Residual} {Networks}.
\newblock \emph{arXiv:1605.07146 [cs]}, June 2017.
\newblock arXiv: 1605.07146.

\bibitem[Zheng et~al.(2020)Zheng, Chen, and Huang]{zheng_out--domain_2020}
Yinhe Zheng, Guanyi Chen, and Minlie Huang.
\newblock Out-of-{Domain} {Detection} for {Natural} {Language} {Understanding}
  in {Dialog} {Systems}.
\newblock \emph{IEEE/ACM Transactions on Audio, Speech, and Language
  Processing}, 28:\penalty0 1198--1209, 2020.
\newblock ISSN 2329-9304.
\newblock \doi{10.1109/TASLP.2020.2983593}.
\newblock Conference Name: IEEE/ACM Transactions on Audio, Speech, and Language
  Processing.

\end{thebibliography}

\clearpage
\newpage
\appendix

\section{Method Summary}
\label{sec:method_sum}

\paragraph{Architecture.} Given a deep learning model  $\logit(\bx)=g \circ h(\bx)$ with $L-1$ hidden layers of size $\{D_l\}_{l=1}^L$, \gls{SNGP} makes two changes to the model: 
\begin{enumerate}[leftmargin=2em]
    \item Adding spectral normalization to the hidden weights $\{\bW_l\}_{l=1}^L$, and
    \item Replacing the dense output layer $g(h) = h^\top\bbeta$ with a \gls{GP} layer. Under the \gls{RFF} approximation, the \gls{GP} layer is simply a one-layer network with $D_L$ hidden units $g(h) \propto \cos(-\bW_L h_i + \bb_L)^\top\bbeta$. Here $\{\bW_L, \bb_L\}$ are frozen weights that are initialized from a Gaussian and a uniform distribution, respectively (as described in Equation (\ref{eq:rff_lr})). 
\end{enumerate} 

\paragraph{Training.} Algorithm \ref{alg:training} summarizes the training step. As shown, for every minibatch step, the model first updates the hidden-layer weights $\{\bW_l, \bb_l\}_{l=1}^{L-1}$ and the trainable output  weights $\beta=\{\beta_k\}_{k=1}^K$ via \gls{SGD}, 
then performs spectral normalization using power iteration method (\citep{gouk_regularisation_2018} which has time complexity $O(\sum_{l=1}^{L-1} D_l)$), and finally performs precision matrix update (Equation (\ref{eq:gp_posterior_approx}), time complexity $O(D^2_L)$). Since $\{D_l\}_{l=1}^{L-1}$ are fixed for a given architecture and usually $D_L \leq 1024$, the computation scales linearly with respect to the sample size. We use $D_L=1024$ in the experiments.

\paragraph{Prediction.} Algorithm \ref{alg:prediction} summarizes the prediction step. The model first performs the conventional forward pass to compute the final hidden feature $\Phi(\bx)_{D_L \times 1}$, and then compute the posterior mean  $\hat{m}_k(\bx)=\Phi^\top\bbeta_k$ (time complexity $O(D_L)$) and the predictive variance $\hat{\sigma}_k(\bx)^2=\Phi(\bx)^\top\hat{\Sigma}\Phi(\bx)$ (time complexity $O(D^2_L)$). 

To estimate the predictive distribution $p_k = \exp(m_k) / \sum_k \exp(m_k)$ where $m_k \sim N \big(\hat{m}_k(\bx), \hat{\sigma}_k^2(\bx)\big)$, we calculate its posterior mean using either Monte Carlo averaging or mean-field approaximation. Notice that this Monte Carlo averaging is computationally  cheap since it only involves sampling from a closed-form distribution whose parameters $(\hat{m}, \hat{\sigma}^2)$ are already computed by the single feed-forward pass (i.e., a single call to {\small\texttt{tf.random.normal}}). This is different from the full Monte Carlo sampling used by \gls{MCD} or deep ensembles which require multiple forward passes and are computationally expensive. 

However, in applications where the inference latency is of high priority (e.g., in the on-device settings), we can reduce the computational overhead further by replacing the Monte Carlo averaging with the mean-field approximation as described in the main text (Equation \ref{eq:mean_field_approx}) \citep{daunizeau_semi-analytical_2017}.


\subsection{Extension to Regression and Multi-class Classification}
\label{sec:multiclass_app}

As introduced in the main text (\ref{eq:laplace}), for an arbitrary model posterior $p(\beta|\Dsc)$, its corresponding Laplace posterior is:
\begin{align*}
p(\beta | \Dsc) \approx MVN(\hat{\beta}, \hat{\bSigma}=\hat{\bH}^{-1}), 
\quad \mbox{where} \quad 
\hat{\bH}_{(i,j)} = -\frac{\partial^2}{\partial \beta_i \partial \beta_j} \log \, p(\beta|\Dsc)|_{\beta=\hat{\beta}}.
\end{align*}
Consequently, to compute the Laplace posterior for any data likelihood $p(\beta|\Dsc)$, it is sufficient to first compute the \gls{MAP} estimate $\hat{\beta}$ as done in (\ref{eq:gp_objective}), and then derive the model Hessian $\hat{\Sigma}=\frac{\partial^2}{\partial \beta_i \partial \beta_j} \log \, p(\beta|\Dsc)|_{\beta=\hat{\beta}}$.

In the main text, we introduced the expression for $\hat{\bSigma}$ for a sigmoid cross entropy likelihood $\log p(\beta|\Dsc)=-\sum_{i=1}^n \big[ I(y_i=1)\log p_i + I(y_i=0)\log (1-p_i) \big] + \frac{1}{2}||\bbeta||^2_2$ based on the classic result from \cite{rasmussen_gaussian_2006}. In this section, we show how to apply (\ref{eq:laplace}) to other common likelihood functions such as regression (using a squared loss) and $K$-class classification (using a softmax cross entropy loss) by deriving their Hessian.

\paragraph{Regression.}

Under squared loss, The model posterior likelihood is:
\begin{align*}
    -\log p(\beta_{D_L \times 1}|\Dsc) 
    = \sum_{i=1}^n \Big[\frac{1}{2}(y_i - g_i)^2\Big] + \frac{1}{2} ||\beta||^2,
\end{align*}
where $g_i = \Phi_i\beta^\top$. Then the model Hessian is:
\begin{align*}
    \hat{\bH}_{D_L \times D_L} = \sum_{i=1}^n \Phi_i \Phi_i^\top + \bI_{D_L}
\end{align*}

\paragraph{Multi-class Classification.}

Under a $K$-class softmax cross entropy loss, the model posterior likelihood is:

\begin{align*}
    -\log p(\beta|\Dsc) 
    =  \Big[\sum_{i=1}^n \sum_{k=1}^K y_{i,k} * \log p_{i,k} \Big] + \frac{1}{2} ||\beta||^2,
\end{align*}
where $\beta_{D_L \times K}=\{\beta_k\}_{k=1}^K$ is the output weight for the $K$ classes, $p_{i} = \mathrm{softmax}(g_{i})$ is the length-$K$ vector of class probability, and $g_i$ is the multi-class logits defined as $g_{i} = \beta_{D_L \times K} \Phi^\top_{i, 1 \times D_L}$.

Then, computing the Hessian with respect to $-\log p(\beta|\Dsc)$ yields \citep{kunstner2019limitations}:

\begin{align*}
    \hat{\bH}_{K D_L \times K D_L} 
    &= 
    \sum_{i=1}^n
    [\nabla_\beta g_i]^\top_{KD_L \times K} [\diag(p_i) - p_i p_i^\top]_{K \times K} [\nabla_\beta g_i]_{K \times KD_L} + \bI_{KD_L} \\
    &= [\Phi_i \otimes \bI_{K}] [\diag(p_i) - p_i p_i^\top] [\Phi_i^\top  \otimes \bI_{K}] + \bI_{KD_L} \\
    &= 
    \sum_{i=1}^n
    \begin{bmatrix} 
    \Phi_i & \dots & 0 \\ & \ddots & \\ 0 & \dots & \Phi_i
    \end{bmatrix}_{KD_L \times K} [\diag(p_i) - p_i p_i^\top]_{K \times K}
     \begin{bmatrix} 
    \Phi_i^\top & \dots & 0 \\ & \ddots & \\ 0 & \dots & \Phi_i^\top 
    \end{bmatrix}_{K \times K D_L} + \bI_{KD_L}.
\end{align*}
This leads to a $K D_L \times K D_L$ matrix with $D_L \times D_L$ diagonal and off-diagonal blocks as:
\begin{align*}
    \hat{\bH}_{k,k'} = 
    \begin{cases}
    \sum_{i=1}^n p_{i,k} (1-p_{i,k}) \Phi_i\Phi_i^\top + \bI_{D_L}  & \mbox{if } k=k'\\
    \sum_{i=1}^n -p_{i,k} p_{i,k'} \Phi_i\Phi_i^\top & \mbox{otherwise.}
    \end{cases}    
\end{align*}
Consequently, the Laplace posterior for each class is:
\begin{align}
\beta_k \sim MVN(\hat{\beta}_k, \hat{\bSigma}_k=\hat{\bH}_k^{-1}), 
\quad \mbox{where} \quad 
\hat{\bH}_k = \sum_{i=1}^n p_{i,k} (1-p_{i,k}) \Phi_i\Phi_i^\top + \bI_{D_L}.
\label{eq:multiclass_likelihood_app}
\end{align}
In the case where computing the class-specific covariance matrix is either infeasible (e.g., the number of output classes is simply too large) or not of interest (e.g., we are just interested in computing an scalar uncertainty statistic for an input example), one can consider computing an upper-bound of the covariance matrices $\hat{\bSigma}_k$ in (\ref{eq:multiclass_likelihood_app}) as:
\begin{align}
\hat{\bSigma}=\hat{\bH}^{-1} \quad \mbox{where} \quad 
\hat{\bH} = \sum_{i=1}^n p^*_{i} (1-p^*_{i}) \Phi_i\Phi_i^\top + \bI_{D_L},
\label{eq:multiclass_cov_upperbound_app}
\end{align}
where $p^*_i = \max_k(p_{i,1}, \dots, p_{i,k})$ is the maximum class probability. From an  information theoretic perspective, quantifying model uncertainty using (\ref{eq:multiclass_cov_upperbound_app}) can be understood adopting the maximum entropy distribution among the family of all class-specific distributions. To ensure computational feasibility for tasks with high-dimensional output (e.g., CIFAR100, CLINC, and ImageNet which correspond to 100, 150 and 1000 classes), we use (\ref{eq:multiclass_cov_upperbound_app}) in our experiments.

\subsection{Hyperparameter Configuration}
\label{sec:hyper_app}

\gls{SNGP} is composed of two components: Spectral Normalization (SN) and Gaussian Process (GP) layer, both are available at the open-source \texttt{edward2} probabilistic programming library \footnote{\url{https://github.com/google/edward2}}. 
\begin{enumerate}[leftmargin=2em]
    \item \textbf{Spectral normalization} contains two hyperparameters: the number of power iterations and the upper bound for spectral norm (i.e., $c$ in Equation (\ref{eq:spec_norm})). In our experiments, we find it is sufficient to fix power iteration to 1. The value for the spectral norm bound $c$ controls the trade-off between the expressiveness and the distance awareness of the residual block, where a small value of $c$ may shrink the residual block back to identity mapping hence harming the expressiveness, while a large value of $c$ may lead to the loss of bi-Lipschitz property (Proposition \ref{thm:resnet_lipschitz}). Furthermore, the proper range of $c$ depends on the layer type: for dense layers (e.g., the intermediate and the output dense layers of a Transformer), it is sufficient to set $c$ to a value between $(0.95, 1)$. For the convolutional layers, the norm bound needs to be set to a larger value to not impact the model's predictive performance. This is likely caused by the fact that the current spectral normalization technique does not have a precise control of the true spectral norm of the convolutional kernel, in conjuction with the fact that the other regularization mechanisms (e.g., BatchNorm and Dropout) may rescale a layer's spectral norm in unexpected ways \citep{gouk_regularisation_2018, miyato_spectral_2018}. In general, we recommend performing a grid search for $c \in \{0.9, 0.95, 1, 2, ...\}$ to identify the smallest possible values of $c$ that still retains the predictive performance of the original model. In the experiments, we set the norm bound to $c=6$ for a WideResNet model.
    \item \textbf{The Gaussian process layer} (Equation \ref{eq:rff_lr}) contains 3 hyperparameters, which 
    are (1) the \textit{hidden dimension} ($D_{L}$, i.e., the number of random features), (2) the \textit{length-scale parameter} $l$ for the \gls{RBF} kernel, 
    and (3) the kernel amplitude $\sigma$. In the experiments, we find the model's performance to be not very sensitive to the hidden dimension or the length-scale parameter. Setting $D_L$ in a range between $[512, 2048]$ and the length-scale $l=2.0$ are sufficient in most cases. This is likely due to the fact that, contrary to the classic \gls{GP} case without \gls{DNN}, the DNN hidden mapping in SNGP is capable of adjust itself to adapt to the kernel parameters during SGD learning. Finally, as discussed in the main text, the model's calibration performance is sensitive to the kernel amplitude $\sigma$, since it functions in a manner that is similar to the temperature parameter in temperature scaling \citep{guo_calibration_2017}.
    \item \textbf{The posterior covariance} (Equation \ref{eq:gp_posterior_approx}) does not contain hyperparameter. However, if one wishes to use a moving average estimator for the covariance matrix, i.e.,
    $$\Sigma_{k, 0}^{-1} = s*\bI, 
    \quad 
    \Sigma_{k, t}^{-1} = m * \Sigma_{k, t-1}^{-1} + (1-m) * \sum_{i=1}^M \hat{p}_{ik}(1-\hat{p}_{ik})\Phi_i\Phi_i^\top.$$
    Then there will be two additional hyperparameters: the ridge factor $s$ and the discount factor $m$. The ridge factor $s$ serves to control the stability of matrix inverse (if the number of sample size $n$ is small), and $m$ controls how fast the moving average update converges to the population value $\Sigma_{k}=s\bI + \sum_{i=1}^n \hat{p}_{ik}(1-\hat{p}_{ik})\Phi_i\Phi_i^\top$. Similar to other moving-average update method, these two parameters can impact the quality of learned covariance matrix in non-trivial ways. In general, we recommend conducting some small scale experiments on the data to validate the learning quality of the moving average update in approximating the population covariance. In the experiments, we set $s=0.001$ and $m=0.999$, which is sufficient for our setting where the number of minibatch steps per epoch is large. 
    We use the exact update formula (\ref{eq:gp_posterior_approx}) in the CIFAR and Genomics experiments, and use the moving average update as described above for larger tasks such as ImageNet.
\end{enumerate}

In summary, when applying SNGP method to a new dataset it is sufficient to sweep the spectral norm bound $c$ and kernel amplitude parameter $\sigma$ on a holdout validation dataset, and fix all other parameters to their default values (Table \ref{tb:hyper_param_default}). We report the values $(c, \sigma)$ for each experiment in Section \ref{sec:exp_detail}.

\begin{table}[ht]
\centering
\scalebox{0.80}{
\begin{tabular}{|c|c|c|c|c|c|}
\toprule
 \multicolumn{2}{|c|}{Spectral Normalization}  &  
 \multicolumn{2}{|c|}{Gaussian Process Layer}  &  
 \multicolumn{2}{|c|}{Covariance Matrix}  \\ 
\midrule
Power Iteration & 1  &  
Hidden Dimension & 2048 {\tiny (BERT)} / 1024 {\tiny (Otherwise)} &
(Optional) & 
\\ 
\textbf{Spectral Norm Bound} & Model dependent  & 
Length-scale Parameter &  2.0  &
\textbf{Ridge Factor} & 0.001
\\ 
&  & \textbf{Kernel amplitude} & Task dependent & 
\textbf{Discount Factor} & 0.999 \\ 
\bottomrule
\end{tabular}
}
\caption{Hyperparameters of SNGP used in the experiments, where important hyperparameters are highlighted in bold.}
\label{tb:hyper_param_default}
\end{table}

As an aside, we also implemented two additional functionalities for GP layers:  \textit{input dimension projection} and \textit{input layer normalization}. The input dimension project serves to project the hidden dimension of the penultimate layer $D_{L-1}$ to a lower value $D'_{L-1}$ (using a random Gaussian matrix $\bW_{D_{L-1} \times D'_{L-1}}$), it can be projected down to a smaller dimension. \textit{Input layer normalization} applies Layer Normalization to the input hidden features, which is akin to performing \gls{ARD}-style variable selection to the input features. 
Ablation studies revealed that the model performance is not sensitive to these changes.

\section{Formal Statements}
\label{sec:formal}

\paragraph{Minimax Solution to Uncertainty Risk Minimization.}
\label{sec:minimax_app}

The expression in (\ref{eq:minimax_solution}) seeks to answer the following question: \textit{assuming we know the true domain probability $p^*(\bx \in \Xsc_{\texttt{IND}})$, and given a model $p(y|\bx, \bx\in \Xsc_{\texttt{IND}})$ that we have already learned from data, what is the best solution we can construct to minimize the minimax objective (\ref{eq:minimax_loss})?} The interest of this conclusion is not to construct a practical algorithm, but to highlight the theoretical necessity of taking into account the domain probability in constructing a good solution for uncertainty quantification. If the domain probability is not necessary, then the expression of the unique and optimal solution to the minimax probability should not contain $p^*(\bx \in \Xsc_{\texttt{IND}})$ even if it is available. However the expression of (\ref{eq:minimax_solution}) shows this is not the case.

To make the presentation clear, we formalize the statement about (\ref{eq:minimax_solution}) into the below  proposition:
\begin{proposition}[Minimax Solution to Uncertainty Risk Minimization] $ $\\
Given:
\begin{enumerate}[leftmargin=2em]
    \item[(a)] $p(y|\bx, \bx \in \Xsc_{\textup{\texttt{IND}}})$ the model's predictive distribution learned from data $\Dsc=\{y_i, \bx_i\}_{i=1}^N$
    \item[(b)] $p^*(\bx \in \Xsc_{\textup{\texttt{IND}}})$ the true domain probability,
\end{enumerate}
then there exists an unique optimal solution to the minimax problem (\ref{eq:minimax_loss}), and it can be constructed using (a) and (b) as:
\begin{align}
    p(y |\bx) 
    &= 
    p(y|\bx, \bx \in \Xsc_{\textup{\texttt{IND}}}) \times p^*(\bx \in \Xsc_{\textup{\texttt{IND}}}) 
    + 
    p_{\textup{\texttt{uniform}}}(y|\bx, \bx \not\in \Xsc_{\textup{\texttt{IND}}}) \times p^*(\bx \not\in \Xsc_{\textup{\texttt{IND}}})
    \label{eq:minimax_solution_app}
\end{align}
where $p_{\textup{\texttt{uniform}}}(y|\bx, \bx \not\in \Xsc_{\textup{\texttt{IND}}})=\frac{1}{K}$ is a discrete uniform distribution for $K$ classes.
\label{thm:minimax}
\end{proposition}

As discussed in Section \ref{sec:minimax}, the solution (\ref{eq:minimax_solution_app}) is not only optimal for the minimax Brier risk, but is in fact optimal for a wide family of strictly proper scoring rules known as the (separable) \textit{Bregman score} \citep{parry_proper_2012}:
\begin{align}
    s(p, p^*|\bx)= \sum_{k=1}^K 
    \Big\{
    [p^*(y_k|\bx) - p(y_k|\bx)]\psi'(p^*(y_k|\bx)) - \psi(p^*(y_k|\bx))
    \Big\}
    \label{eq:bregman_score}
\end{align}
where $\psi$ is a strictly concave and differentiable function. Bregman score reduces  to the log score when $\psi(p)=p * \log (p)$, and reduces to the Brier score when $\psi(p)=p^2 - \frac{1}{K}$. 

Therefore we will show (\ref{eq:minimax_solution_app}) for the Bregman score. The proof relies on the following key lemma:

\begin{lemma}[$p_{\textup{\texttt{uniform}}}$ is Optimal for Minimax Bregman Score in $\bx \not\in \Xsc_{IND}$] $ $\\
Consider the Bregman score in (\ref{eq:bregman_score}). At a location $\bx \not\in \Xsc_{IND}$ where the model has no information about $p^*$ other than $\sum_{k=1}^K p(y_k|\bx) = 1$, the solution to the minimax problem
$$\inf_{p\in \Psc} \sup_{p^* \in \Psc^*} s(p, p^*|\bx)$$
is the discrete uniform distribution, i.e., $p_{\textup{\texttt{uniform}}}(y_k|\bx)=\frac{1}{K} \;\;\; \forall k \in \{1, \dots, K\}$.
\label{thm:minimax_lemma}
\end{lemma}
The proof for Lemma \ref{thm:minimax_lemma} is in Section \ref{sec:minimax_lemma_proof}. It is worth noting that Lemma \ref{thm:minimax_lemma} only holds for a \textit{strictly} proper scoring rule \citep{gneiting_probabilistic_2007}. For a non-strict proper scoring rule (e.g., the \gls{ECE}), there can exist infinitely many optimal solutions, making the minimax  problem ill-posed. 

We are now ready to prove Proposition \ref{thm:minimax}:
\noindent{\it Proof.\quad}
Denote $\Xsc_{\texttt{OOD}} = \Xsc / \Xsc_{\texttt{IND}}$.
Decompose the overall Bregman risk by domain:
\begin{align*}
    S(p, p^*) 
    &= 
    E_{x \in \Xsc}\big(s(p, p^*|\bx )\big) = 
    \int_{\Xsc} s\big(p, p^*|\bx \big)p^*(\bx) d\bx \\
    & = 
    \int_{\Xsc} s\big(p, p^*|\bx \big) * \big[
    p^*(\bx|\bx \in \Xsc_{\texttt{IND}})p^*(\bx \in \Xsc_{\texttt{IND}}) + 
    p^*(\bx|\bx \in \Xsc_{\texttt{OOD}})p^*(\bx \in \Xsc_{\texttt{OOD}}) \big]
    d\bx \\
    &= E_{\bx \in \Xsc_{\texttt{IND}}}\big(s(p, p^*|\bx )\big)
    p^*(\bx \in \Xsc_{\texttt{IND}}) +
    E_{\bx \in \Xsc_{\texttt{OOD}}}\big(s(p, p^*|\bx )\big)
    p^*(\bx \in \Xsc_{\texttt{OOD}}) \\
    &= S_{\texttt{IND}}(p, p^*) * p^*(\bx \in \Xsc_{\texttt{IND}}) +
    S_{\texttt{OOD}}(p, p^*) * p^*(\bx \in \Xsc_{\texttt{OOD}}).
\end{align*}
where we have denoted $S_{\texttt{IND}}(p, p^*)=E_{\bx \in \Xsc_{\texttt{IND}}}\big(s(p, p^*|\bx )\big)$ and $S_{\texttt{OOD}}(p, p^*)=E_{\bx \in \Xsc_{\texttt{OOD}}}\big(s(p, p^*|\bx )\big)$.

Now consider decomposing the sup risk $\sup_{p^*} S(p, p^*)$ for a given $p$. Notice that sup risk $\sup_{p^*} S(p, p^*)$ is separable by domain for any $p \in \Psc$. This is because $S_{\texttt{IND}}(p, p^*)$ and $S_{\texttt{OOD}}(p, p^*)$ has disjoint support, and we do not impose assumption on $p^*$: 
\begin{align*}
    \sup_{p^*} S(p, p^*) = 
    \sup_{p^*} \big[S_{\texttt{IND}}(p, p^*)\big] * 
    p^*(\bx \in \Xsc_{\texttt{IND}}) +
    \sup_{p^*} \big[S_{\texttt{OOD}}(p, p^*)\big] * 
    p^*(\bx \in \Xsc_{\texttt{OOD}})
\end{align*}
We are now ready to decompose the minimax risk $\inf_{p}\sup_{p^*} S(p, p^*)$. Notice that the minimax risk is also separable by domain due to the disjoint in support:
\begin{align}
    \inf_{p}\sup_{p^*} S(p, p^*) 
    &= 
    \inf_{p}\Big[ 
    \sup_{p^*} \big[S_{\texttt{IND}}(p, p^*)\big] * 
    p^*(\bx \in \Xsc_{\texttt{IND}}) +
    \sup_{p^*} \big[S_{\texttt{OOD}}(p, p^*)\big] * 
    p^*(\bx \in \Xsc_{\texttt{OOD}})
    \Big] \nonumber \\
    &= \inf_{p}\sup_{p^*} \big[S_{\texttt{IND}}(p, p^*)\big] * 
    p^*(\bx \in \Xsc_{\texttt{IND}}) + 
    \inf_{p}\sup_{p^*} \big[S_{\texttt{OOD}}(p, p^*)\big] * 
    p^*(\bx \in \Xsc_{\texttt{OOD}}),
    \label{eq:minimax_decomposed}
\end{align}
also notice that the in-domain minimax risk $\inf_{p}\sup_{p^*} \big[S_{\texttt{IND}}(p, p^*)\big]$ is fixed due to condition $(a)$. \\

Therefore, to show that (\ref{eq:minimax_solution_app}) is the optimal and unique solution to (\ref{eq:minimax_decomposed}), we only need to show $p_{\texttt{uniform}}$ is the optimal and unique solution to $\inf_{p}\sup_{p^*} \big[S_{\texttt{OOD}}(p, p^*)\big]$. To this end, notice that for a given $p$:
\begin{align}
    \sup_{p^*\in \Psc^*} \big[S_{\texttt{OOD}}(p, p^*)\big] 
    = 
    \int_{\Xsc_{\texttt{OOD}}} 
    \sup_{p^*}[s(p, p^*|\bx)] p(\bx|\bx\in \Xsc_{\texttt{OOD}}) d\bx,
\end{align}
due to the fact that we don't impose assumption on $p^*$ (therefore $p^*$ is free to attain the global supreme by maximizing $s(p, p^*|\bx)$ at every single location $\bx \in \Xsc_{\texttt{OOD}}$). Furthermore, there exists $p$ that minimize $\sup_{p^*} s(p, p^*|\bx)$ at every location of $\bx\in \Xsc_{\texttt{OOD}}$, then it minimizes the integral \citep{berger_statistical_1985}. By Lemma \ref{thm:minimax_lemma}, such $p$ exists and is unique, i.e.:
\begin{align*}
    p_{\textup{\texttt{uniform}}} = \underset{p\in \Psc}{\mathrm{arginf}}\sup_{p^*\in \Psc^*} S_{\texttt{OOD}}(p, p^*).
\end{align*}
In conclusion, we have shown that $p_{\textup{\texttt{uniform}}}$ is the unique solution to $\inf_{p}\sup_{p^*} S_{\texttt{OOD}}(p, p^*)$. Combining with condition (a)-(b), we have shown that the unique solution to (\ref{eq:minimax_decomposed}) is 
(\ref{eq:minimax_solution_app}).
\hfill\BlackBox

\section{Experiment Details}
\label{sec:exp_app}


\subsection{Model Configuration}
\label{sec:exp_detail}

For CIFAR-10 and CIFAR-100, we followed the original Wide ResNet work to apply the standard data augmentation (horizontal flips and random crop-ping with 4x4 padding) and used the same hyperparameter and training setup \citep{zagoruyko_wide_2017}. The only exception is the learning rate and training epochs, where we find a smaller learning rate ($0.04$ for CIFAR-10 and $0.08$ for CIFAR100, v.s. $0.1$ for the original WRN model) and longer epochs ($250$ for SNGP v.s. $200$ for the original WRN model) leads to better performance. \

For CLINC OOS intent understanding data, we pre-tokenized the sentences using the standard BERT tokenizer\footnote{\url{https://github.com/google-research/bert}} with maximum sequence length 32, and created standard binary input mask for the BERT model that returns 1 for valid tokens and 0 otherwise. Following the original BERT work, we used the Adam optimizer with  weight decay rate $0.01$ and warmup proportion $0.1$. We initialize the model from the official BERT$_{\texttt{Base}}$ checkpoint\footnote{\url{https://storage.googleapis.com/bert_models/2020_02_20/uncased_L-12_H-768_A-12.zip}}.
For this fine-tuning task, we using a smaller step size ($5e-5$ for SNGP .v.s. $1e-4$ for the original BERT model) but shorter epochs ($40$ for SNGP v.s. $150$ for the original BERT model) leads to better performance. When using spectral normalization, we set the hyperparameter $c=0.95$ and apply it to the pooler dense layer of the classification token. We do not apply spectral normalization to the hidden transformer layers, as we find the pre-trained BERT representation is already competent in preserving input distance due to the masked language modeling training, and further regularization may in fact harm its predictive and calibration performance. 

For Genomics sequence data, we consider a 1D CNN model following the prior work \citep{ren2019likelihood}. Specifically, the model is composed by one convolutional layer of 2,000 filters of length 20, one max-pooling layer, one dense layer of 2,000 units, and a final dense layer with softmax activation for predicting class probabilities. To build the SNGP model, we add spectral normalization to both the convolutional layer and the dense layer, and we replace the last layer with Gaussian process layer. For MC Dropout, we add filter-wise dropout before the convolutional layer with dropout rate 0.1. For Deep Ensemble, we ensemble 5 models trained based on random initialization of network parameters and random shuffling of training inputs.
The model is trained using the batch size 128, the learning rate 1e-4, and Adam optimizer. The training step is set for 1 million, but we choose the best step when validation loss is at the lowest value. Due to the large size of test OOD dataset, we randomly select 100,000 OOD samples to pair with the same number of in-distribution samples.

For each of the experiment, we fix the SNGP hyperparameters to their recommended values as in Table \ref{tb:hyper_param_default}, and sweep the Spectral Norm Bound $m$ and the kernel amplitude $\sigma$ with respect to the negative log likelihood on the validation data. The final values found for each experiment is summarized at Table \ref{tb:hyper_param}. As shown, the optimal value for spectral norm bound usually depends on the layer type (1D Convolution v.s. 2D Convolution v.s. Dense), while the optimal value of kernel amplitude seems to be sensitive to the data modality, the model type, and also the type of covariance estimator (i.e., exact estimator v.s. moving average estimator) being used
(also see earlier discussion in Section \ref{sec:hyper_app}).

\begin{table}[ht]
\centering
\scalebox{0.80}{
\begin{tabular}{|c|c|c|}
\toprule
Task & Spectral Norm Bound & Kernel Amplitude \\
\midrule
CIFAR-10 & 6.0 & 20 / 30 (+Ensemble) / 7.5 (+AugMix) \\
CIFAR-100 & 6.0 & 7.5  / 5.0 (+Ensemble) / 1 (+AugMix)  \\
ImageNet & 6.0 & 1. \\
CLINC & 0.95 & 0.1 \\
Genomics & 10.0 & 0.001 \\
\bottomrule
\end{tabular}
}
\caption{Hyperparameters of SNGP used in the experiments.}
\label{tb:hyper_param}
\end{table}

All models are implemented in TensorFlow and are trained on 8-core Cloud TPU v2 with 8 GiB of high-bandwidth memory (HBM) for each TPU core. We use batch size 32 per core.

\subsection{Evaluation} 
\label{sec:ood_metrics}
For CIFAR-10 and CIFAR-100, we evaluate the model's predictive accuracy and calibration error under both clean and corrupted versions of the CIFAR testing data. The corrupted data, termed CIFAR10-C, includes 15 types of corruptions, e.g., noise, blurring, pixelation, etc, over 5 levels of corruption intensity  \citep{hendrycks_benchmarking_2019}. 

We assess the model's calibration performance using the empirical estimate of \gls{ECE}: $\hat{ECE}=\sum_{m=1}^M \frac{|B_m|}{n} |acc(B_m) - conf(B_m)|$ which estimates the difference in model's accuracy and confidence by partitioning model prediction into $M$ bins $\{B_m\}_{m=1}^M$ \citep{guo_calibration_2017}. In this work, we choose $M=15$. 

We also evaluate the model performance in \gls{OOD} detection by using the CIFAR-10/CIFAR-100 model's uncertainty estimate as a predictive score for \gls{OOD} classification, where we consider a standard \gls{OOD} task by testing CIFAR-10/CIFAR-100 model's ability in detecting samples from the \glsfirst{SVHN} dataset \citep{netzer_reading_2011}, and a more difficult \gls{OOD} task by testing CIFAR-10's ability in detecting samples from the CIFAR-100 dataset, and vice versa. 
For CIFAR dataset, since the input data is normalized by the sample mean and variance, we perform the same normalization for the test OOD data.
We evaluate the OOD uncertainty scores for the inputs in in-distribution test set, and the inputs in OOD test set, and we use AUROC to measure the separation between the two sets based on the OOD uncertainty score. 

For all models, we compute the confidence score (or reverse equivalently OOD uncertainty score) for an input $\bx$ as the Maximum of Softmax Probability (MSP) \citep{hendrycks_baseline_2017}, which is $\max_k(\softmax(g_k(\bx))), k=1,2,,K$, given logits for $K$ classes $\{g_k(\bx)\}_{k=1}^K$. The results presented in the main text are based on MSP. 
We study other commonly used uncertainty scores, including the so-called \textit{Dempster-Shafer metric} \citep{sensoy_evidential_2018}, the Mahalanobis distance \cite{lee2018simple}, and the relative Mahalanobis distance \cite{ren2021simple}, to compare with the MSP.

The \textit{Dempster-Shafer metric} computes its uncertainty for a test example $\bx_i$ as:
\begin{align}
    u(\bx) = \frac{K}{K + \sum_{k=1}^K \exp\big(h_k(\bx)\big)},
\end{align}
and the confidence score is defined as $\mathrm{score}_{\mathrm{DS}}(\bx) = 1 - u(\bx)$,
As shown, for a distance-aware model where the magnitude of the logits reflects the distance from the observed data manifold, $u(\bx_i)$ can be a more effective metric since it is monotonic with respect to the magnitude of the logits, similar to the energy-based OOD score in \cite{liu2020energy}. On the other hand, the maximum probability $p_{\texttt{max}}=\max_k \; \exp(g_k)/\sum_{k=1}^K \exp\big(g_k(\bx_i)\big)$ does not take advantage of this information since it normalizes over the exponentiated logits.

\textit{Mahalanobis distance} (MD) method was proposed by \citet{lee2018simple} to fit a Gaussian distribution to the class-conditional embeddings and use the Mahalanobis distance for OOD detection. 
Let $h(\bx)$ denote the embedding (e.g. the penultimate layer before computing the logits) of an input $\bx$.
We fit a Gaussian distribution to the embeddings of the training data, computing per-class mean 
$\bmu_k = \tfrac{1}{N_k} \sum_{i:y_j=k} h(\bx_j)$
and a shared covariance matrix $\Sigma = \tfrac{1}{N}\sum_{k=1}^K \sum_{i:y_j=k}\bigl(h(\bx_j)-\bmu_k \bigr)\bigl(h(\bx_j)-\bmu_k\bigr)^{\top}.$
The confidence score (negative of the Mahalanobis distance) is then computed as:  
$\mathrm{score}_{\mathrm{Maha}}(\bx) = - \min_k \left ( 
  \bigl(h(\bx)-\bmu_k\bigr) \Sigma^{-1} \bigl(h(\bx)-\bmu_k\bigr)^\top
  \right) = -\min_k \left (\text{MD}_k(\bx) \right).$ 
  
\textit{Relative Mahalanobis distance} (RMD) was proposed by \citet{ren2021simple} to fix the degradation of the raw Mahalanobis distance in near-OOD scenarios.  
The confidence score
$\mathrm{score}_{\mathrm{RMaha}}(\bx) = - \min_k \left( \text{MD}_k(\bx) - \text{MD}_0(\bx) \right)$, where $\text{MD}_0(\bx)$ is the Mahalanobis distance to a  background Gaussian distribution fitted to the entire training data not considering the class labels. 

Table \ref{tab:cifar-ood} shows the OOD performance for CIFAR dataset based on the above four different confidence scores. 
First, it shows that Dempster Shafer score in general has better performance than MSP.
Second, we noticed that Mahalanobis distance has much worse performance on GP based models, while Relative Mahalanobis distance corrects for the degradation. For example, for CIFAR-10 vs. CIFAR-100 task, Mahalanobis distance based on SNGP model's embeddings has only 0.742 AUROC, while other OOD methods achieve around 0.90 AUROC based on SNGP model, including Relative Mahalanobis distance.
This suggests that the GP based models probably preserved background features which confounds the raw Mahalanobis distance score, and the relative Mahalanobis distance corrects for the background features and fix the performance. 
It is interesting to see that the Relative Mahalanobis distance has the best performance for the most challenging near-OOD task CIFAR-100 vs. CIFAR-10 for all models. 

{\color{black}We also additionally evaluate another two simple OOD datasets, random Gaussian noise and texture dataset (i.e. DTD), used in the prior work \citep{hendrycks_deep_2018,liang2017enhancing,liu2020energy}. The results are included in Table 11. As shown, SNGP provides the best performance when using the default MSP method for OOD detection: 0.999 AUROC for detecting random Gaussian noise, and 0.959 AUROC for detecting DTD. 
When combined with more advanced OOD detection signals, such as Dempster Shafer, the performance can be further improved to 1.000 for random Gaussian and 0.988 for DTD. 
Furthermore, for the uncertainty metrics whose performance hinders on the qualities of both the hidden representation and the last-layer (i.e., MSP and Dempster-Shafer), the SNGP model attains the strongest performance when compared to its ablated counterparts. 
Finally, we observed some vanilla methods (e.g., Mahalanobis distance based on vanilla DNN) also achieves strong performance for these simple datasets. However, this advantage starts to break down on the more difficult datasets (e.g., CIFAR100 v.s. SVHN)}.

In conclusion, when comparing to the baseline approaches, SNGP provides the best out-of-box performance when using the default MSP method for OOD detection (Section \ref{sec:cifar}), and it can be combined with more advanced OOD detection signals to further improve performance.

\begin{table}[ht]
\centering
\scalebox{0.9}{%
\begin{tabular}{ccccc}
\toprule
\multicolumn{1}{c}{}             & \textbf{MSP}                            & \textbf{Dempster Shafer}                & \multicolumn{1}{l}{\textbf{Mahalanobis}} & \multicolumn{1}{l}{\textbf{Relative Mahalanobis}}\\ \midrule
\multicolumn{1}{c}{\multirow{1}{*}{}} & \multicolumn{4}{c}{Near-OOD} \\ \midrule
\multicolumn{1}{c}{\multirow{1}{*}{}} & \multicolumn{4}{c}{CIFAR-10 vs. CIFAR-100} \\ \midrule
DNN      & 0.893 $\pm$ 0.003 & \textbf{0.894 $\pm$ 0.004} & 0.891 $\pm$ 0.002 & 0.890 $\pm$ 0.002   \\ 
DNN-SN   & \textbf{0.893 $\pm$ 0.002} & \textbf{0.893 $\pm$ 0.003} & 0.892 $\pm$ 0.002 & 0.889 $\pm$ 0.002  \\ 
DNN-GP   & \textbf{0.903 $\pm$ 0.003} & 0.902 $\pm$ 0.005 & 0.767 $\pm$ 0.016 & 0.899 $\pm$ 0.002  \\
SNGP & \textbf{0.905 $\pm$ 0.002} & 0.903 $\pm$ 0.005 & 0.742 $\pm$ 0.011 & 0.899 $\pm$ 0.002   \\ \midrule
\multicolumn{1}{c}{\multirow{1}{*}{}} & \multicolumn{4}{c}{CIFAR-100 vs. CIFAR-10} \\ \midrule
DNN      & 0.795 $\pm$ 0.001      & 0.804 $\pm$ 0.002     & 0.780 $\pm$ 0.003       & \textbf{0.809 $\pm$ 0.002}        \\
DNN-SN   & 0.793 $\pm$ 0.003      & 0.801 $\pm$ 0.002     & 0.785 $\pm$ 0.003       & \textbf{0.809 $\pm$ 0.002}        \\
DNN-GP   & 0.797 $\pm$ 0.001      & 0.782 $\pm$ 0.004     & 0.559 $\pm$ 0.005       & \textbf{0.804 $\pm$ 0.003}        \\
SNGP   & 0.798 $\pm$ 0.001      & 0.784 $\pm$ 0.003     & 0.576 $\pm$ 0.006       & \textbf{0.804 $\pm$ 0.002}     \\ \midrule   
\multicolumn{1}{c}{\multirow{1}{*}{}} & \multicolumn{4}{c}{Far-OOD} \\ \midrule
\multicolumn{1}{c}{\multirow{1}{*}{}} & \multicolumn{4}{c}{CIFAR-10 vs. SVHN } \\ \midrule
DNN      & 0.946 $\pm$ 0.005 & 0.954 $\pm$ 0.008 & \textbf{0.959 $\pm$ 0.006} & 0.924 $\pm$ 0.011  \\
DNN-SN   & 0.945 $\pm$ 0.005 & 0.958 $\pm$ 0.006 & \textbf{0.960 $\pm$ 0.005} & 0.929 $\pm$ 0.010  \\ 
DNN-GP   & 0.964 $\pm$ 0.006 & \textbf{0.984 $\pm$ 0.004} & 0.914 $\pm$ 0.012 & 0.932 $\pm$ 0.005  \\
SNGP & 0.960 $\pm$ 0.004 & \textbf{0.982 $\pm$ 0.004} & 0.915 $\pm$ 0.014 & 0.931 $\pm$ 0.016 \\ \midrule
\multicolumn{1}{c}{\multirow{1}{*}{}} & \multicolumn{4}{c}{CIFAR-100 vs. SVHN} \\ \midrule
DNN   & 0.799 $\pm$ 0.020 & \textbf{0.841 $\pm$ 0.019} & 0.764 $\pm$ 0.020 & 0.755 $\pm$ 0.022 \\
DNN-SN   & 0.798 $\pm$ 0.022 & \textbf{0.832 $\pm$ 0.021} & 0.752 $\pm$ 0.025 & 0.727 $\pm$ 0.025 \\
DNN-GP   & 0.835 $\pm$ 0.021 & \textbf{0.887 $\pm$ 0.025} & 0.550 $\pm$ 0.073 & 0.724 $\pm$ 0.033 \\
SNGP   & 0.846 $\pm$ 0.019 & \textbf{0.894 $\pm$ 0.020} & 0.533 $\pm$ 0.039 & 0.712 $\pm$ 0.021 \\
\midrule     
\multicolumn{1}{c}{\multirow{1}{*}{}} & \multicolumn{4}{c}{CIFAR-10 vs Random Gaussian} \\ \midrule
DNN    & 0.981 $\pm$ 0.014 & 0.990 $\pm$ 0.008 & \textbf{1.000 $\pm$ 0.000} & 0.990 $\pm$ 0.007  \\
DNN-SN & 0.976 $\pm$ 0.013 & 0.996 $\pm$ 0.003 & \textbf{1.000 $\pm$ 0.000} & 0.983 $\pm$ 0.013  \\
DNN-GP & 0.997 $\pm$ 0.003 & \textbf{1.000 $\pm$ 0.000} & 0.999 $\pm$ 0.000 & 0.968 $\pm$ 0.032  \\
SNGP   & 0.999 $\pm$ 0.002 & \textbf{1.000 $\pm$ 0.000} & 0.999 $\pm$ 0.000 & 0.982 $\pm$ 0.013 \\
\midrule
\multicolumn{1}{c}{\multirow{1}{*}{}} & \multicolumn{4}{c}{CIFAR-10 vs DTD} \\ \midrule
DNN    & 0.902 $\pm$ 0.020 & 0.915 $\pm$ 0.011 & \textbf{0.994 $\pm$ 0.001} & 0.906 $\pm$ 0.022  \\
DNN-SN & 0.909 $\pm$ 0.016 & 0.909 $\pm$ 0.024 & \textbf{0.994 $\pm$ 0.001} & 0.901 $\pm$ 0.028  \\
DNN-GP & 0.946 $\pm$ 0.008 & 0.982 $\pm$ 0.008 & \textbf{0.987 $\pm$ 0.002} & 0.925 $\pm$ 0.015  \\
SNGP   & 0.959 $\pm$ 0.009 & \textbf{0.988 $\pm$ 0.004} & 0.985 $\pm$ 0.003 & 0.938 $\pm$ 0.004  \\
\midrule
\end{tabular}
}
\caption{
OOD detection performance (AUROC) based on different confidence scores, including MSP, Dempster Shafer, Mahalanobis distance, and Relative Mahalanobis distance. The highest values in each row are highlighted. 
}
\label{tab:cifar-ood}
\end{table}






\clearpage
{\color{black}
\subsection{Theoretical Convergence to Optimal Behaviour}
\label{sec:optimal_sngp}
In this section, we discuss the asymptotic behaviour of the SNGP algorithm on OOD datapoints far from the training distribution, and show how the predictive distribution converges to the optimal distribution suggested by Equation (\ref{eq:minimax_solution}). We formalize this in the following proposition:
\begin{proposition}[SNGP Convergence to Optimal Behaviour]$ $ \\
\label{thm:optimal_sngp}
Given:
\begin{enumerate}[leftmargin=2em]
    \item[(a)] A distance-preserving hidden mapping $h: \mathcal{X} \rightarrow \mathcal{H}$, so that
    $$
    L_1 * d_X(\bx, \bx^*) \leq  \left\|h(\bx)-h\left(\bx^{*}\right)\right\|_2^2
    \leq L_2 * d_X(\bx, \bx^*),
    $$
    for $0 < L_1 < L_2$ two positive constants.
    \item[(a)] A dual form formulation of the SNGP model assuming a Laplace approximation over the posterior, with posterior mean and variance given by
    \begin{align*}
        m(\bx) &= \mathbf{k}^*(\bx)^{\top}(\mathbf{K}+\tau \mathbf{I})^{-1} \mathbf{y} \\
       \nu(\bx) &= \tau^{-1} *\left[k(\bx, \bx)-\mathbf{k}^*(\bx)^{\top}(\mathbf{K}+\tau \mathbf{I})^{-1} \mathbf{k}^*(\bx)\right]
    \end{align*}
    where $k(\bx, \bx)_{1 \times 1}=\phi(\bx)^{\top} \phi(\bx), \mathbf{k}^*(\bx)_{N \times 1}=\boldsymbol{\phi}(\bx)^{\top} \Phi^{\top} \text { and } \mathbf{K}_{N \times N}=\Phi \Phi^{\top}$ are kernel matrices approximating those under the RBF kernel $k\left(\bx, \bx^{\prime}\right) \propto \exp \left(-\left\|h(\bx)-h\left(\bx^{\prime}\right)\right\|_2^2\right)$.
    \item[(b)] The predictive distribution of the SNGP model for classification $p(y|\bx)$ given by
    \begin{align*}
        p(y \mid \bx) &= \int_{g \sim N(m(\bx), \nu(\bx))} \operatorname{softmax}(g) d g
    \end{align*}
\end{enumerate}
then as test points $\bx^*$ tend away from the training manifold $\bx$ (i.e $d_{X}(\bx, \bx^*) \rightarrow \infty$), the limit of the predictive distribution $p(y|\bx)$ is either exact equal to the optimal distribution $p_{\text{uniform}}$ as given by (\ref{eq:minimax_solution}) (under the mean-field approximation), or can be closely approximated by $p_{\textup{\texttt{uniform}}}$ (under the Monte Carlo approximation).
\end{proposition}

The proof for Proposition \ref{thm:optimal_sngp} follows:
\noindent{\it Proof.\quad}
Given condition (a), $h$ is a bi-Lipschitz, distance-preserving function, and therefore we can write down the asymptotic convergence of the RBF kernel $k(\bx, \bx^*)$ as follows
\begin{align*}
\lim_{d_{X}(\bx, \bx^*) \rightarrow \infty} k(\bx, \bx^*) 
&= 
\lim_{d_{X}(\bx, \bx^*) \rightarrow \infty} C \exp \left( 
- \left\|h(\bx)-h\left(\bx^{*}\right)\right\|_2^2 
\right) \\
&\leq 
\lim_{d_{X}(\bx, \bx^*) \rightarrow \infty} C 
\exp\left( 
- L_1 * d_{X}(\bx, \bx^*)
\right)
= 0,
\end{align*}
where $L_1$ and $C$ are constants, and the convergence is guaranteed by the sandwich theorem, since by the distance preservation property, $\left\|h(\bx)-h\left(\bx^{*}\right)\right\|_2^2 \geq L_1 \times d_X(\bx, \bx^*)$.

Leveraging the above result, we can now reason about the asymptotic behaviour of the posterior moments $(m(\bx^*), \nu(\bx^*))$:
\begin{align*}
    \lim_{d_{X}(\bx, \bx^*) \rightarrow \infty} m(\bx^*) &= \lim_{d_{X}(\bx, \bx^*) \rightarrow \infty} \mathbf{k}^*(\mathbf{x^*})^{\top}(\mathbf{K}+\tau \mathbf{I})^{-1} \mathbf{y} = 0\\
    \lim_{d_{X}(\bx, \bx^*) \rightarrow \infty} \nu(\bx^*) &= \lim_{d_{X}(\bx, \bx^*) \rightarrow \infty} \tau^{-1} *\left[k(\mathbf{x^*}, \mathbf{x^*})-\mathbf{k}^*(\mathbf{x^*})^{\top}(\mathbf{K}+\tau \mathbf{I})^{-1} \mathbf{k}^*(\bx)\right] \\ &= \tau^{-1} * k(\mathbf{x^*}, \mathbf{x^*}) = c 
\end{align*}
for some data-independent constant $c$. 

As a result, we can now consider the asymptotic behaviour of the predictive distribution $p(y|\bx)$ by leveraging the asymptotic behaviour of the moments. 

\paragraph{Mean-field Approximation.} First notice that by mean-field theorem, the sigmoid and softmax transformation of Gaussian is approximated as below \citep{bishop_pattern_2011, lu2020mean}:
\begin{align*}
p(y \mid \mathbf{x})
&= 
\sigma \left(\frac{m(\bx)}{\sqrt{1+\lambda * \nu(\bx)}}\right).
\end{align*}
Then, since $m(\bx^*) \rightarrow \bzero$ and $\nu(\bx^*) \rightarrow c$ as $d(\bx, \bx^*) \rightarrow \infty$, the predictive distribution $p(y|\bx^*)$ converges to the uniform distribution:
\begin{align}
     \lim_{d_{X}(\bx, \bx^*) \rightarrow \infty} p(y \mid \mathbf{x^{*}}) 
     = \sigma\left(\frac{m(\bx^*)}{\sqrt{1+\lambda * \nu(\bx^*)}}\right) 
     = \sigma(0) = p_{\textup{\texttt{uniform}}},
\end{align}
where the second equality follows from the continuous mapping theorem.

\paragraph{Monte Carlo Approximation.} Notice that since the Monte Carlo approximation of $p(y \mid \mathbf{x^{*}})$ is an unbiased estimator of the true integral, it is sufficient to investigate the asymptotic behavior of the integral:
\begin{align}
    \lim_{d_{X}(\bx, \bx^*) \rightarrow \infty} p(y \mid \mathbf{x^{*}}) &= \lim_{d_{X}(\bx, \bx^*) \rightarrow \infty} \int_{g \sim N(m(\mathbf{x^{*}}), \nu(\mathbf{x^{*}}))} \sigma(g) d g 
    \nonumber \\
    &= \int \lim_{d_{X}(\bx, \bx^*) \rightarrow \infty} \sigma(g) * f_N(g |m(\mathbf{x^{*}}), \nu(\mathbf{x^{*}})) d g 
    \nonumber \\
    &= \int \sigma(g) *  f_N(g |0, c) d g
    \label{eq:gaussian_softmax_app}
\end{align}
where $f_N(g |m, \nu)$ denotes the probability density function of a Gaussian-distributed random variable $g$ with mean $m$ and standard deviation $\nu$. In the above, the second equality (i.e., switching of limit and integral) follows by the bounded convergence theorem since the integrand $ \sigma(g) * f_N(g)$ is a bounded function. Intuitively, \Cref{eq:gaussian_softmax_app} is again a Gaussian-softmax integral whose expectation can be approximated closely via the mean-field approximation \citep{bishop_pattern_2011, lu2020mean}. That is, by following the same line of argument as above, we now that the limiting distribution $\lim_{d_{X}(\bx, \bx^*) \rightarrow \infty} p(y \mid \mathbf{x^{*}})$ of the Monte Carlo approximation can be closely approximated by a uniform distribution, i.e.,
\begin{align*}
 \lim_{d_{X}(\bx, \bx^*) \rightarrow \infty} p(y \mid \mathbf{x^{*}}) = E_{g \sim N(0,c)} \big[ \sigma(g) \big]
 \approx 
 \sigma(0) = p_{\textup{\texttt{uniform}}}.
\end{align*}
Alternatively, we can argue about convergence of \Cref{eq:gaussian_softmax_app} by appealing to the multivariate continuous mapping theorem. Specifically, since $(m(\bx^*), \nu(\bx^*)) \rightarrow (\bzero, c)$, we have $g \rightarrow g' \stackrel{d}{\sim} N(\bzero, c\bI)$ and $\sigma(g) \stackrel{d}{\sim} \sigma(g')$, where $\stackrel{d}{\sim}$ denotes the convergence in distribution. Consequently, $E(g) \rightarrow E(g')$. Now, for a $K$-dimensional $g'$, consider the $j^{th}$ coordinate of $log \sigma(g')$:
\begin{align*}
    \sigma(g')_j = \frac{exp(g'_j)}{\sum_{k=1}^K exp(g'_{k})},
\end{align*}
As shown, since $g'_{k} \stackrel{i.i.d.}{\sim} N(0, c)$, we expect $E(\sigma(g')_j)$ to be equal $\forall j \in \{1, \dots, K\}$. Consequently, denoting $E(\sigma(g')_k)=p' \; \forall k$ and notice $\sigma(g')$ sum to 1, we have:
\begin{align*}
    E(\sigma(g')_k)&=p' \quad \forall k=1, \dots, K, \\
    \sum_{k} E(\sigma(g')_k)&= 1,
\end{align*}
which leads to the unique solution of $p'=\frac{1}{K}$, i.e., $E(\sigma(g')) = p_{\textup{\texttt{uniform}}}$, which again shows:
\begin{align*}
    \lim_{d_{X}(\bx, \bx^*) \rightarrow \infty} p(y \mid \mathbf{x^{*}}) 
    = E(\sigma(g)) \rightarrow E(\sigma(g')) = p_{\textup{\texttt{uniform}}}.
\end{align*}
\hfill\BlackBox
}

\clearpage
\section{An Example Formalization of ``Semantic Distance".}
\label{sec:semantic_distance_formalization}
In this section, we develop an example formalization of the intuition notion of \textit{semantic distance} using languages from the metric embedding theory \citep{abraham2011advances, matouvsek2013lecture, chennuru2018measures}. Here, our goal is to supply an example formalization of this often intuitive notion, with then goal of facilitating a rigorous understanding of \Cref{sec:discussion} \citep{abraham2011advances, matouvsek2013lecture, chennuru2018measures}. Indeed, the term ``semantic distance" has a long history in the literature,  
and it is out of the scope of this work to provide an authoritative, all-encompassing mathematical construction that unifies its diverse usages across many fields such as manifold learning, representation learning, natural language processing, and cognitivle psychology
\citep{tenenbaum2000global, mohammad2006distributional, deselaers2011visual, hashimoto2016word, higgins2018towards, khemakhem2020variational, chandrasekaran2021evolution, reilly2022semantic}. 

\paragraph{Defining ``semantic space" in terms of a metric space.}

We consider the setting where all examples $\bx$ within a problem domain $\Xsc$ can be sufficiently described by a large collection of attributes, with each attribute being either discrete (e.g., types of entities that appeared in a image and their relationships) or continuous (e.g., color intensity or camera angle). The number of attributes $D$ is finite but allowed to be very large. Furthermore, we assume different attribute impacts the semantic similarity between examples to a different degree, so that the variations along only a subset of attributes constitute a meaningful difference between the examples.

We can formalize the above intuition terms of metric space \citep{rudin1976principles}. Specifically, we can assume the examples $\bx$ has a semantic representation $\bx$ that reside in a \textit{semantic space} $\Xsc$, which is a $P$-dimensional metric space with its dimensions correspond to the discrete and continuous attributes. Specifically, $\Xsc$ can be expressed as a product of $D$ \textit{attribute subspaces} $\{\Asc_j\}_{j=1}^D$:
\begin{align}
    \Xsc = \Asc_1 \times \Asc_2 \times \dots \times \Asc_D,
\end{align}
where each attribute subspace $(\Asc_j, d_j)$ is a metric space that corresponds to a continuous or discrete attribute, and is equipped with a well-defined metric $d_j$. For example, $\Asc_j$ can represent a \textit{continuous} attribute such as the color, which implies  $\Asc_j=\real^3$ and $d_j$ is the standard Euclidean metric for the RGB space. $\Asc_j$ can also represent a \textit{discrete} attribute such as entity types, which implies $\Asc_j$ is a discrete space that is supported on a large amount of candidate entities (e.g., from a knowledge graph), and the metric between entities $d_j$ can be defined by a certain graph metric with respect to a pre-established concept hierarchy (e.g., WordNet) \citep{chandrasekaran2021evolution}. As a result, we can write the semantic representation of an example $\bx \in S$ as $\bx=[\bx_1, \dots, \bx_D] \in S$ where $\bx_j \in \Asc_j$, and the differences between two examples $(\bx, \bx')$ in the $j^{th}$ attribute can be described as $d_j(\bx_j, \bx'_j)$.

Consequently, a ``semantic distance" can be defined in terms of a metric function $d_X: \Xsc \times \Xsc \rightarrow \real$ for the product space $\Xsc$, so that $(\Xsc, d_X)$ is a valid metric space. To this end, a proper choice of $d_X$ should satisfy the metric axioms (positivity, symmetry, and triangle inequality) while aligns well with the intuitive notion of ``semantic similarity" between examples. For example, consider the below definition of ${\color{black}d_X}$:
\begin{align}
    {\color{black}d_X}(\bx, \bx') = \sum_{j} w_j d_j(\bx_j, \bx_j') 
    + \sum_{j}\sum_k w_{jk} d_{j,k}(\bx_j, \bx_j')d_k(\bx_k, \bx_k'),
    \label{eq:semantic_distance}
\end{align}
i.e., ${\color{black}d_X}$ is a weighted sum of attribute-specific metrics $d_j$ and their pairwise products\footnote{It is also possible to define ${\color{black}d_X}$ with even higher-order products, e.g., $d_i(\bx_i, \bx_i')d_j(\bx_j, \bx_j')d_k(\bx_k, \bx_k')$, which we don't explore here for the simplicity of exposition.}.  Here $\bw = [w_1, w_2, \dots, w_D, w_{11}, w_{12}, \dots, w_{DD}]$ is the set of positive weights that sum to 1. We see the definition of semantic distance ${\color{black}d_X}$ in \Cref{eq:semantic_distance} is flexible, as it not only allows the domain expert to define what constitutes a ``semantically-meaningful difference" by assigning different weights among attributes, but also allows the attributes to interact to define the overall metric. We also see that ${\color{black}d_X}$ is a valid metric, as the positive $\bw$ guarantees positivity, and triangle inequality is closed under the summation and multiplication of positive terms\footnote{That is, if the attribute-specific metrics satisfy positivity and triangle inequality, then their sum and product also satisfy triangle inequality. For example, $d_i(\bx_i,\bx_i'') + d_j(\bx_j,\bx_j'') \leq (d_i(\bx_i,\bx_i')+d_i(\bx_i',\bx_i'')) + (d_j(\bx_j,\bx_j')+d_j(\bx_j',\bx_j''))= (d_i(\bx_i,\bx_i')+d_j(\bx_i,\bx_i')) +  (d_i(\bx_i',\bx_i'')+d_j(\bx_j',\bx_j'')).$ and $d_i(\bx_i,\bx_i'')d_j(\bx_j,\bx_j'') = (d_i(\bx_i,\bx_i')+d_i(\bx_i',\bx_i''))(d_j(\bx_j,\bx_j')+d_j(\bx_j',\bx_j''))\leq d_i(\bx_i,\bx_i')d_j(\bx_j,\bx_j') + d_i(\bx_i',\bx_i'')d_j(\bx_j',\bx_j'').$}.

\paragraph{From semantic space to data space via metric embedding.} So far, we have described a formal definition of \textit{semantic space} $\Xsc$ and the associated \textit{semantic distance} ${\color{black}d_X}$ in terms of a metric space $(\Xsc, {\color{black}d_X})$. Further, $(\Xsc, {\color{black}d_X})$ is constructed as the product of a collection of attribute metric spaces $\{(\Asc_j, d_j)\}_{j=1}^D$, so that the  coordinates of $\Xsc$ adopt meaningful interpretations in terms of well-defined attributes. In practice, we often do not have direct access to $(\Xsc, {\color{black}d_X})$, and can only work with its surface-form data representation $(\Ssc, d_S)$, where the coordinates of $\Ssc$ and the surface-form distance $d_S$ is less meaningful. However, the elements in $(\Xsc, {\color{black}d_X})$ still has a unique (i.e., one-to-one) correspondence with respect to $\bx$'s in the semantic space. For example, an image is often represented as a tensor of dimension $(W, H, C)$ (i.e., $\Xsc=\real^{W \times H \times C}$). However, by visually inspecting this tensor, a human can still discern the various attributes underlying the image, implying that an inverse mapping exists there exists from the surface-form data space $\Ssc$ to the semantic space $\Xsc$.

Formally, this means we can define an \textit{embedding function} $\psi: \Xsc \rightarrow \Ssc$, which is a mapping from the semantic space $(\Xsc, {\color{black}d_X})$ to the data space $(\Ssc, d_S)$. As a result, every example $\bx \in \Xsc$ adopts a surface-form representation $\textbf{s} = \psi(\bx) \in \Ssc$, which means when measuring based on the data space, the distance between a pair of examples $(\bx, \bx')$ becomes:
\begin{align*}
 d_S(\bx, \bx') \coloneqq d_S(\textbf{s}, \textbf{s}') 
 = d_S(\psi(\bx), \psi(\bx')).
\end{align*}

Consequently, the \textit{distortion} (introduced in the \Cref{sec:discussion} in the discussion section) between the semantic distance ${\color{black}d_X}$ and the surface-form distance $d_S$ can be expressed as:
\begin{align*}
    \rho(\bx, \bx') 
    = \frac{{\color{black}d_X}(\bx, \bx')}{d_S(\bs, \bs')} 
    =  \frac{{\color{black}d_X}(\bx, \bx')}{d_S(\psi(\bx), \psi(\bx'))}.
\end{align*}
As shown, the distortion is induced by both the embedding function $\psi$ and the difference in the metrics (i.e., ${\color{black}d_X} \, v.s. \, d_S$). From this perspective, when learning a neural network model $logit(\bx)= h(\bx)^\top \beta$ based on the observed data $\{(\bx_i, y_i)\}_{i=1}^n$, a semantic \textit{distance preserving} representation $h: \Xsc \rightarrow \Hsc$ should ideally satisfying the \textit{bi-Lipschitz} condition (\Cref{eq:dp}) with respect to ${\color{black}d_X}$:
\begin{align*}
    L_1 \times {\color{black}d_X}(\bx_1, \bx_2) \leq ||h(\bx_1) - h(\bx_2)||_H \leq L_2 \times {\color{black}d_X}(\bx_1, \bx_2),
\end{align*}
so that $||.||_H$ has a reasonable correspondence with the semantic distance.

\clearpage
\section{Proof}
\subsection{Proof of Proposition \ref{thm:resnet_lipschitz}}
\label{sec:resnet_lipschitz_proof}

The proof for Proposition \ref{thm:resnet_lipschitz} is an adaptation of the classic result of \citep{bartlett_representing_2018} to our current context:

\noindent{\it Proof.\quad}
First establish some notations. We denote $I(\bx)=\bx$ the identity function such that for $h(\bx)=\bx + g(\bx)$, we can write $g = h - I$. 
For $h: \Xsc \rightarrow \Hsc$, denote $||h||= \sup \Big\{ \frac{||f(\bx)||_H}{||\bx||_X} \mbox{ for } \bx \in \Xsc, ||\bx|| > 0 \Big\}$. Also denote the Lipschitz seminorm for a function $h$ as:
\begin{align}
    ||h||_L = \sup\bigg\{
    \frac{||h(\bx)-h(\bx')||_H}{{\color{black}d_X}(\bx, \bx')} \quad \mbox{for} \quad 
    \bx, \bx' \in \Xsc, \bx \neq \bx'
    \bigg\}
\end{align}
It is worth noting that by the above definitions, for two functions $(\bx' - \bx): \Xsc \times \Xsc \rightarrow \Xsc$ and $(h(\bx) - h(\bx')): \Xsc \times \Xsc \rightarrow \Hsc$ who shares the same input space, the Lipschitz inequality can be expressed using the $||.||$ norm, i.e., $||h(\bx) - h(\bx')||_H \leq \alpha {\color{black}d_X}(\bx, \bx')$ implies $||h(\bx') - h(\bx)|| \leq \alpha ||\bx - \bx'||$, and vice versa.\\

Now assume $\forall l$, $||g_l||_L=||h_l-I||_L\leq \alpha < 1$. We will show  Proposition \ref{thm:resnet_lipschitz} by first showing:
\begin{align}
(1-\alpha)||\bx-\bx'|| \leq ||h_l(\bx)-h_l(\bx')|| \leq (1+\alpha)||\bx-\bx'||,
\label{eq:bilip_single_original}
\end{align}
which is the bi-Lipschitz condition for a single residual block.

First show the left hand side:
\begin{align*}
    ||\bx - \bx'||
    &\leq ||\bx  - \bx' - (h_l(\bx) - h_l(\bx')) + (h_l(\bx) - h_l(\bx'))||
    \nonumber \\
    &\leq ||(h_l(\bx')-\bx') - (h_l(\bx) - \bx)|| + ||h_l(\bx) - h_l(\bx')||
    \nonumber \\
    &\leq ||g_l(\bx') - g_l(\bx)|| + ||h_l(\bx) - h_l(\bx')||
    \nonumber \\
    &\leq \alpha||\bx' - \bx|| + ||h_l(\bx) - h_l(\bx')||,
\end{align*}
where the last line follows by the assumption $||g_l||_L \leq \alpha$. Rearranging, we get:
\begin{align}
    (1 - \alpha)||\bx - \bx'|| \leq ||h_l(\bx) - h_l(\bx')||.
    \label{eq:bilip_lhs}
\end{align}

Now show the right hand side:
\begin{align*}
    ||h_l(\bx) - h_l(\bx')|| 
    &= ||\bx + g_l(\bx) - (\bx' + g_l(\bx'))||
    \leq ||\bx  - \bx'|| + ||g_l(\bx) - g_l(\bx'))||
    \leq (1 + \alpha) ||\bx  - \bx'||.
    \label{eq:bilip_rhs}
\end{align*}
Combining (\ref{eq:bilip_lhs})-(\ref{eq:bilip_rhs}), we have shown (\ref{eq:bilip_single_original}), which also implies:
\begin{align}
(1-\alpha)||\bx-\bx'||_X \leq ||h_l(\bx)-h_l(\bx')||_H \leq (1+\alpha)||\bx-\bx'||_X
\end{align}

Now show the bi-Lipschitz condition for a $L$-layer residual network $h=h_L\circ h_{L-1} \circ \dots \circ h_1$. It is easy to see that by induction:
\begin{align}
(1-\alpha)^L||\bx-\bx'||_X \leq ||h(\bx)-h(\bx')||_H \leq (1+\alpha)^L||\bx-\bx'||_X
\end{align}
Denoting $L_1 = (1-\alpha)^L$ and $L_2=(1+\alpha)^L$, we have arrived at  expression in Proposition \ref{thm:resnet_lipschitz}.
\hfill\BlackBox

\subsection{Proof of Lemma \ref{thm:minimax_lemma}}
\label{sec:minimax_lemma_proof}
\noindent{\it Proof Sketch.\quad}
This proof is an application of the generalized maximum entropy theorem to the case of Bregman score. We shall first state the generalized maximum entropy theorem to make sure the proof is self-contained. Briefly, the generalized maximum entropy theorem verifies that for a general scoring function $s(p, p^*|\bx)$ with entropy function $H(p|\bx)$, the maximum-entropy distribution $p' = \underset{p}{argsup} \, H(p|\bx)$ attains the minimax optimality :

\begin{theorem}[Maximum Entropy Theorem for General Loss  \citep{grunwald_game_2004}]
Let $\Psc$ be a convex, weakly closed and tight set of distributions. Consider a general score function $s(p, p^*|\bx)$ with an associated entropy function defined as $H(p|\bx) = \inf_{p^*\in \Psc^*} s(p, p^*|\bx)$. Assume below conditions on $H(p|\bx)$ hold:
\begin{itemize}[leftmargin=2em]
    \item (Well-defined) For any $p \in \Psc$, $H(p|\bx)$ exists and is finite. 
    \item (Lower-semicontinous) For a weakly converging sequence $p_n \rightarrow p_0 \in \Psc$ where $H(p_n|\bx)$ is bounded below, we have $s(p, p_0|\bx) \leq \liminf_{n\rightarrow \infty} s(p, p_n|\bx)$ for all $p \in \Psc$.
\end{itemize}
Then there exists an maximum-entropy distribution $p'$ such that 
$$p' = \sup_{p \in \Psc}H(p)=\sup_{p \in \Psc} \inf_{p^*\in \Psc^*} s(p, p^*|\bx) = \inf_{p \in \Psc} \sup_{p^* \in \Psc^*} s(p, p^*|\bx).$$
\label{thm:minimax_general}
\end{theorem}
Above theorem states that the maximum-entropy distribution attains the minimax optimality for a scoring function $s(p, p^*|\bx)$, assuming its entropy function satisfying certain regularity conditions. Authors of \citep{grunwald_game_2004}  showed that the entropy function of a Bregman score satisfies conditions in Theorem 1. Consequently, to show that the discrete uniform distribution is minimax optimal for Bregman score at $\bx \not\in \Xsc_{\texttt{IND}}$, we only need to show discrete uniform distribution is the maximum-entropy distribution.

Recall the definition of the \textit{strictly} proper Bregman score \citep{parry_proper_2012}:
\begin{align}
    s(p, p^*|\bx)= \sum_{k=1}^K 
    \Big\{
    [p^*(y_k|\bx) - p(y_k|\bx)]\psi'(p^*(y_k|\bx)) - \psi(p^*(y_k|\bx))
    \Big\}
\end{align}
where $\psi$ is differentiable and \textit{strictly} concave. Moreover, its entropy function is:
\begin{align}
    H(p|\bx) = -\sum_{k=1}^K \psi(p(y_k|\bx))
    \label{eq:bregman_entropy}
\end{align}
Our interest is to show that for $\bx \in \Xsc_{\texttt{OOD}}$, the maximum-entropy distribution for the Bregman score is the discrete uniform distribution $p(y_k|\bx)=\frac{1}{K}$. To this end, we notice that in the absence of any information, the only constraint on the predictive distribution is that $\sum_k p(y_k|\bx) = 1$. Therefore, denoting $p(y_k|\bx)=p_k$, we can set up the optimization problem with respect to Bregman entropy (\ref{eq:bregman_entropy}) using the Langrangian form below:
\begin{align}
L(p|\bx) &= H(p|\bx) + \lambda * (\sum_k p_k - 1)
= -\sum_{k=1}^K \psi(p_k) + \lambda * (\sum_k p_k - 1)
\end{align}
Taking derivative with respect to $p_k$ and $\lambda$:
\begin{align}
    \deriv{p_k} L &= -\psi'(p_k) + \lambda = 0\\
    \deriv{\lambda} L &= \sum_{k=1}^K p_k - 1 = 0
\end{align}
Notice that since $\psi(p)$ is \textit{strictly} concave, the function $\psi'(p)$ is monotonically decreasing and therefore invertible.
As a result, to solve the maximum entropy problem, we can solve the above systems of equation by finding a inverse function $\psi^{' -1}(p)$, which lead to the simplification:
\begin{align}
    p_k &= \psi^{' -1}(\lambda);  \quad \quad \sum_{k=1}^K p_k = 1.
\end{align}
Above expression essentially states that all $p_k$'s should be equal and sum to 1. The only distribution satisfying the above is the discrete uniform distribution, i.e.,  $p_k = \frac{1}{K} \; \forall k$.
\hfill\BlackBox

\end{document}